Takaaki Fujita • Florentin Smarandache

# Theoretical Foundations of Superhypergraph and Plithogenic Graph Neural Networks

Revised and Expanded Edition

**Takaaki Fujita, Florentin Smarandache**

# Theoretical Foundations of Superhypergraph and Plithogenic Graph Neural Networks

(Revised and Expanded Edition)

*Editor:*

Neutrosophic Science International Association (NSIA)
Publishing House
https://fs.unm.edu/NSIA/

Division of Mathematics and Sciences
University of New Mexico
705 Gurley Ave., Gallup Campus
NM 87301, United States of America

University of Guayaquil
Av. Kennedy and Av. Delta
"Dr. Salvador Allende" University Campus
Guayaquil 090514, Ecuador

***Peer-Reviewers:***

**Maikel Leyva-Vázquez**
Universidad de Guayaquil, Guayas, ECUADOR
maikel.leyvav@ug.edu.ec

**Victor Christianto**
Malang Institute of Agriculture (IPM), Malang, INDONESIA
victorchristianto@gmail.com

# Contents in this book

# Abstract

Hypergraphs generalize classical graphs by allowing an edge to connect multiple vertices, and thereby provide a natural language for modeling higher-order interactions. Superhypergraphs further extend this paradigm by accommodating nested, set-valued entities and relations, enabling the representation of hierarchical and multi-level structures that cannot be captured by ordinary graphs or hypergraphs. In parallel, neural networks—especially Graph Neural Networks (GNNs)—have become a standard tool for learning from relational data, and recent years have witnessed rapid developments of Hypergraph Neural Networks (HGNNs) and their theoretical properties.

To model uncertainty and multi-aspect attributes in complex networks, several graded and multi-valued graph frameworks have been introduced, including fuzzy graphs and neutrosophic graphs. The plithogenic graph framework unifies and refines these approaches by incorporating multi-valued attributes together with membership and contradiction mechanisms, offering a flexible representation for heterogeneous and partially inconsistent information.

This book develops theoretical foundations for SuperHyperGraph Neural Networks (SHGNNs) and Plithogenic Graph Neural Networks, aiming to extend neural message-passing principles to these advanced higher-order structures. We provide rigorous definitions, establish core structural properties, and prove well-definedness results for key constructions, with particular emphasis on strengthened formulations of Soft Graph Neural Networks and Rough Graph Neural Networks. While the present volume focuses on mathematical development rather than empirical evaluation, it also outlines directions for future computational experiments.

This book is an expanded and improved version of our earlier paper [1]. The revision refines definitions, theorems, and proofs; streamlines passages that were overly elementary; improves the coverage and accuracy of the related literature; and enhances the overall presentation and readability.

# 1 Introduction

## 1.1 Hypergraphs and SuperHyperGraphs

Graph theory studies relational structures consisting of a set of vertices (nodes) and a set of edges (links) between them [2]. Because graphs encode adjacency and dependency patterns in a mathematically explicit way, they serve as a common modeling language across many disciplines, including:

- Computational sciences: circuit design, data-flow analysis, and combinatorial optimization [3];
- Chemistry and biology: molecular graph theory and interaction networks in bioinformatics [4, 5];
- Project management: workflow representation, dependency analysis, and scheduling [6, 7];
- Probabilistic modeling: Bayesian networks for conditional dependence structures [8, 9];
- Graph databases: storage and query of highly connected data [10].

A hypergraph is a standard generalization of a graph in which an edge may join any nonempty subset of the vertex set [11, 12]. Hypergraphs naturally model higher-order relations and have found applications in areas such as machine learning, biology, and social network analysis (e.g., [13, 14, 15]). From a set-theoretic viewpoint, a hyperedge is simply a subset of the vertex set; thus, the edge set of a hypergraph is typically a subfamily of the powerset of its vertices.

More recently, SuperHyperGraphs have been proposed as higher-order extensions obtained by iterating powerset-type constructions and enriching them with suitable operations and relations [16, 17, 18]. SuperHyperGraphs model hierarchical multiway interactions via iterated powersets, unifying graphs/hypergraphs, enabling richer invariants, algorithms, and GNN architectures for multiscale uncertain relational data. For reference, a concise comparison of graphs, hypergraphs, and superhypergraphs is provided in Table 1.1.

Table 1.1: Concise comparison of graphs, hypergraphs, and superhypergraphs.

| Object | Notation | Vertex domain | Edge (hyperedge/superedge) type |
|---|---|---|---|
| Graph | $G = (V, E)$ | $V$ (atoms) | $E \subseteq \{\{u, v\} \subseteq V \mid u \neq v\}$ (each edge joins exactly two vertices). |
| Hypergraph | $H = (V, \mathcal{E})$ | $V$ (atoms) | $\mathcal{E} \subseteq \mathcal{P}(V) \setminus \{\emptyset\}$ (each hyperedge joins any nonempty subset of $V$). |
| SuperHyperGraph | $\mathrm{SHG} = (V, E)$ over $V_0$ | $V \subseteq \mathcal{P}(V_0) \setminus \{\emptyset\}$ (set-valued vertices) | $E \subseteq \mathcal{P}(V) \setminus \{\emptyset\}$ (each superedge joins a nonempty family of supervertices). |

In this perspective, a SuperHyperGraph may be viewed as an instance of a broader SuperHyperStructure, where the underlying “universe” consists of set-valued objects living at higher levels of an iterated powerset hierarchy.

Definition 1.1.1 ($n$-th powerset and $n$-th nonempty powerset). (cf. [18, 19]) Let $H$ be a set and let $\mathcal{P}(H)$ denote its powerset. Define the iterated powersets $\mathcal{P}^n(H)$ for $n \geq 0$ by

$$\mathcal{P}^0(H) := H, \qquad \mathcal{P}^{n+1}(H) := \mathcal{P}(\mathcal{P}^n(H)).$$

Likewise, define the iterated nonempty powersets $\mathcal{P}_*^n(H)$ by

$$\mathcal{P}_*^0(H) := H, \qquad \mathcal{P}_*^{n+1}(H) := \mathcal{P}(\mathcal{P}_*^n(H)) \setminus \{\emptyset\}.$$

Definition 1.1.2 (SuperHyperStructure). (cf. [18, 19]) Let $H$ be a nonempty set and fix $n \in \mathbb{N}$. A SuperHyperStructure of level $n$ is a pair

$$\mathcal{S} = (\mathcal{P}_*^n(H), \mathcal{O}),$$

where $\mathcal{O}$ is a specified family of operations and/or relations on $\mathcal{P}_*^n(H)$. Elements of $\mathcal{O}$ are called SuperHyperOperators.

Example 1.1.3 (Binary SuperHyperOperator). (cf. [18, 19]) A typical SuperHyperOperator is a binary mapping

$$\circ : \mathcal{P}_*^n(H) \times \mathcal{P}_*^n(H) \longrightarrow \mathcal{P}_*^n(H).$$

For instance, one may define $\circ$ by specifying a rule $f : \mathcal{P}_*^n(H) \times \mathcal{P}_*^n(H) \to \mathcal{P}_*^n(H)$ and setting $A \circ B := f(A, B)$. Concrete choices of $f$ depend on the intended application and the algebraic or relational axioms imposed on $\mathcal{O}$.

SuperHyperStructures encompass a variety of higher-order frameworks, including superhyperalgebras [20, 21, 22], superhypertopology [23, 24], superhyperfunctions [25, 26], and superhypersoft sets [27, 28]. Accordingly, the study of hypergraphs and SuperHyperGraphs is motivated both by foundational considerations—higher-order relational modeling and iterated set constructions—and by applications in domains where multi-way or hierarchical interactions must be represented explicitly. For reference, a comparison of structures, hyperstructures, and superhyperstructures is presented in Table 1.2.

Table 1.2: Concise comparison of structures, hyperstructures, and superhyperstructures.

| Concept | Carrier (universe) | Typical operation signature |
|---|---|---|
| Structure | $H$ | $\#_0 : H^m \to H$ |
| Hyperstructure | $H$ | $\#_1 : H^m \to \mathcal{P}^*(H)$ |
| Superhyperstructure (level $n \geq 2$) | $H$ | $\#_n : H^m \to \mathcal{P}^{\langle n \rangle}(H)$ |

Here $\mathcal{P}^*(H) := \mathcal{P}(H) \setminus \{\emptyset\}$ is the nonempty powerset. The nonempty powerset tower $\big(\mathcal{P}^{\langle n \rangle}(H)\big)_{n \geq 0}$ is defined by

$$\mathcal{P}^{\langle 0 \rangle}(H) := H, \qquad \mathcal{P}^{\langle n+1 \rangle}(H) := \mathcal{P}(\mathcal{P}^{\langle n \rangle}(H)) \setminus \{\emptyset\}.$$

Thus, $\#_1$ is set-valued (hyperoperation), while $\#_n$ is higher-order set-valued, returning elements of an iterated nonempty powerset.

## 1.2 Graph Neural Networks

This subsection briefly reviews Graph Neural Networks (GNNs). In recent years, machine learning [29, 30, 3], artificial intelligence [31, 32], and big-data analytics [33, 34] have become central research areas, and neural-network methods play a key role in many of these developments.

A neural network is a computational model inspired by biological neural systems and is widely used for tasks such as representation learning, pattern recognition, classification, and prediction [35, 36]. A graph neural network extends this paradigm to data endowed with a graph structure by learning from node/edge features together with the graph connectivity, thereby enabling models that explicitly exploit relational information [37, 38, 39].

Several important variants have been proposed to capture richer relational patterns. Hypergraph neural networks (HGNNs) generalize GNNs by replacing pairwise edges with hyperedges, allowing the model to represent higher-order interactions among multiple vertices simultaneously [40, 41]. Closely related ideas include hypernetworks, which have been studied as a mechanism for parameter generation and adaptive architectures [42, 43]. Moreover, GNNs on directed graphs (e.g., directed graph neural networks) [44, 45] and models for mixed graph structures (e.g., mixed graph neural networks) [46] have also attracted attention.

Given the breadth of applications and the need to learn from structured relational data, the further development of GNN theory and methodology remains an important research direction.

## 1.3 Uncertain Graphs

The notion of a fuzzy set was introduced by Zadeh in 1965 [47]. Fuzzy sets provide a flexible mathematical language for representing graded membership and have been widely applied to uncertainty modeling in areas such as graph theory, algebra, topology, and logic. To capture richer and more intricate forms of uncertainty, several extensions have been proposed, including Intuitionistic Fuzzy Sets[48], vague sets[49, 50], neutrosophic sets [51, 52], double-valued neutrosophic sets[53, 54], hesitant fuzzy sets[55, 56], picture fuzzy sets[57, 58], spherical fuzzy sets[59, 60, 61], plithogenic sets[62, 63], and uncertain set[64]. which explicitly distinguish different sources of imprecision.

These uncertainty-aware formalisms are well aligned with practical needs, since real data are often incomplete, noisy, or mutually inconsistent [65, 66]. In particular, a neutrosophic set assigns to each element three membership degrees—truth, indeterminacy, and falsity—and is therefore useful when evidence simultaneously supports, contradicts, or fails to determine a statement. Representative application areas include:

- Healthcare decision-making: balancing effectiveness (truth), uncertainty (indeterminacy), and risk (falsity) when clinical information is incomplete or conflicting [67, 68];
- Social network analysis: modeling relations such as trust, suspicion, and disagreement under ambiguous or inconsistent observations [69, 70];
- Fault diagnosis in engineering: incorporating uncertain and contradictory diagnostic signals when identifying faults in mechanical and industrial systems [71, 72, 73];
- Market analysis: integrating positive feedback (truth), ambiguous responses (indeterminacy), and negative feedback (falsity) in customer-preference modeling [74, 75, 76].

Motivated by these developments, this book considers several models of uncertain graphs, including fuzzy, intuitionistic fuzzy, neutrosophic, and plithogenic graphs. Such models extend classical graph theory by attaching degrees of uncertainty to vertices and/or edges, enabling finer analysis of vague, heterogeneous, and complex relational information [77].

Typical examples of uncertain graph models include:

- Fuzzy graph: a graph in which membership functions quantify the presence or strength of vertices and edges, thereby supporting graded adjacency and flexible relationship modeling [78, 79]. Related concepts include intuitionistic fuzzy graphs[80], picture fuzzy graphs [81, 82, 83], spherical fuzzy graphs[84, 60], hesitant fuzzy graphs[85, 86], bipolar fuzzy graphs[87, 79], and $m$-polar fuzzy graphs[88, 89, 90].
- Neutrosophic graph: a refinement of the fuzzy-graph paradigm in which each vertex and edge is described by three degrees (truth, indeterminacy, falsity), yielding a richer representation of uncertainty and inconsistency [91, 92]. It is well known that neutrosophic graphs generalize fuzzy graphs. Moreover, related notions such as bipolar neutrosophic graphs [93, 94] and complex neutrosophic graphs [95, 96] have also been studied.
- Plithogenic graph: a graph model equipped with multi-valued attributes together with membership and contradiction functions, designed to represent complex relations with heterogeneous and possibly conflicting criteria [97, 98, 62]. It is widely recognized that plithogenic graphs generalize neutrosophic graphs.

Uncertain graph models, and their set-based foundations more broadly, have been used in decision-making frameworks [60], and they also appear in learning-based settings such as neural networks [99, 100, 101] and machine learning [102]. These connections further motivate the systematic study of uncertain graphs and their associated mathematical structures. For convenience, Table 1.3 summarizes, in a unified notation, the canonical information assigned to vertices and edges in several representative graph extensions.

Table 1.3: Representative graph extensions and the canonical information stored on vertices and/or edges.

| Graph Type | Canonical data attached to vertices/edges |
|---|---|
| Fuzzy Graph | Vertex membership $\sigma : V \to [0,1]$ and edge membership $\mu : E \to [0,1]$ (typically with $\mu(uv) \le \sigma(u) \wedge \sigma(v)$). |
| Intuitionistic Fuzzy Graph | Vertex degrees $(\mu_A, \nu_A) : V \to [0,1]^2$ and edge degrees $(\mu_B, \nu_B) : E \to [0,1]^2$ with $\mu + \nu \le 1$; the residual represents hesitation. |
| Neutrosophic Graph | Vertex triple $(T_A, I_A, F_A) : V \to [0,1]^3$ and edge triple $(T_B, I_B, F_B) : E \to [0,1]^3$ (truth, indeterminacy, falsity). |
| Quadripartitioned Neutrosophic Graph | Vertex quadruple $(T, C, U, F) : V \to [0,1]^4$ and edge quadruple $(T, C, U, F) : E \to [0,1]^4$, typically encoding truth, contradiction, unknown, and falsity. |
| Pentapartitioned Neutrosophic Graph | Vertex quintuple $(T, C, U, F, S) : V \to [0,1]^5$ and edge quintuple $(T, C, U, F, S) : E \to [0,1]^5$, i.e., a five-component refinement of neutrosophic information. |
| Plithogenic Graph | Vertex structure $PM = (M, \ell, M_\ell, \mathrm{adf}, \mathrm{aCf})$ and edge structure $PN = (N, m, N_m, \mathrm{bdf}, \mathrm{bCf})$, where $\mathrm{adf} : M \times M_\ell \to [0,1]^s$ and $\mathrm{bdf} : N \times N_m \to [0,1]^s$ encode $s$-dimensional appurtenance, while aCf and bCf are symmetric contradiction maps in $[0,1]^t$. |

## 1.4 Our Contribution

In this book, we introduce the SuperHyperGraph Neural Network (SHGNN), a mathematical extension of Hypergraph Neural Networks that leverages the unique structural properties of SuperHyperGraphs. Additionally, we examine uncertain graph neural models, such as Neutrosophic Graph Neural Networks and Plithogenic Graph Neural Networks, which address similar challenges. Importantly, we demonstrate that both Neutrosophic and Plithogenic Graph Neural Networks serve as mathematical generalizations of Fuzzy Graph Neural Networks.

This work is theoretical in nature, focusing on establishing the mathematical framework for SHGNNs and PGNNs. It does not include computational experiments or practical implementations. Therefore, we hope that computational experiments will be conducted in the future by experts and readers alike. For precise definitions and detailed notations, readers are encouraged to consult the relevant literature, such as [40]. In this book, we conduct a theoretical examination of the relationships between Graph Neural Networks.

Although not directly related to the Graph Neural Networks discussed earlier, this book also explores several extended concepts in hypergraph theory, including Multilevel k-way Hypergraph Partitioning, Superhypergraph Random Walk, and the Superhypergraph Turán Problem. As these investigations are limited to theoretical considerations, it is hoped that computational experiments and practical validations will be conducted in the future as needed. For reference, a comparison of GNN, HGNN, and SHGNN is provided in Table 1.4.

Table 1.4: Concise comparison of GNN, HGNN, and SHGNN (incidence-based formulation).

| Aspect | Graph Neural Network (GNN) | Hypergraph Neural Network (HGNN) | SuperHyperGraph Neural Network (SHGNN) |
|---|---|---|---|
| Underlying object | Graph $G = (V, E)$ | Hypergraph $H = (V, E)$ | $n$-SuperHyperGraph $\mathrm{SHG}^{(n)} = (V_n, E)$ over a base set $V_0$ |
| Vertex type | atomic vertices $v \in V$ | atomic vertices $v \in V$ | supervertices $v \in V_n \subseteq \mathcal{P}^n(V_0)$ (set-valued, hierarchical) |
| Edge type | pairwise edges $e = \{u, v\}$ | hyperedges $e \subseteq V$ | superedges $e \subseteq V_n$ (relations among supervertices) |
| Linear operator used for propagation | adjacency/Laplacian ($A$, $L$, normalized variants) | incidence-based operator $D_V^{-1/2} H W D_E^{-1} H^\top D_V^{-1/2}$ | typically reduced to an expanded hypergraph on $V_0$: $\mathrm{Exp}^{(n)}(\mathrm{SHG}^{(n)}) = (V_0, E^{\mathrm{exp}})$, then apply HGNN operator on $(V_0, E^{\mathrm{exp}})$ |
| Message-passing semantics | aggregate from 1-hop neighbors in $G$ | alternate vertex–hyperedge–vertex aggregation via incidence $H$ | aggregate via superedges after expansion (base-vertex messages induced by set-valued supervertices/superedges) |
| Typical representation domain | node embeddings $h_v \in \mathbb{R}^d$ for $v \in V$ | node embeddings $h_v \in \mathbb{R}^d$ for $v \in V$ (hyperedge embeddings optional) | base-vertex embeddings $h_x \in \mathbb{R}^d$ for $x \in V_0$ (or supervertex embeddings if working directly on $V_n$) |
| Computational footprint (per layer, schematic) | $\mathcal{O}(\|E\|d)$ (sparse graphs) | $\mathcal{O}(\mathrm{nnz}(H)\, d)$ | $\mathcal{O}(\mathrm{nnz}(H^{\mathrm{exp}})\, d)$ after expansion (may increase with expansion size) |

Remark. Here, "SHGNN" refers to the common incidence-based approach that first maps an $n$-superhypergraph to an expanded hypergraph on the base vertex set $V_0$ and then applies HGNN-style propagation. Alternative SHGNN designs can operate directly on $V_n$ without expansion.

This book is an expanded and improved version of our earlier paper [1]. It primarily focuses on strengthening the developments of Soft Graph Neural Networks and Rough Graph Neural Networks; refining the mathematical content of definitions, theorems, and proofs; revising or streamlining overly simplistic passages; improving the coverage and accuracy of the referenced literature; and enhancing overall presentation and readability. Please note that this book is devoted primarily to theoretical analysis. The authors hope that future work will advance quantitative investigations and real-world applications based on the concepts introduced in this book.

# 2 Preliminaries

In this section, we provide a brief overview of the definitions and notations used throughout this book. While we aim to make the content accessible to readers from various backgrounds, it is not possible to cover all relevant details comprehensively. Readers are encouraged to consult the referenced literature for additional information as needed.

## 2.1 Basic Graph Neural Network Concepts

This section collects several basic definitions and notational conventions used in graph neural networks (GNNs). For broader background and further discussion, see, e.g., the lecture notes and surveys in [103, 104, 105].

Definition 2.1.1 (Adjacency matrix). (cf. [106, 107, 108]) Let $G = (V, E)$ be a (simple) graph with vertex set $V = \{v_1, \ldots, v_n\}$. The adjacency matrix of $G$ is the $n \times n$ matrix $A = [a_{ij}]$ defined by

$$a_{ij} = \begin{cases} 1, & \text{if } (v_i, v_j) \in E, \\ 0, & \text{otherwise.} \end{cases}$$

(For weighted graphs, $a_{ij}$ is commonly replaced by the edge weight between $v_i$ and $v_j$.)

Definition 2.1.2 (Weight matrix). (cf. [109, 110]) In many learning models, a weight matrix encodes trainable linear map parameters. Let $\mathbf{X} \in \mathbb{R}^{n \times d}$ be a data matrix whose $n$ rows correspond to data points and whose $d$ columns correspond to input features. A weight matrix $\mathbf{W} \in \mathbb{R}^{d \times p}$ maps the input feature space $\mathbb{R}^d$ to an output space $\mathbb{R}^p$ via

$$\mathbf{Z} = \mathbf{XW},$$

where $\mathbf{Z} \in \mathbb{R}^{n \times p}$ is the resulting representation matrix. Equivalently, the entry $w_{ij}$ quantifies how strongly the $i$-th input feature contributes to the $j$-th output coordinate.

Definition 2.1.3 (Feature vector). (cf. [111, 112, 113]) Let $\mathcal{O}$ be an object (or observation) and let $F = \{f_1, \ldots, f_n\}$ be a collection of feature maps $f_i : \mathcal{O} \to \mathbb{R}$. The feature vector of $\mathcal{O}$ (with respect to $F$) is

$$\mathbf{x} = \left[f_1(\mathcal{O}),\ f_2(\mathcal{O}),\ \ldots,\ f_n(\mathcal{O})\right]^\top \in \mathbb{R}^n.$$

Definition 2.1.4 (Dataset). (cf. [114]) A dataset is a finite collection of data points. Formally, for an input space $\mathcal{X}$, one may write

$$D = \{\mathbf{x}_i \in \mathcal{X} \mid i = 1, 2, \ldots, n\},$$

where $n$ is the dataset size.

Definition 2.1.5 (Normalization). (cf. [115, 116]) Normalization rescales numerical values so that they lie in a prescribed range, often to stabilize learning. Given scalars $x_1, \ldots, x_n$, min–max normalization to $[0, 1]$ is defined by

$$x'_i = \frac{x_i - \min(x)}{\max(x) - \min(x)}, \qquad \min(x) = \min_{1 \le j \le n} x_j, \quad \max(x) = \max_{1 \le j \le n} x_j.$$

To rescale to $[-1, 1]$, one may use

$$x'_i = 2 \cdot \frac{x_i - \min(x)}{\max(x) - \min(x)} - 1.$$

Definition 2.1.6 (Graph neural network (GNN)). (cf. [117]) Let $G = (V, E)$ be a (finite) graph with $V = \{v_1, \ldots, v_n\}$ and $E \subseteq V \times V$. Assume that each node $v_i \in V$ is equipped with a feature vector $\mathbf{x}_i \in \mathbb{R}^{d_0}$, and that each edge $(v_i, v_j) \in E$ may optionally carry an edge feature $\mathbf{e}_{ij} \in \mathbb{R}^{k_0}$. A graph neural network (GNN) is a parametric family of functions that, for layers $t = 0, 1, \ldots, T$, computes node representations

$$\mathbf{h}_i^{(t)} \in \mathbb{R}^{d_t} \qquad (i = 1, \ldots, n)$$

by combining the graph structure (connectivity) with node/edge features, typically via neighborhood aggregation (message passing).

Definition 2.1.7 (Key components of a message-passing GNN). (cf. [117]) A common message-passing GNN is specified by the following steps.

1. Initialization. Set the initial hidden states to the input features:

$$\mathbf{h}_i^{(0)} = \mathbf{x}_i, \qquad i = 1, \ldots, n.$$

2. Message passing. For each layer $t \in \{0, \ldots, T-1\}$, node $v_i$ receives messages from its (graph) neighborhood $\mathcal{N}(i) \subseteq V$ (cf. [118, 119]) via

$$\mathbf{m}_i^{(t+1)} = \sum_{v_j \in \mathcal{N}(i)} \phi_m^{(t)}\big(\mathbf{h}_i^{(t)}, \mathbf{h}_j^{(t)}, \mathbf{e}_{ij}\big),$$

where $\phi_m^{(t)}$ is a learnable message function (its output dimension matches the input expected by the update map).

3. Node update. Each node state is updated using the aggregated message (cf. [120]):

$$\mathbf{h}_i^{(t+1)} = \phi_u^{(t)}\big(\mathbf{h}_i^{(t)}, \mathbf{m}_i^{(t+1)}\big),$$

where $\phi_u^{(t)}$ is a learnable update function.

4. Readout (graph-level representation). For graph-level tasks, a global embedding is obtained by applying a permutation-invariant readout to the final node states:

$$\mathbf{z}_G = \phi_r\left(\{\mathbf{h}_i^{(T)} \mid v_i \in V\}\right),$$

where $\phi_r$ is typically chosen as sum/mean/max pooling or another symmetric aggregation.

Example 2.1.8 (Common readout functions). (cf. [121, 122]) A readout function $\phi_r$ maps the multiset of final-layer node embeddings $\{\mathbf{h}_i^{(T)}\}_{i=1}^n$ to a single graph embedding $\mathbf{z}_G$. Typical permutation-invariant choices include:

- Mean readout (cf. [123, 124]):

$$\phi_r\left(\{\mathbf{h}_i^{(T)} \mid v_i \in V\}\right) = \frac{1}{|V|} \sum_{v_i \in V} \mathbf{h}_i^{(T)}.$$

- Max-pooling readout (cf. [125, 126, 127]):

$$\phi_r\left(\{\mathbf{h}_i^{(T)} \mid v_i \in V\}\right) = \max_{v_i \in V} \mathbf{h}_i^{(T)},$$

where max is applied element-wise across coordinates.

- Sum readout (cf. [128, 129]):

$$\phi_r\left(\{\mathbf{h}_i^{(T)} \mid v_i \in V\}\right) = \sum_{v_i \in V} \mathbf{h}_i^{(T)}.$$

Definition 2.1.9 (Matrix-form framework). (cf. [117]) Let $\mathbf{H}^{(t)} \in \mathbb{R}^{n\times d_t}$ denote the matrix whose $i$-th row is $(\mathbf{h}_i^{(t)})^\top$. Let $\mathbf{A} \in \mathbb{R}^{n\times n}$ be the (possibly weighted) adjacency matrix of $G$. A broad class of GNN layers can be written abstractly as

$$\mathbf{H}^{(t+1)} = \Phi^{(t)}\big(\mathbf{A}, \mathbf{H}^{(t)}; \mathbf{W}^{(t)}\big),$$

where $\Phi^{(t)}$ is a differentiable aggregation–transformation map parametrized by learnable weights $\mathbf{W}^{(t)}$.

Definition 2.1.10 (Graph convolutional network (GCN)). (cf. [130]) A (Kipf–Welling style) graph convolutional network layer is given by

$$\mathbf{H}^{(t+1)} = \sigma\Big(\widehat{\mathbf{A}}\,\mathbf{H}^{(t)}\mathbf{W}^{(t)}\Big),$$

where $\sigma$ is an element-wise activation function (e.g., ReLU), $\mathbf{W}^{(t)} \in \mathbb{R}^{d_t\times d_{t+1}}$ is trainable, and

$$\widetilde{\mathbf{A}} = \mathbf{A} + \mathbf{I}, \qquad \widetilde{\mathbf{D}} = \operatorname{diag}\left(\sum_{j=1}^{n} \widetilde{a}_{ij}\right)_{i=1}^{n}, \qquad \widehat{\mathbf{A}} = \widetilde{\mathbf{D}}^{-1/2}\,\widetilde{\mathbf{A}}\,\widetilde{\mathbf{D}}^{-1/2}.$$

Here $\mathbf{I}$ is the $n \times n$ identity matrix (adding self-loops), $\widetilde{\mathbf{D}}$ is the degree matrix of $\widetilde{\mathbf{A}}$, and $\widehat{\mathbf{A}}$ is the symmetrically normalized adjacency.

Example 2.1.11 (Intuition for a GCN layer). Consider a social network (cf. [131]) in which each person is a node and friendships are edges. Suppose each person $v_i$ has an initial feature vector $\mathbf{x}_i$ encoding, for example, interests in topics such as sports, music, and technology. A GCN layer updates the representation of each person by:

1. aggregating information from neighbors (and from the node itself, via self-loops);
2. weighting these contributions through the normalized matrix $\widehat{\mathbf{A}}$, which balances the influence of high- and low-degree nodes;
3. applying a learned linear transform $\mathbf{W}^{(t)}$ followed by a nonlinearity $\sigma$.

Stacking multiple layers allows each node representation to incorporate information from progressively larger neighborhoods (e.g., friends-of-friends after two layers), supporting tasks such as node classification, graph classification, and link prediction.

## 2.2 Hypergraph Concepts

A hypergraph extends the classical notion of a graph by allowing an edge (called a hyperedge) to join an arbitrary number of vertices, rather than being restricted to pairs [11, 132]. This additional flexibility makes hypergraphs well suited for modeling higher-order interactions in areas such as computer science and biology [133, 134]. We recall the basic definitions below.

Definition 2.2.1 (Hypergraph). [11, 135] A (finite) hypergraph is a pair $H = (V(H), E(H))$ such that

- $V(H)$ is a finite, nonempty set (the vertices);
- $E(H) \subseteq \mathcal{P}(V(H)) \setminus \{\emptyset\}$ is a finite family of nonempty subsets of $V(H)$ (the hyperedges).

Example 2.2.2 (Hypergraph). Let $H = (V, E)$ be given by

$$V = \{v_1, v_2, v_3, v_4\}, \qquad E = \big\{\{v_1, v_2\},\ \{v_2, v_3, v_4\},\ \{v_1\}\big\}.$$

Then $\{v_1, v_2\}$ and $\{v_2, v_3, v_4\}$ are hyperedges of sizes 2 and 3, respectively, and $\{v_1\}$ is a singleton hyperedge.

Proposition 2.2.3. Every (simple) graph is a special case of a hypergraph.

Proof. Let $G = (V, E)$ be a (simple, undirected) graph, where $E \subseteq \binom{V}{2}$. Define a hypergraph $H = (V, E)$ by viewing each edge $\{u, v\} \in E$ as a 2-element hyperedge. Then $E \subseteq \mathcal{P}(V) \setminus \{\emptyset\}$, so $H$ is a hypergraph, and its hyperedges coincide with the edges of $G$. □

Definition 2.2.4 (Induced subhypergraph). [135] Let $H = (V, E)$ be a hypergraph and let $X \subseteq V$. The induced subhypergraph on $X$ is

$$H[X] := \big(X,\ E[X]\big), \qquad E[X] := \{\, e \in E \mid e \subseteq X \,\}.$$

Moreover, the hypergraph obtained by deleting the vertices in $X$ is

$$H \setminus X := H[V \setminus X].$$

## 2.3 $n$-SuperHyperGraphs

An $n$-SuperHyperGraph is a higher-order combinatorial structure in which supervertices may live at the $n$-th level of an iterated (nonempty) powerset tower, so that vertices can be set-valued, set-of-set-valued, and so on. One convenient formalization starts from a finite base vertex set $V_0$ and lets supervertices be elements of $\mathcal{P}^{\langle n \rangle}(V_0)$, while superedges are (nonempty) subsets of the chosen supervertex family (cf. [17, 136, 137]).

Definition 2.3.1 ($n$-SuperHyperGraph). (cf. [17]) Fix $n \in \mathbb{N}_0$ and let $V_0$ be a finite, nonempty base set. An $n$-SuperHyperGraph over $V_0$ is an ordered pair $H^{(n)} = (V, E)$ such that

- $V \subseteq \mathcal{P}^{\langle n \rangle}(V_0)$ is a finite set whose elements are called supervertices;
- $E \subseteq \mathcal{P}(V) \setminus \{\emptyset\}$ is a finite set whose elements are called superedges.

Example 2.3.2 (2-SuperHyperGraph). Let $V_0 = \{x_1, x_2, x_3\}$ and $n = 2$. Since $\mathcal{P}^{\langle 1 \rangle}(V_0) = \mathcal{P}(V_0) \setminus \{\emptyset\}$, an element of $\mathcal{P}^{\langle 2 \rangle}(V_0)$ is a nonempty family of nonempty subsets of $V_0$. Define supervertices

$$v_1 := \big\{\{x_1, x_2\}, \{x_3\}\big\}, \qquad v_2 := \big\{\{x_1\}\big\}, \qquad v_3 := \big\{\{x_2, x_3\}\big\},$$

and set

$$V := \{v_1, v_2, v_3\}, \qquad E := \big\{\{v_1, v_2\},\ \{v_1, v_3\}\big\}.$$

Then $H^{(2)} = (V, E)$ is a 2-SuperHyperGraph over $V_0$. For reference, an overview schematic is provided in Fig. 2.1.

To compare an $n$-SuperHyperGraph with an ordinary hypergraph on the base set $V_0$, we use a canonical flattening map that turns an $n$-level object into a subset of $V_0$.

Definition 2.3.3 (Flattening map). Fix $n \in \mathbb{N}_0$ and a base set $V_0$. Define maps $\mathrm{flat}_k : \mathcal{P}^{\langle k \rangle}(V_0) \to \mathcal{P}(V_0) \setminus \{\emptyset\}$ recursively by

$$\mathrm{flat}_0(x) := \{x\} \quad (x \in V_0), \qquad \mathrm{flat}_{k+1}(U) := \bigcup_{u \in U} \mathrm{flat}_k(u) \quad \big(U \in \mathcal{P}^{\langle k+1 \rangle}(V_0)\big).$$

Definition 2.3.4 (Expanded hypergraph of an $n$-SuperHyperGraph). Let $H^{(n)} = (V, E)$ be an $n$-SuperHyperGraph over $V_0$. Its expanded hypergraph is the hypergraph on the base vertex set $V_0$ defined by

$$\mathrm{Exp}_n(H^{(n)}) := (V_0, E^{\exp}), \qquad E^{\exp} := \Big\{ \bigcup_{v \in e} \mathrm{flat}_n(v) \ \Big|\ e \in E \Big\}.$$

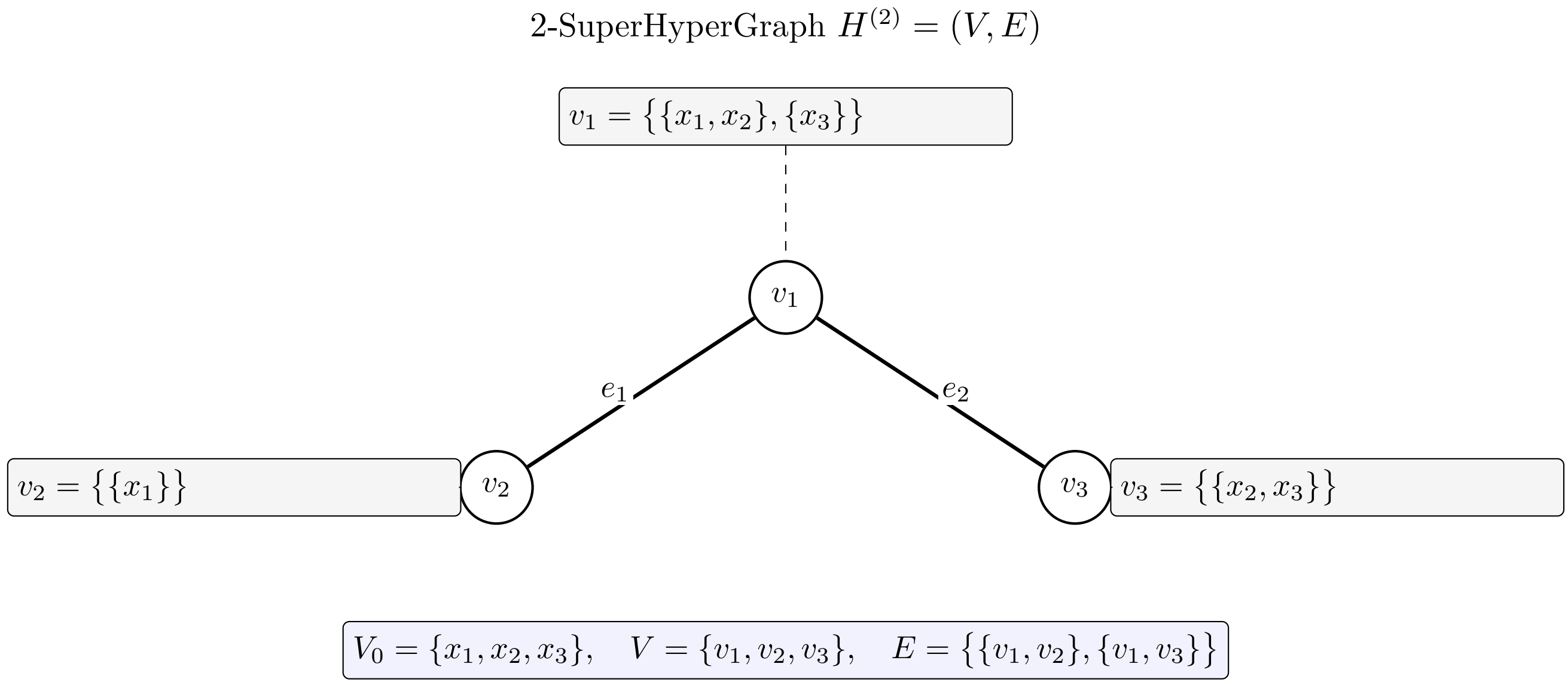

Figure 2.1: A schematic illustration of the 2-SuperHyperGraph in Example 2.3.2.

Example 2.3.5 (Expansion in Example 2.3.2). For the 2-SuperHyperGraph in Example 2.3.2,

$$\mathrm{flat}_2(v_1) = \{x_1, x_2, x_3\}, \quad \mathrm{flat}_2(v_2) = \{x_1\}, \quad \mathrm{flat}_2(v_3) = \{x_2, x_3\}.$$

Hence

$$\bigcup_{v \in \{v_1, v_2\}} \mathrm{flat}_2(v) = \{x_1, x_2, x_3\}, \qquad \bigcup_{v \in \{v_1, v_3\}} \mathrm{flat}_2(v) = \{x_1, x_2, x_3\},$$

.

Theorem 2.3.6. For every finite hypergraph $\mathcal{H} = (V_0, \mathcal{E})$ and every $n \in \mathbb{N}_0$, there exists an $n$-SuperHyperGraph $H^{(n)}$ over $V_0$ such that $\mathrm{Exp}_n(H^{(n)}) = \mathcal{H}$.

Proof. Fix $n \in \mathbb{N}_0$. Define an iterated singleton embedding $\iota_k : V_0 \to \mathcal{P}^{\langle k \rangle}(V_0)$ by

$$\iota_0(x) := x, \qquad \iota_{k+1}(x) := \{\iota_k(x)\} \quad (x \in V_0).$$

Set

$$V := \{\iota_n(x) \mid x \in V_0\} \subseteq \mathcal{P}^{\langle n \rangle}(V_0), \qquad E := \Big\{ \{\iota_n(x) \mid x \in e\} \;\Big|\; e \in \mathcal{E} \Big\}.$$

Then $H^{(n)} = (V, E)$ is an $n$-SuperHyperGraph.

We claim that $\mathrm{flat}_n(\iota_n(x)) = \{x\}$ for all $x \in V_0$. This follows by induction on $n$: for $n = 0$, $\mathrm{flat}_0(\iota_0(x)) = \{x\}$ by definition; and if true for $n$, then

$$\mathrm{flat}_{n+1}(\iota_{n+1}(x)) = \mathrm{flat}_{n+1}(\{\iota_n(x)\}) = \bigcup_{u \in \{\iota_n(x)\}} \mathrm{flat}_n(u) = \mathrm{flat}_n(\iota_n(x)) = \{x\}.$$

Now take any hyperedge $e \in \mathcal{E}$ and the corresponding superedge $\hat{e} = \{\iota_n(x) \mid x \in e\} \in E$. By the claim,

$$\bigcup_{v \in \hat{e}} \mathrm{flat}_n(v) = \bigcup_{x \in e} \mathrm{flat}_n(\iota_n(x)) = \bigcup_{x \in e} \{x\} = e.$$

Hence $E^{\exp} = \mathcal{E}$, i.e. $\mathrm{Exp}_n(H^{(n)}) = (V_0, \mathcal{E}) = \mathcal{H}$. □

Remark 2.3.7. For $n = 1$, we have $\mathcal{P}^{\langle 1 \rangle}(V_0) = \mathcal{P}(V_0) \setminus \{\emptyset\}$, so Definition 2.3.1 recovers the common "set-valued vertex" formalization of a SuperHyperGraph.

## 2.4 Some Classes of SuperHyperGraphs: Hierarchical and Recursive SuperHyperGraphs

In this section, we formalize two important extensions of the superhypergraph framework: hierarchical superhypergraphs and recursive superhypergraphs. The former allows supervertices to appear across multiple iterated-powerset levels, while the latter allows edges to be recursively nested up to a prescribed depth. We then combine these two viewpoints into a unified $(n, k)$-recursive SuperHyperGraph model.

A hierarchical superhypergraph is a superhypergraph whose vertices may lie on different powerset levels, and whose edges may connect supervertices from mixed levels, subject to a natural downward-closure coherence condition (cf. [138, 139]).

Definition 2.4.1 (Support map and edge support). Let $V_0$ be a finite nonempty base set, and let $r \geq 0$. Define the hierarchical universe

$$\mathcal{U}_r(V_0) := \bigcup_{k=0}^{r} \Big(\mathcal{P}^k(V_0) \setminus \{\emptyset\}\Big),$$

where

$$\mathcal{P}^0(V_0) := V_0, \qquad \mathcal{P}^{k+1}(V_0) := \mathcal{P}\big(\mathcal{P}^k(V_0)\big) \quad (k \geq 0).$$

The support map

$$\mathrm{supp} : \mathcal{U}_r(V_0) \longrightarrow \mathcal{P}(V_0) \setminus \{\emptyset\}$$

is defined recursively by

$$\mathrm{supp}(v) := \{v\} \quad (v \in V_0), \qquad \mathrm{supp}(X) := \bigcup_{y \in X} \mathrm{supp}(y) \quad (X \in \mathcal{U}_r(V_0) \setminus V_0).$$

If $e \subseteq \mathcal{U}_r(V_0)$ is nonempty, its base support is defined by

$$\sigma(e) := \bigcup_{x \in e} \mathrm{supp}(x) \in \mathcal{P}(V_0) \setminus \{\emptyset\}.$$

Definition 2.4.2 (Hierarchical SuperHyperGraph of height $r$). (cf. [138, 139]) Let $V_0$ be a finite nonempty base set, and let $r \geq 0$. A hierarchical SuperHyperGraph of height $r$ on $V_0$ is a pair

$$\mathbb{H}^{\langle r \rangle} = (V, E)$$

satisfying the following conditions:

(H1) (Hierarchical vertex set) $V$ is a finite nonempty set such that

$$V \subseteq \mathcal{U}_r(V_0).$$

Elements of $V$ are called hierarchical supervertices.

(H2) (Cross-level superhyperedges) $E$ is a finite set of nonempty subsets of $V$:

$$E \subseteq \mathcal{P}(V) \setminus \{\emptyset\}.$$

Elements of $E$ are called hierarchical superhyperedges. In particular, a superhyperedge may contain supervertices from different levels.

(H3) (Downward-closure coherence) If $X \in V$ and $X \notin V_0$ (equivalently, $X$ is a higher-level supervertex), then

$$X \subseteq V.$$

Thus, whenever a higher-level supervertex is present, all of its immediate constituents are also present as supervertices.

For $x \in \mathcal{U}_r(V_0)$, define its level by

$$\ell(x) := \min\{\, k \in \{0, 1, \dots, r\} : x \in \mathcal{P}^k(V_0)\,\}.$$

For each $k \in \{0, \dots, r\}$, define the $k$-th layer by

$$V_k := \{\, x \in V : \ell(x) = k\,\}.$$

Then $V = \bigcup_{k=0}^{r} V_k$, and this union is disjoint.

Remark 2.4.3. When $r = 0$, one has $\mathcal{U}_0(V_0) = V_0$, so a hierarchical SuperHyperGraph of height 0 reduces to an ordinary hypergraph on a (nonempty) vertex subset of $V_0$.

Example 2.4.4 (A hierarchical SuperHyperGraph of height $r = 2$). Let the base set be

$$V_0 = \{x_1, x_2, x_3, x_4\}.$$

We take $r = 2$, so

$$\mathcal{U}_2(V_0) = \Big(V_0\Big) \cup \Big(\mathcal{P}(V_0) \setminus \{\emptyset\}\Big) \cup \Big(\mathcal{P}^2(V_0) \setminus \{\emptyset\}\Big).$$

Step 1: Hierarchical supervertices (mixed levels). Define the following elements of $\mathcal{U}_2(V_0)$:

$$\begin{aligned} &\text{Level 0:} \quad v_1 := x_1,\ v_2 := x_2,\ v_3 := x_3,\ v_4 := x_4, \\ &\text{Level 1:} \quad v_5 := \{x_1, x_2\}, \qquad v_6 := \{x_3, x_4\}, \\ &\text{Level 2:} \quad v_7 := \big\{\{x_1, x_2\}, \{x_3, x_4\}\big\}. \end{aligned}$$

Set

$$V := \{v_1, v_2, v_3, v_4, v_5, v_6, v_7\} \subseteq \mathcal{U}_2(V_0).$$

Thus (H1) holds.

Step 2: Downward-closure coherence. The higher-level supervertices satisfy

$$v_5 = \{x_1, x_2\} \subseteq V, \qquad v_6 = \{x_3, x_4\} \subseteq V,$$

and

$$v_7 = \big\{\{x_1, x_2\}, \{x_3, x_4\}\big\} = \{v_5, v_6\} \subseteq V.$$

Hence (H3) holds.

Step 3: Cross-level superhyperedges. Define the following superhyperedges (each is a nonempty subset of $V$):

$$e_1 := \{v_5, v_6\}, \qquad e_2 := \{v_7, v_2, v_3\}, \qquad e_3 := \{v_1, v_5, v_4\}.$$

Set

$$E := \{e_1, e_2, e_3\} \subseteq \mathcal{P}(V) \setminus \{\emptyset\}.$$

Thus (H2) holds. Note that $e_2$ is cross-level: it contains the level-2 supervertex $v_7$ together with the level-0 vertices $v_2 = x_2$ and $v_3 = x_3$.

Layer decomposition. Using the level function $\ell$ from Definition 2.4.2, we obtain

$$V_0 = \{x_1, x_2, x_3, x_4\} = \{v_1, v_2, v_3, v_4\},$$

$$V_1 = \{v_5, v_6\}, \qquad V_2 = \{v_7\}, \qquad V = V_0 \,\dot\cup\, V_1 \,\dot\cup\, V_2.$$

Therefore,

$$\mathbb{H}^{\langle 2\rangle} = (V, E)$$

is a hierarchical SuperHyperGraph of height 2 on $V_0$ in the sense of Definition 2.4.2.

A recursive hypergraph is a hypergraph-like structure in which an edge may contain ordinary vertices and also recursively constructed set-valued objects, thereby allowing nested incidence up to a prescribed recursion depth [140, 141, 142, 143].

Definition 2.4.5 (Depth-$k$ powerset universe). [140, 143] Let $S$ be a nonempty set and let $k \in \mathbb{N} \cup \{0\}$. Define a sequence of sets $(S_i)_{i \geq 0}$ recursively by

$$S_0 := S, \qquad S_i := \mathcal{P}\Big(\bigcup_{j=0}^{i-1} S_j\Big) \quad (i \geq 1).$$

Set

$$U_{S,k} := \bigcup_{i=0}^{k} S_i.$$

The depth-$k$ powerset universe over $S$ is

$$2_{S,k} \ := \ \mathcal{P}(U_{S,k}).$$

Definition 2.4.6 ($k$-recursive hypergraph). [140, 143] Let $V$ be a finite nonempty set and let $k \in \mathbb{N} \cup \{0\}$. A $k$-recursive hypergraph is a pair

$$H = (V, E)$$

such that

$$E \subseteq 2_{V,k} \setminus \{\emptyset\},$$

where $2_{V,k}$ is the depth-$k$ powerset universe from Definition 2.4.5 (with $S = V$).

In particular, when $k = 0$, one has $U_{V,0} = V$ and hence

$$2_{V,0} = \mathcal{P}(V),$$

so $E \subseteq \mathcal{P}(V) \setminus \{\emptyset\}$, i.e., $H$ reduces to an ordinary hypergraph.

We now combine the hierarchical supervertex mechanism (via iterated powersets) with recursive superhyperedges (via bounded-depth recursive incidence).

Definition 2.4.7 ($(n, k)$-recursive SuperHyperGraph). Let $V_0$ be a finite nonempty base set, and let $n, k \in \mathbb{N} \cup \{0\}$. An $(n, k)$-recursive SuperHyperGraph over $V_0$ is a pair

$$\mathrm{RSHG}^{(n,k)} = (V, E)$$

satisfying:

(R1) (Hierarchical supervertex set of height $n$) $V$ is a finite nonempty set such that

$$V \subseteq \mathcal{U}_n(V_0),$$

where $\mathcal{U}_n(V_0)$ is the hierarchical universe from Definition 2.4.1 (with $r = n$).

(R2) (Downward-closure coherence) If $X \in V$ and $X \notin V_0$, then

$$X \subseteq V.$$

(R3) (Recursive superhyperedge family of depth $k$) $E$ is a finite set satisfying

$$E \subseteq 2_{V,k} \setminus \{\emptyset\},$$

where $2_{V,k}$ is the depth-$k$ powerset universe constructed from $S = V$ as in Definition 2.4.5.

Remark 2.4.8 (Limiting cases). The definition above is compatible with the standard models in the expected limiting cases:

(1) If $k = 0$, then $2_{V,0} = \mathcal{P}(V)$, and $\mathrm{RSHG}^{(n,0)}$ is exactly a hierarchical SuperHyperGraph of height $n$.

(2) If $n = 0$, then $V \subseteq V_0$, and $\mathrm{RSHG}^{(0,k)}$ becomes a $k$-recursive hypergraph on the (possibly proper) vertex set $V$.

Example 2.4.9 ($(2, 1)$-recursive SuperHyperGraph). Let the finite base set be

$$V_0 = \{x_1, x_2, x_3\}.$$

We take $n = 2$ and $k = 1$. Recall that $\mathcal{U}_2(V_0) = \big(V_0\big) \cup \big(\mathcal{P}(V_0) \setminus \{\emptyset\}\big) \cup \big(\mathcal{P}^2(V_0) \setminus \{\emptyset\}\big)$.

Step 1: Choose a coherent hierarchical supervertex set $V \subseteq \mathcal{U}_2(V_0)$. Define the following supervertices:

$$v_1 := x_1, \qquad v_2 := x_2, \qquad v_3 := x_3 \quad \text{(level 0)},$$
$$v_4 := \{x_1, x_2\}, \qquad v_5 := \{x_2, x_3\} \quad \text{(level 1)},$$
$$v_6 := \{\{x_1, x_2\}, \{x_3\}\} \quad \text{(level 2)}.$$

Set

$$V := \{v_1, v_2, v_3, v_4, v_5, v_6\} \subseteq \mathcal{U}_2(V_0).$$

This $V$ satisfies the downward-closure coherence condition (R2) because

$$v_4 = \{x_1, x_2\} \subseteq V, \qquad v_5 = \{x_2, x_3\} \subseteq V, \qquad v_6 = \{\{x_1, x_2\}, \{x_3\}\} = \{v_4, v_3\} \subseteq V.$$

Step 2: Construct the depth-1 powerset universe over $V$. For $k = 1$, Definition 2.4.5 gives

$$S_0 := V, \qquad S_1 := \mathcal{P}(S_0) = \mathcal{P}(V), \qquad U_{V,1} = S_0 \cup S_1 = V \cup \mathcal{P}(V),$$

and hence

$$2_{V,1} = \mathcal{P}(U_{V,1}) = \mathcal{P}\big(V \cup \mathcal{P}(V)\big).$$

Therefore a 1-recursive superhyperedge may contain both supervertices $v \in V$ and (sub)sets of supervertices $X \subseteq V$ as elements.

Step 3: Define a recursive superhyperedge family $E \subseteq 2_{V,1} \setminus \{\emptyset\}$. Let

$$e_1 := \{\, v_6,\ \{v_4, v_5\}\,\}, \qquad e_2 := \{\, v_4,\ \{v_1, v_2\},\ v_5\,\}.$$

Then $e_1, e_2 \subseteq V \cup \mathcal{P}(V) = U_{V,1}$, hence $e_1, e_2 \in 2_{V,1} \setminus \{\emptyset\}$. Set

$$E := \{e_1, e_2\}.$$

Intuitively, $e_1$ contains the level-2 supervertex $v_6$ together with the nested pair $\{v_4, v_5\}$, while $e_2$ mixes a level-1 supervertex $v_4$, a nested subset $\{v_1, v_2\}$ (which corresponds to $\{x_1, x_2\}$), and another level-1 supervertex $v_5$.

Conclusion. The pair

$$\mathrm{RSHG}^{(2,1)} = (V, E)$$

satisfies:

- $V \subseteq \mathcal{U}_2(V_0)$ (R1),
- downward-closure coherence (R2), and
- $E \subseteq 2_{V,1} \setminus \{\emptyset\}$ (R3).

Hence $\mathrm{RSHG}^{(2,1)}$ is a concrete $(2,1)$-recursive SuperHyperGraph over $V_0$.

If we set $k = 0$, then edges must be ordinary subsets of $V$ and the nested edge objects $\{v_4, v_5\}$ and $\{v_1, v_2\}$ would no longer be admissible as elements of a superedge. This illustrates how the recursion depth $k$ enlarges the admissible edge domain.

## 2.5 Uncertain Graphs

Since the introduction of fuzzy sets by Zadeh [47], a broad family of graph-based models has been developed to represent uncertainty in relational data. In this section we present a unified set-theoretic framework for several widely used uncertain graph formalisms, including fuzzy graphs, intuitionistic fuzzy graphs, neutrosophic graphs, hesitant fuzzy graphs, quadripartitioned neutrosophic graphs, and single-valued pentapartitioned neutrosophic graphs. For survey-style overviews and additional variants, see, e.g., [144, 77].

### 2.5.1 Crisp graphs

Fuzzy and related graph models are typically built upon an underlying crisp graph.

Definition 2.5.1 (Crisp graph). (cf. [144]) A crisp graph is a finite simple undirected graph $G = (V, E)$, where $V$ is a finite nonempty set of vertices and $E \subseteq \binom{V}{2} = \{\{u, v\} \subseteq V \mid u \neq v\}$ is a set of (unordered) edges. If $E = \emptyset$, then $G$ is called edgeless.

### 2.5.2 A unified framework

We now describe several uncertain graph models by attaching suitable membership information to vertices and edges. Throughout, let $G = (V, E)$ be a crisp graph.

Definition 2.5.2 (Unified framework for uncertain graphs). (cf. [77]) An uncertain graph model on $G = (V, E)$ consists of a vertex labeling $\sigma : V \to \mathcal{D}_V$ and an edge labeling $\mu : E \to \mathcal{D}_E$, where the codomains $\mathcal{D}_V, \mathcal{D}_E$ encode the chosen uncertainty formalism, together with constraints that relate $\mu(\{u, v\})$ to $\sigma(u)$ and $\sigma(v)$.

Definition 2.5.3 (Fuzzy graph). (cf. [145, 78, 37]) A fuzzy graph on $G = (V, E)$ is a pair of maps

$$\sigma : V \to [0, 1], \qquad \mu : E \to [0, 1],$$

such that for every edge $\{u, v\} \in E$,

$$\mu(\{u, v\}) \;\leq\; \min\{\sigma(u), \sigma(v)\}.$$

Definition 2.5.4 (Intuitionistic fuzzy graph). (cf. [146, 147]) An intuitionistic fuzzy graph (IFG) on $G = (V, E)$ consists of vertex maps

$$\sigma_\mu, \sigma_\nu : V \to [0, 1] \quad \text{with} \quad \sigma_\mu(v) + \sigma_\nu(v) \leq 1 \;\; (v \in V),$$

and edge maps

$$\mu_\mu, \mu_\nu : E \to [0, 1] \quad \text{with} \quad \mu_\mu(e) + \mu_\nu(e) \leq 1 \;\; (e \in E),$$

satisfying, for every $e = \{u, v\} \in E$,

$$\mu_\mu(e) \leq \min\{\sigma_\mu(u), \sigma_\mu(v)\}, \qquad \mu_\nu(e) \geq \max\{\sigma_\nu(u), \sigma_\nu(v)\}.$$

Definition 2.5.5 (Neutrosophic graph). (cf. [148, 149]) A (single-valued) neutrosophic graph on $G = (V, E)$ is given by

$$\sigma : V \to [0, 1]^3, \qquad \mu : E \to [0, 1]^3,$$

where $\sigma(v) = (\sigma_T(v), \sigma_I(v), \sigma_F(v))$ and $\mu(e) = (\mu_T(e), \mu_I(e), \mu_F(e))$. Typically one assumes $0 \leq \sigma_T, \sigma_I, \sigma_F \leq 1$ (and similarly for $\mu$), without requiring the sum to equal 1. A standard compatibility condition is that for each $e = \{u, v\} \in E$,

$$\mu_T(e) \leq \min\{\sigma_T(u), \sigma_T(v)\}, \qquad \mu_I(e) \leq \min\{\sigma_I(u), \sigma_I(v)\}, \qquad \mu_F(e) \geq \max\{\sigma_F(u), \sigma_F(v)\}.$$

Definition 2.5.6 (Hesitant fuzzy graph). (cf. [150, 151]) A hesitant fuzzy graph on $G = (V, E)$ consists of mappings

$$\sigma : V \to \mathcal{P}_{\text{fin}}([0, 1]) \setminus \{\emptyset\}, \qquad \mu : E \to \mathcal{P}_{\text{fin}}([0, 1]) \setminus \{\emptyset\},$$

assigning to each vertex/edge a nonempty finite set of possible membership degrees. (Additional constraints relating $\mu(\{u, v\})$ to $\sigma(u), \sigma(v)$ may be imposed depending on the chosen aggregation rule.)

Definition 2.5.7 (Quadripartitioned neutrosophic graph). (cf. [152, 153]) A quadripartitioned neutrosophic graph (QNG) on $G = (V, E)$ is specified by

$$\sigma : V \to [0, 1]^4, \qquad \mu : E \to [0, 1]^4,$$

where $\sigma(v) = (\sigma_1(v), \sigma_2(v), \sigma_3(v), \sigma_4(v))$ and $\mu(e) = (\mu_1(e), \mu_2(e), \mu_3(e), \mu_4(e))$, often with the boundedness constraint $\sum_{i=1}^{4} \sigma_i(v) \leq 4$ (and similarly for edges). A typical edge–vertex compatibility requirement is that for each $e = \{u, v\} \in E$,

$$\mu_1(e) \leq \min\{\sigma_1(u), \sigma_1(v)\}, \quad \mu_2(e) \leq \min\{\sigma_2(u), \sigma_2(v)\},$$
$$\mu_3(e) \leq \min\{\sigma_3(u), \sigma_3(v)\}, \quad \mu_4(e) \leq \min\{\sigma_4(u), \sigma_4(v)\}.$$

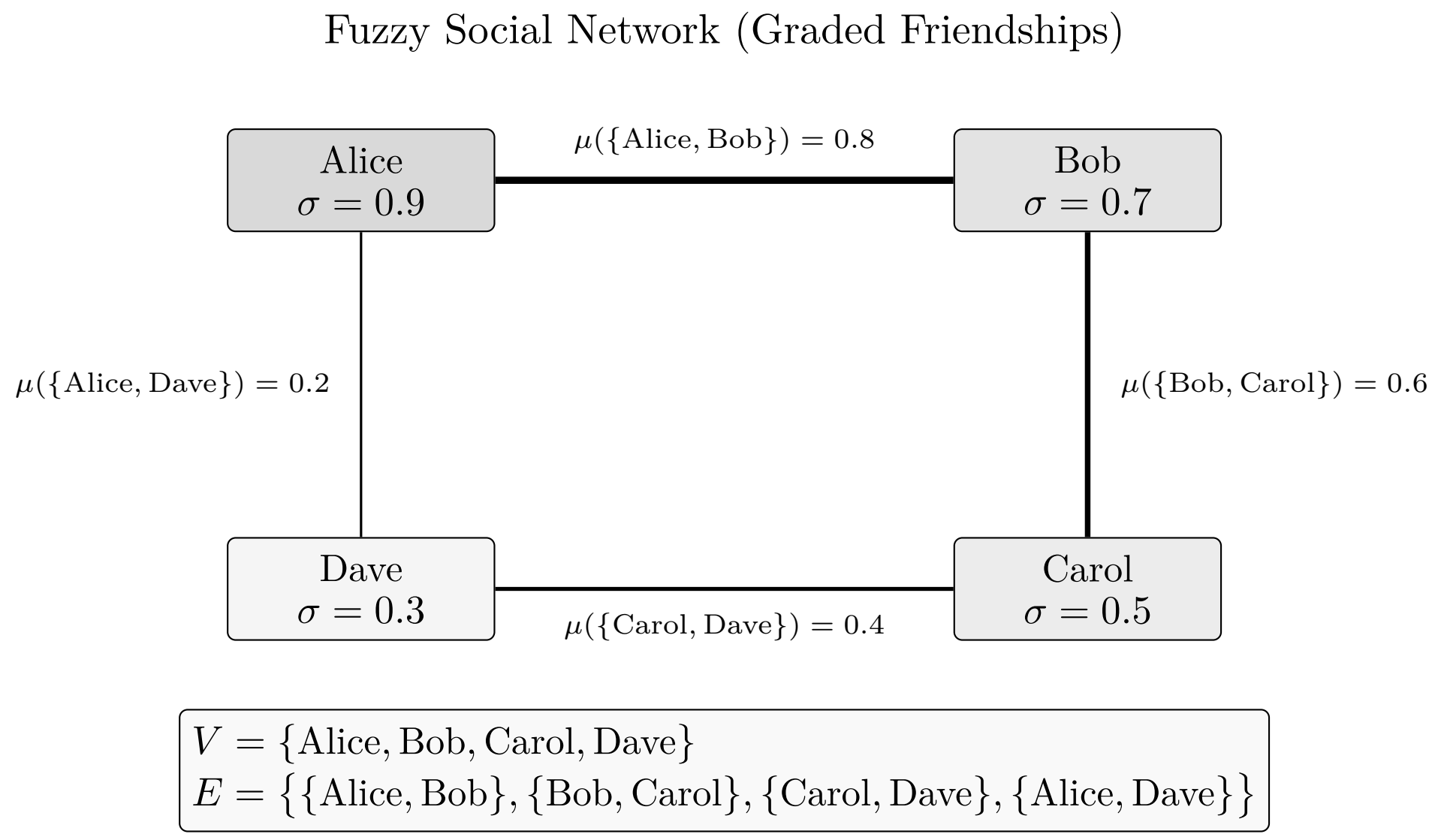

Figure 2.2: A fuzzy graph representation of a social network with graded vertex memberships $\sigma$ and edge memberships $\mu$ (Example 2.5.9).

Definition 2.5.8 (Single-valued pentapartitioned neutrosophic graph). (cf. [154, 155, 156]) A single-valued pentapartitioned neutrosophic graph on $G = (V, E)$ is given by

$$\sigma : V \to [0,1]^5, \qquad \mu : E \to [0,1]^5,$$

where $\sigma(v) = (\sigma_1(v), \ldots, \sigma_5(v))$ and $\mu(e) = (\mu_1(e), \ldots, \mu_5(e))$, often with $\sum_{i=1}^{5} \sigma_i(v) \le 5$ (and similarly for edges). A standard compatibility condition is that for each $e = \{u, v\} \in E$ and each $k = 1, \ldots, 5$,

$$\mu_k(e) \le \min\{\sigma_k(u), \sigma_k(v)\}.$$

We next give illustrative examples showing how uncertain graphs capture graded and multi-source uncertainty in applications (cf. [157, 158]).

Example 2.5.9 (Fuzzy graph: social network with graded friendship strengths). Consider a social network with users

$$V = \{\text{Alice}, \text{Bob}, \text{Carol}, \text{Dave}\},$$

and edges representing friendships, e.g.,

$$E = \big\{\{\text{Alice}, \text{Bob}\}, \{\text{Bob}, \text{Carol}\}, \{\text{Carol}, \text{Dave}\}, \{\text{Alice}, \text{Dave}\}\big\}.$$

Define vertex memberships (activity/influence levels) by

$$\sigma(\text{Alice}) = 0.9, \quad \sigma(\text{Bob}) = 0.7, \quad \sigma(\text{Carol}) = 0.5, \quad \sigma(\text{Dave}) = 0.3,$$

and edge memberships (friendship strengths) by

$$\mu(\{\text{Alice}, \text{Bob}\}) = 0.8, \quad \mu(\{\text{Bob}, \text{Carol}\}) = 0.6, \quad \mu(\{\text{Carol}, \text{Dave}\}) = 0.4, \quad \mu(\{\text{Alice}, \text{Dave}\}) = 0.2.$$

Then $\mu(\{u, v\}) \le \min\{\sigma(u), \sigma(v)\}$ holds for all edges, so $(\sigma, \mu)$ defines a fuzzy graph. Such a model supports graded analysis of social relations, e.g., for recommendation or community detection (cf. [159, 160]). For reference, an overview schematic is provided in Fig. 2.2.

Example 2.5.10 (Neutrosophic graph: disease-transmission network under uncertain evidence). In epidemiology, infection states and transmission events may be partially observed or mutually inconsistent. Let

$$V = \{\text{Patient1}, \text{Patient2}, \text{Patient3}, \text{Patient4}\},$$

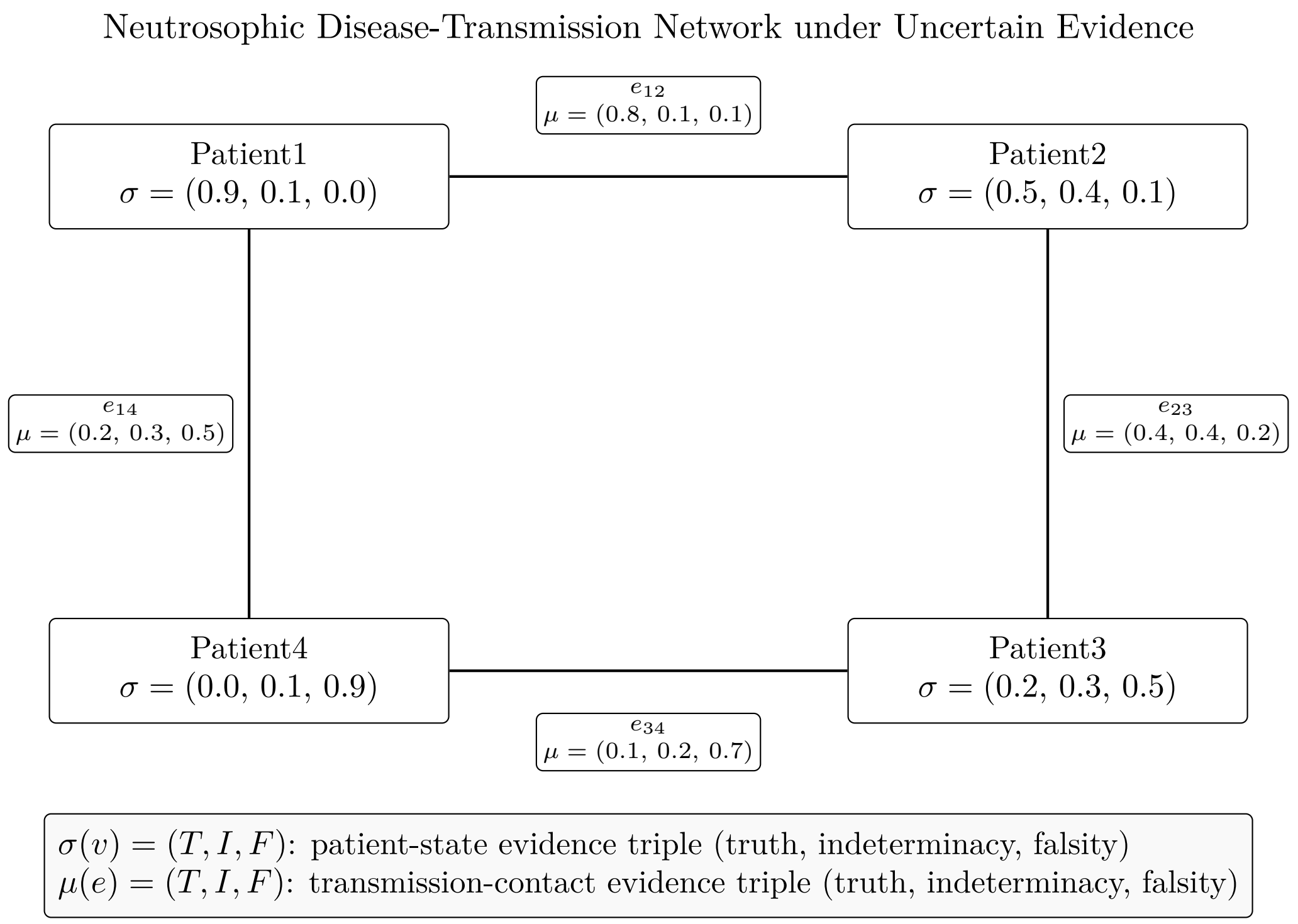

Figure 2.3: A neutrosophic graph model for a disease-transmission network with uncertain and possibly inconsistent evidence (Example 2.5.10).

and let $E$ encode potential transmission contacts. Assign to each patient a neutrosophic triple $\sigma(v) = (\sigma_T(v), \sigma_I(v), \sigma_F(v))$ and to each contact an edge triple $\mu(e) = (\mu_T(e), \mu_I(e), \mu_F(e))$, for example:

$$\sigma(\text{Patient1}) = (0.9, 0.1, 0.0), \quad \sigma(\text{Patient2}) = (0.5, 0.4, 0.1),$$
$$\sigma(\text{Patient3}) = (0.2, 0.3, 0.5), \quad \sigma(\text{Patient4}) = (0.0, 0.1, 0.9),$$

and

$$\mu(\{\text{Patient1}, \text{Patient2}\}) = (0.8, 0.1, 0.1), \quad \mu(\{\text{Patient2}, \text{Patient3}\}) = (0.4, 0.4, 0.2),$$
$$\mu(\{\text{Patient3}, \text{Patient4}\}) = (0.1, 0.2, 0.7), \quad \mu(\{\text{Patient1}, \text{Patient4}\}) = (0.2, 0.3, 0.5).$$

This representation separates supportive evidence (truth), ambiguity (indeterminacy), and counterevidence (falsity), aiding risk assessment and resource allocation under uncertainty (cf. [161]). For reference, an overview schematic is provided in Fig. 2.3.

Proposition 2.5.11. Neutrosophic graphs generalize fuzzy graphs.

Proof. Given a fuzzy graph $(\sigma, \mu)$ on $G = (V, E)$, define a neutrosophic labeling by

$$\sigma'(v) := (\sigma(v),\, 0,\, 1 - \sigma(v)), \qquad \mu'(e) := (\mu(e),\, 0,\, 1 - \mu(e)).$$

Then $\sigma', \mu'$ take values in $[0,1]^3$, and the edge compatibility $\mu(e) \leq \min\{\sigma(u), \sigma(v)\}$ implies $\mu'_T(e) \leq \min\{\sigma'_T(u), \sigma'_T(v)\}$ and $\mu'_F(e) \geq \max\{\sigma'_F(u), \sigma'_F(v)\}$. Hence $(\sigma', \mu')$ is a neutrosophic graph whose truth components recover the original fuzzy graph. □

A plithogenic graph is a graph-theoretic model derived from the notion of a plithogenic set, in which objects are described by (possibly multi-valued) attributes together with degrees of appurtenance and degrees of contradiction. This framework is often presented as a unifying generalization encompassing, as special cases, fuzzy-type and neutrosophic-type graph structures [97].

Definition 2.5.12 (Plithogenic graph). [62, 162] Let $G = (V, E)$ be a (crisp) graph, where $V$ is the vertex set and $E \subseteq V \times V$ is the edge set. A plithogenic graph over $G$ is a pair

$$PG = (PM, PN),$$

where:

1. Plithogenic vertex structure. $PM = (M, \ell, M_\ell, adf, aCf)$, with
   - $M \subseteq V$ (the set of vertices under consideration);
   - $\ell$ a chosen vertex attribute;
   - $M_\ell$ the set (range) of possible values of $\ell$;
   - $adf : M \times M_\ell \to [0,1]^s$ the degree of appurtenance function (DAF) for vertices;
   - $aCf : M_\ell \times M_\ell \to [0,1]^t$ the degree of contradiction function (DCF) for vertex-attribute values.
2. Plithogenic edge structure. $PN = (N, m, N_m, bdf, bCf)$, with
   - $N \subseteq E$ (the set of edges under consideration);
   - $m$ a chosen edge attribute;
   - $N_m$ the set (range) of possible values of $m$;
   - $bdf : N \times N_m \to [0,1]^s$ the degree of appurtenance function (DAF) for edges;
   - $bCf : N_m \times N_m \to [0,1]^t$ the degree of contradiction function (DCF) for edge-attribute values.

The pair $PG = (PM, PN)$ is required to satisfy the following conditions:

1. Edge appurtenance constraint. For all $(x, a), (y, b) \in M \times M_\ell$, if $xy \in N$ denotes the edge between $x$ and $y$ (in the underlying crisp graph), then

$$bdf\big(xy, (a, b)\big) \;\leq\; \min\{\, adf(x, a),\; adf(y, b) \,\},$$

   where $(a, b) \in N_m$ denotes the corresponding edge-attribute value induced from the end-point attribute values.

2. Contradiction constraint. For all $(a, b), (c, d) \in N_m \times N_m$,

$$bCf\big((a, b), (c, d)\big) \;\leq\; \min\{\, aCf(a, c),\; aCf(b, d) \,\}.$$

3. Reflexivity and symmetry of contradiction functions. For all admissible attribute values,

$$\begin{array}{llll} aCf(a, a) = 0, & \forall\, a \in M_\ell, & aCf(a, b) = aCf(b, a), & \forall\, a, b \in M_\ell, \\ bCf(\alpha, \alpha) = 0, & \forall\, \alpha \in N_m, & bCf(\alpha, \beta) = bCf(\beta, \alpha), & \forall\, \alpha, \beta \in N_m. \end{array}$$

Example 2.5.13 (Plithogenic graph: collaboration network with role-based uncertainty). Let $G = (V, E)$ be an undirected crisp graph, where

$$V = \{\text{Alice}, \text{Bob}, \text{Carol}, \text{Dave}\},$$

and

$$E = \big\{\{\text{Alice}, \text{Bob}\}, \{\text{Alice}, \text{Carol}\}, \{\text{Bob}, \text{Carol}\}, \{\text{Carol}, \text{Dave}\}\big\}.$$

We construct a plithogenic graph $PG = (PM, PN)$ on $G$ as follows (for simplicity, we take scalar-valued DAF/DCF, i.e. $s = t = 1$).

1) Plithogenic vertex structure $PM = (M, \ell, M_\ell, adf, aCf)$.

- $M := V$.

- $\ell$: professional role level.
- $M_\ell := \{\mathsf{Sr}, \mathsf{Md}, \mathsf{Jr}\}$ (Senior, Mid, Junior).
- The vertex DAF $adf : M \times M_\ell \to [0, 1]$ is given by:

$$\begin{aligned} adf(\text{Alice}, \mathsf{Sr}) &= 0.85, & adf(\text{Alice}, \mathsf{Md}) &= 0.15, & adf(\text{Alice}, \mathsf{Jr}) &= 0.00, \\ adf(\text{Bob}, \mathsf{Sr}) &= 0.40, & adf(\text{Bob}, \mathsf{Md}) &= 0.50, & adf(\text{Bob}, \mathsf{Jr}) &= 0.10, \\ adf(\text{Carol}, \mathsf{Sr}) &= 0.10, & adf(\text{Carol}, \mathsf{Md}) &= 0.70, & adf(\text{Carol}, \mathsf{Jr}) &= 0.20, \\ adf(\text{Dave}, \mathsf{Sr}) &= 0.05, & adf(\text{Dave}, \mathsf{Md}) &= 0.20, & adf(\text{Dave}, \mathsf{Jr}) &= 0.75. \end{aligned}$$

- The vertex contradiction function $aCf : M_\ell \times M_\ell \to [0, 1]$ is defined by

$$aCf(\mathsf{Sr}, \mathsf{Sr}) = aCf(\mathsf{Md}, \mathsf{Md}) = aCf(\mathsf{Jr}, \mathsf{Jr}) = 0,$$

$$aCf(\mathsf{Sr}, \mathsf{Md}) = 0.3, \qquad aCf(\mathsf{Md}, \mathsf{Jr}) = 0.4, \qquad aCf(\mathsf{Sr}, \mathsf{Jr}) = 0.8,$$

  and $aCf$ is symmetric.

2) Plithogenic edge structure $PN = (N, m, N_m, bdf, bCf)$.

- $N := E$.
- $m$: ordered endpoint role-pair attribute (using a fixed endpoint order, e.g. alphabetical order).
- $N_m := M_\ell \times M_\ell$.
- For each edge $e \in N$, assign an edge reliability factor $\rho_e \in [0, 1]$:

$$\rho_{\{\text{Alice,Bob}\}} = 0.90, \quad \rho_{\{\text{Alice,Carol}\}} = 0.60, \quad \rho_{\{\text{Bob,Carol}\}} = 0.80, \quad \rho_{\{\text{Carol,Dave}\}} = 0.70.$$

  Define $bdf : N \times N_m \to [0, 1]$ by

$$bdf\big(e, (a, b)\big) := \rho_e \cdot \min\{adf(x, a), adf(y, b)\},$$

  where $e = \{x, y\} \in N$ and $(a, b) \in N_m$.

- Define $bCf : N_m \times N_m \to [0, 1]$ by

$$bCf\big((a, b), (c, d)\big) := \min\{aCf(a, c),\ aCf(b, d)\}.$$

3) Verification of the plithogenic constraints.

1. For every $e = \{x, y\} \in N$ and $(a, b) \in N_m$,

$$bdf\big(e, (a, b)\big) = \rho_e \min\{adf(x, a), adf(y, b)\} \le \min\{adf(x, a), adf(y, b)\},$$

   since $0 \le \rho_e \le 1$. Hence the edge appurtenance constraint holds.

2. For all $(a, b), (c, d) \in N_m$,

$$bCf\big((a, b), (c, d)\big) = \min\{aCf(a, c), aCf(b, d)\} \le \min\{aCf(a, c), aCf(b, d)\},$$

   so the contradiction constraint holds (indeed, with equality).

3. By construction, $aCf$ is reflexive on the diagonal and symmetric. The same properties are inherited by $bCf$, because

$$bCf(\alpha, \alpha) = 0, \qquad bCf(\alpha, \beta) = bCf(\beta, \alpha).$$

Therefore, $PG = (PM, PN)$ is a well-defined plithogenic graph.

For instance, using the dominant role values

$$\text{Alice} \sim \mathsf{Sr}, \quad \text{Bob} \sim \mathsf{Md}, \quad \text{Carol} \sim \mathsf{Md}, \quad \text{Dave} \sim \mathsf{Jr},$$

we obtain the following representative edge appurtenance values:

$$\begin{aligned} bdf\big(\{\text{Alice}, \text{Bob}\}, (\mathsf{Sr}, \mathsf{Md})\big) &= 0.90 \cdot \min\{0.85, 0.50\} = 0.45, \\ bdf\big(\{\text{Alice}, \text{Carol}\}, (\mathsf{Sr}, \mathsf{Md})\big) &= 0.60 \cdot \min\{0.85, 0.70\} = 0.42, \\ bdf\big(\{\text{Bob}, \text{Carol}\}, (\mathsf{Md}, \mathsf{Md})\big) &= 0.80 \cdot \min\{0.50, 0.70\} = 0.40, \\ bdf\big(\{\text{Carol}, \text{Dave}\}, (\mathsf{Md}, \mathsf{Jr})\big) &= 0.70 \cdot \min\{0.70, 0.75\} = 0.49. \end{aligned}$$

This model is suitable for collaboration analysis under attribute uncertainty, where vertices and edges are evaluated relative to multiple potentially contradictory attribute values.
2) Plithogenic edge structure $PN = (N, m, N_m, bdf, bCf)$.

- $N := E$.
- $m$: ordered endpoint role-pair attribute (using a fixed endpoint order, e.g. alphabetical order).
- $N_m := M_\ell \times M_\ell$.
- For each edge $e \in N$, assign an edge reliability factor $\rho_e \in [0, 1]$:

  $$\rho_{\{\text{Alice},\text{Bob}\}} = 0.90, \quad \rho_{\{\text{Alice},\text{Carol}\}} = 0.60, \quad \rho_{\{\text{Bob},\text{Carol}\}} = 0.80, \quad \rho_{\{\text{Carol},\text{Dave}\}} = 0.70.$$

  Define $bdf : N \times N_m \to [0, 1]$ by

  $$bdf\big(e, (a, b)\big) \;:=\; \rho_e \cdot \min\{adf(x, a), adf(y, b)\},$$

  where $e = \{x, y\} \in N$ and $(a, b) \in N_m$.
- Define $bCf : N_m \times N_m \to [0, 1]$ by

  $$bCf\big((a, b), (c, d)\big) \;:=\; \min\{aCf(a, c),\ aCf(b, d)\}.$$

3) Verification of the plithogenic constraints.

1. For every $e = \{x, y\} \in N$ and $(a, b) \in N_m$,

   $$bdf\big(e, (a, b)\big) = \rho_e \min\{adf(x, a), adf(y, b)\} \le \min\{adf(x, a), adf(y, b)\},$$

   since $0 \le \rho_e \le 1$. Hence the edge appurtenance constraint holds.
2. For all $(a, b), (c, d) \in N_m$,

   $$bCf\big((a, b), (c, d)\big) = \min\{aCf(a, c), aCf(b, d)\} \le \min\{aCf(a, c), aCf(b, d)\},$$

   so the contradiction constraint holds (indeed, with equality).
3. By construction, $aCf$ is reflexive on the diagonal and symmetric. The same properties are inherited by $bCf$, because

   $$bCf(\alpha, \alpha) = 0, \qquad bCf(\alpha, \beta) = bCf(\beta, \alpha).$$

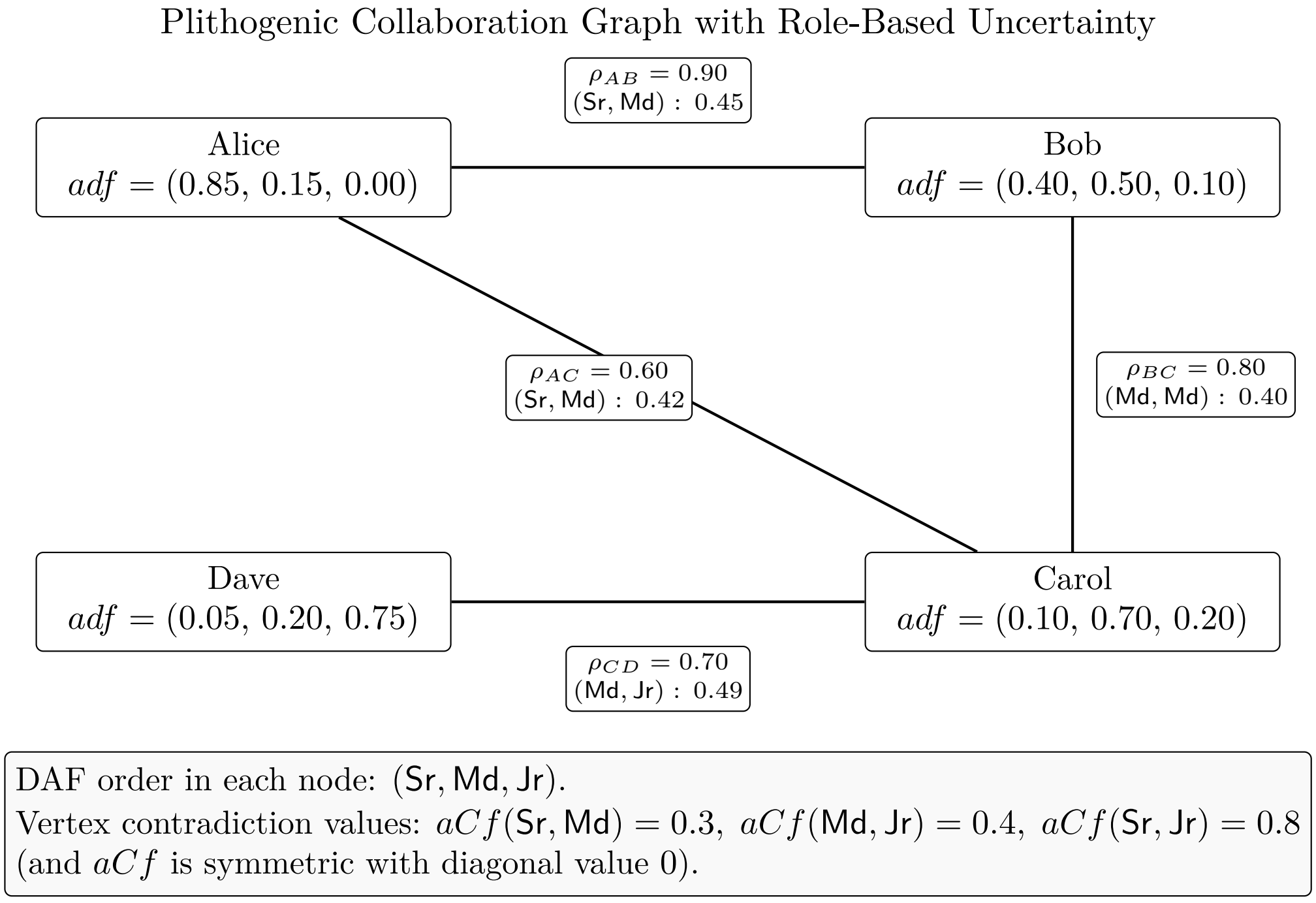

Figure 2.4: A concrete plithogenic graph example. Each vertex carries a degree-of-appurtenance vector (DAF) for the role-level attribute $(\mathsf{Sr}, \mathsf{Md}, \mathsf{Jr})$, and each edge is equipped with a plithogenic edge appurtenance induced from endpoint attributes.

Therefore, $PG = (PM, PN)$ is a well-defined plithogenic graph.

For instance, using the dominant role values

$$\text{Alice} \sim \mathsf{Sr}, \quad \text{Bob} \sim \mathsf{Md}, \quad \text{Carol} \sim \mathsf{Md}, \quad \text{Dave} \sim \mathsf{Jr},$$

we obtain the following representative edge appurtenance values:

$$\begin{aligned}
bdf\big(\{\text{Alice}, \text{Bob}\}, (\mathsf{Sr}, \mathsf{Md})\big) &= 0.90 \cdot \min\{0.85, 0.50\} = 0.45, \\
bdf\big(\{\text{Alice}, \text{Carol}\}, (\mathsf{Sr}, \mathsf{Md})\big) &= 0.60 \cdot \min\{0.85, 0.70\} = 0.42, \\
bdf\big(\{\text{Bob}, \text{Carol}\}, (\mathsf{Md}, \mathsf{Md})\big) &= 0.80 \cdot \min\{0.50, 0.70\} = 0.40, \\
bdf\big(\{\text{Carol}, \text{Dave}\}, (\mathsf{Md}, \mathsf{Jr})\big) &= 0.70 \cdot \min\{0.70, 0.75\} = 0.49.
\end{aligned}$$

This model is suitable for collaboration analysis under attribute uncertainty, where vertices and edges are evaluated relative to multiple potentially contradictory attribute values. For reference, an overview schematic is provided in Fig. 2.4.

An uncertain set assigns to each element a degree drawn from a prescribed uncertainty model. In this way, it serves as a common umbrella encompassing fuzzy, intuitionistic fuzzy, neutrosophic, plithogenic, and related degree-based formalisms [64, 163]. An uncertain graph is then a graph whose vertices and/or edges carry such model-dependent degrees, so that fuzzy graphs, intuitionistic fuzzy graphs, and neutrosophic graphs arise as particular instances.

Definition 2.5.14 (Uncertain model). [64] Let $U$ denote the class of all uncertain models. An element $M \in U$ is specified by

- a nonempty degree-domain $\mathrm{Dom}(M) \subseteq [0, 1]^k$ of admissible degree tuples for some fixed integer $k \geq 1$; and
- additional algebraic and/or geometric admissibility conditions on $\mathrm{Dom}(M)$ (depending on the model), such as $\mu + \nu \leq 1$ in the intuitionistic fuzzy setting, or $0 \leq T + I + F \leq 3$ in the neutrosophic setting.

Typical examples include:

- Fuzzy model: $\mathrm{Dom}(M) = [0, 1]$;
- Intuitionistic fuzzy model: $\mathrm{Dom}(M) = \{(\mu, \nu) \in [0, 1]^2 : \ \mu + \nu \leq 1\}$;
- Neutrosophic model: $\mathrm{Dom}(M) = \{(T, I, F) \in [0, 1]^3 : \ 0 \leq T + I + F \leq 3\}$;
- Plithogenic model, together with many further extensions.

Definition 2.5.15 (Uncertain set (U-set)). [64] Let $X$ be a nonempty universe, and fix an uncertain model $M$ with degree-domain $\mathrm{Dom}(M) \subseteq [0, 1]^k$. An uncertain set of type $M$ (briefly, a U-set) on $X$ is a pair

$$\mathcal{U} = (X, \mu_M),$$

where

$$\mu_M : X \longrightarrow \mathrm{Dom}(M)$$

is the uncertainty-degree (membership) map. For each $x \in X$, the value $\mu_M(x) \in \mathrm{Dom}(M)$ records the model-prescribed degree tuple describing the membership status of $x$ in $\mathcal{U}$.

We now introduce the corresponding graph-theoretic notion.

Definition 2.5.16 (Uncertain graph). Let $G = (V, E)$ be a finite, undirected, simple graph (i.e., loopless and without multiple edges), and let $M$ be an uncertain model with degree-domain $\mathrm{Dom}(M)$. An uncertain graph of type $M$ is a triple

$$\mathcal{G}_M = (V, E, \mu_M),$$

where

$$\mu_M : V \cup E \longrightarrow \mathrm{Dom}(M)$$

assigns an uncertainty degree in $\mathrm{Dom}(M)$ to each vertex $v \in V$ and each edge $e \in E$. Depending on the chosen model $M$, one may additionally impose compatibility constraints linking edge degrees to their endpoint degrees (for example, upper bounds on $\mu_M(e)$ in terms of $\mu_M(u)$ and $\mu_M(v)$ when $e = \{u, v\}$), but such conditions are model-specific and are not built into the present general definition.

For convenience, Table 2.1 lists representative families of uncertainty graphs, grouped by the dimension $k$ of the degree-domain $\mathrm{Dom}(M) \subseteq [0, 1]^k$.

Definition 2.5.17 (Uncertain hypergraph). [178] Let $H = (V, E)$ be a hypergraph and let $M$ be an uncertain model with degree-domain $\mathrm{Dom}(M)$. An uncertain hypergraph of type $M$ is a triple

$$\mathcal{H}_M = (V, E, \mu_M),$$

where

$$\mu_M : V \cup E \longrightarrow \mathrm{Dom}(M)$$

assigns an uncertainty degree to each vertex $v \in V$ and each hyperedge $e \in E$. As in the graph case, any relations between vertex-degrees and hyperedge-degrees (e.g., bounds of $\mu_M(e)$ in terms of $\mu_M(v)$ for $v \in e$) are governed by the selected model $M$ and its constraints.

Definition 2.5.18 (Uncertain $n$-SuperHyperGraph). [178] Let $V_0$ be a finite base set and let $n \in \mathbb{N}_0$. Assume that an $n$-SuperHyperGraph on $V_0$ is given by

$$\mathrm{SHG}^{(n)} = (V_n, E),$$

where

$$\emptyset \neq V_n \subseteq \mathcal{P}^n(V_0) \quad \text{and} \quad \emptyset \neq E \subseteq \mathcal{P}(V_n) \setminus \{\emptyset\},$$

so that each $n$-superedge $e \in E$ is a nonempty subset of the $n$-supervertex set $V_n$.

Table 2.1: Representative uncertainty-graph families, organized by the dimension $k$ of the degree-domain $\mathrm{Dom}(M) \subseteq [0,1]^k$.

| $k$ | Representative uncertainty-graph type(s) $\mathcal{G}_M = (V, E, \mu_M)$ with $\mu_M : V \cup E \to \mathrm{Dom}(M) \subseteq [0,1]^k$ |
|---|---|
| 1 | Fuzzy graph; $N$-graph; shadowed-graph variants |
| 2 | Intuitionistic fuzzy graph[164]; vague graph[165]; bipolar fuzzy graph[79]; intuitionistic evidence graph; variable fuzzy graph; paraconsistent fuzzy graph; bifuzzy graph[166, 167] |
| 3 | Neutrosophic graph[168][(a)]; hesitant fuzzy graph[86]; tripolar fuzzy graph; three-way fuzzy graph; picture fuzzy graph[169, 82]; spherical fuzzy graph[60]; inconsistent intuitionistic fuzzy graph; ternary fuzzy / neutrosophic-fuzzy graph; neutrosophic vague graph |
| 4 | Quadripartitioned neutrosophic graph[170, 171]; double-valued neutrosophic graph[172]; dual hesitant fuzzy graph[173]; ambiguous graph[(b)]; local-neutrosophic graph; support-neutrosophic graph; turiyam neutrosophic graph[174][(c)] |
| 5 | Pentapartitioned neutrosophic graph[175]; triple-valued neutrosophic graph |
| 6 | Hexapartitioned neutrosophic graph; quadruple-valued neutrosophic graph |
| 7 | Heptapartitioned neutrosophic graph[176]; quintuple-valued neutrosophic graph |
| 8 | Octapartitioned neutrosophic graph |
| 9 | Nonapartitioned neutrosophic graph |
| $n$ | $n$-refined fuzzy graph; multi-valued (fuzzy) graphs; multi-fuzzy graphs[177] |
| $2n$ | $n$-refined intuitionistic fuzzy graph; multi-intuitionistic fuzzy graphs |
| $3n$ | $n$-refined neutrosophic graph; multi-neutrosophic graphs |

(a) Neutrosophic graphs are often treated as broad degree-based frameworks that can specialize to many related graph models under additional constraints.
(b) Ambiguous-graph models are commonly presented as subclasses of certain quadripartitioned and also double-valued neutrosophic graph models.
(c) Turiyam neutrosophic graphs are reported as subclasses of certain quadripartitioned neutrosophic graph models.

Fix an uncertain model $M$ with degree-domain $\mathrm{Dom}(M) \subseteq [0,1]^k$. An uncertain $n$-SuperHyperGraph of type $M$ is a triple

$$\mathcal{S}_M^{(n)} = (V_n, E, \mu_M),$$

where

$$\mu_M : V_n \cup E \longrightarrow \mathrm{Dom}(M)$$

assigns an uncertainty degree in $\mathrm{Dom}(M)$ to each $n$-supervertex $v \in V_n$ and each $n$-superedge $e \in E$.

Any additional relations between superedge-degrees and the degrees of the supervertices they contain (for example, model-specific bounds or aggregation rules) are determined by the chosen uncertain model $M$ and are not fixed at the level of this general definition.

For $n = 0$ and $V_0 = V_n$, this notion reduces to an uncertain hypergraph of type $M$.

## 2.6 Directed SuperHyperGraph

Directed SuperHyperGraphs extend SuperHyperGraphs in the same spirit that directed graphs extend undirected graphs: hyperedges carry an orientation from a source part to a target part[178]. We begin with directed hypergraphs and then introduce their powerset-based $n$-level analogues.

Definition 2.6.1 (Directed hypergraph). [179, 180, 181] A directed hypergraph is a pair $H = (V, E)$, where $V$ is a (typically finite) set of vertices and $E$ is a set of directed hyperedges. Each directed hyperedge $e \in E$ is an ordered pair

$$e = (T_e, H_e),$$

where $T_e \subseteq V$ is the tail (source set) and $H_e \subseteq V$ is the head (target set), and one assumes

$$T_e \neq \emptyset, \qquad H_e \neq \emptyset.$$

When $|T_e| = |H_e| = 1$ for every $e \in E$, the structure reduces to an ordinary directed graph.

Example 2.6.2 (Directed hypergraph in a CI/CD build pipeline). A practical instance of a directed hypergraph arises in modern build systems and CI/CD pipelines. Let $V$ be the set of artifacts (source files, generated files, object files, test binaries, container images, deployment manifests, etc.). Each pipeline step that consumes multiple artifacts and produces multiple artifacts is naturally modeled as a directed hyperedge $e = (T_e, H_e)$, where:

- $T_e$ (tail) is the set of required inputs (e.g., $\{\text{main.c}, \text{util.c}, \text{libssl.a}\}$);
- $H_e$ (head) is the set of outputs (e.g., $\{\text{app.o}, \text{app}\}$ or $\{\text{image:app:v1}, \text{sbom.json}\}$).

This representation captures many-to-many transformations that ordinary digraphs cannot express without losing structure.

Definition 2.6.3 (Underlying digraph). Let $H = (V, E)$ be a directed hypergraph. Its underlying digraph is the digraph $U(H) = (V, A)$ with

$$A := \big\{(u, v) \in V \times V \;\big|\; \exists\, e = (T_e, H_e) \in E : \; u \in T_e, \; v \in H_e\big\}.$$

Thus $u \to v$ is an arc in $U(H)$ whenever some directed hyperedge sends flow from a tail containing $u$ to a head containing $v$.

Example 2.6.4 (Underlying digraph as a dependency graph for incremental builds). Continuing Example 2.6.2, the underlying digraph $U(H)$ provides a convenient coarse dependency view used by incremental compilation and cache invalidation. An arc $u \to v$ appears in $U(H)$ whenever there exists a pipeline step (hyperedge) whose tail contains $u$ and head contains $v$. For instance, if a build step uses util.c and produces app, then util.c $\to$ app is an arc in $U(H)$. Thus, when a file $u$ changes, one can quickly find all potentially affected outputs $v$ by reachability in $U(H)$, even if the original process is fundamentally hyperedge-based.

Definition 2.6.5 (Directed $n$-SuperHyperGraph). (cf.[178, 17]) Let $S$ be a nonempty base set and let $n \geq 0$. Consider the iterated powersets $\mathcal{P}^n(S)$ defined by $\mathcal{P}^0(S) = S$ and $\mathcal{P}^{k+1}(S) = \mathcal{P}(\mathcal{P}^k(S))$. A directed $n$-SuperHyperGraph is a pair

$$\mathrm{DSuHG}^{(n)} = (V, \mathcal{E}),$$

where $V \subseteq \mathcal{P}^n(S)$ is the set of $n$-supervertices, and

$$\mathcal{E} \subseteq \big(\mathcal{P}(V) \setminus \{\emptyset\}\big) \times \big(\mathcal{P}(V) \setminus \{\emptyset\}\big)$$

is the set of directed $n$-superedges. Each $e \in \mathcal{E}$ is written as

$$e = (\mathrm{Tail}(e), \mathrm{Head}(e)),$$

where $\mathrm{Tail}(e)$ and $\mathrm{Head}(e)$ are nonempty subsets of $V$, interpreted as the source and target collections of $n$-supervertices, respectively.

Example 2.6.6 (Directed 2-SuperHyperGraph for hierarchical microservice call flows). A directed $n$-SuperHyperGraph is useful when the system is hierarchical and the units of interaction are nested collections. Consider a Kubernetes-based microservice platform: let $S$ be the set of pods (the base compute instances). Take $n = 2$. Then:

- a 1-supervertex is a service (a subset of pods implementing one API);
- a 2-supervertex is an application (a set of services, i.e., an element of $\mathcal{P}(\mathcal{P}(S)) = \mathcal{P}^2(S)$).

Let $V \subseteq \mathcal{P}^2(S)$ be the set of such applications. In an event-driven architecture, several applications may jointly emit events that trigger several downstream applications (fan-in/fan-out). Model this by a directed 2-superedge $e = (\mathrm{Tail}(e), \mathrm{Head}(e))$, where $\mathrm{Tail}(e) \subseteq V$ is the set of upstream applications (publishers) and $\mathrm{Head}(e) \subseteq V$ is the set of downstream applications (subscribers). This encodes inter-application flow while preserving the internal hierarchy (pods $\to$ services $\to$ applications) within the vertex domain. For reference, an overview schematic is provided in Fig. 2.5.

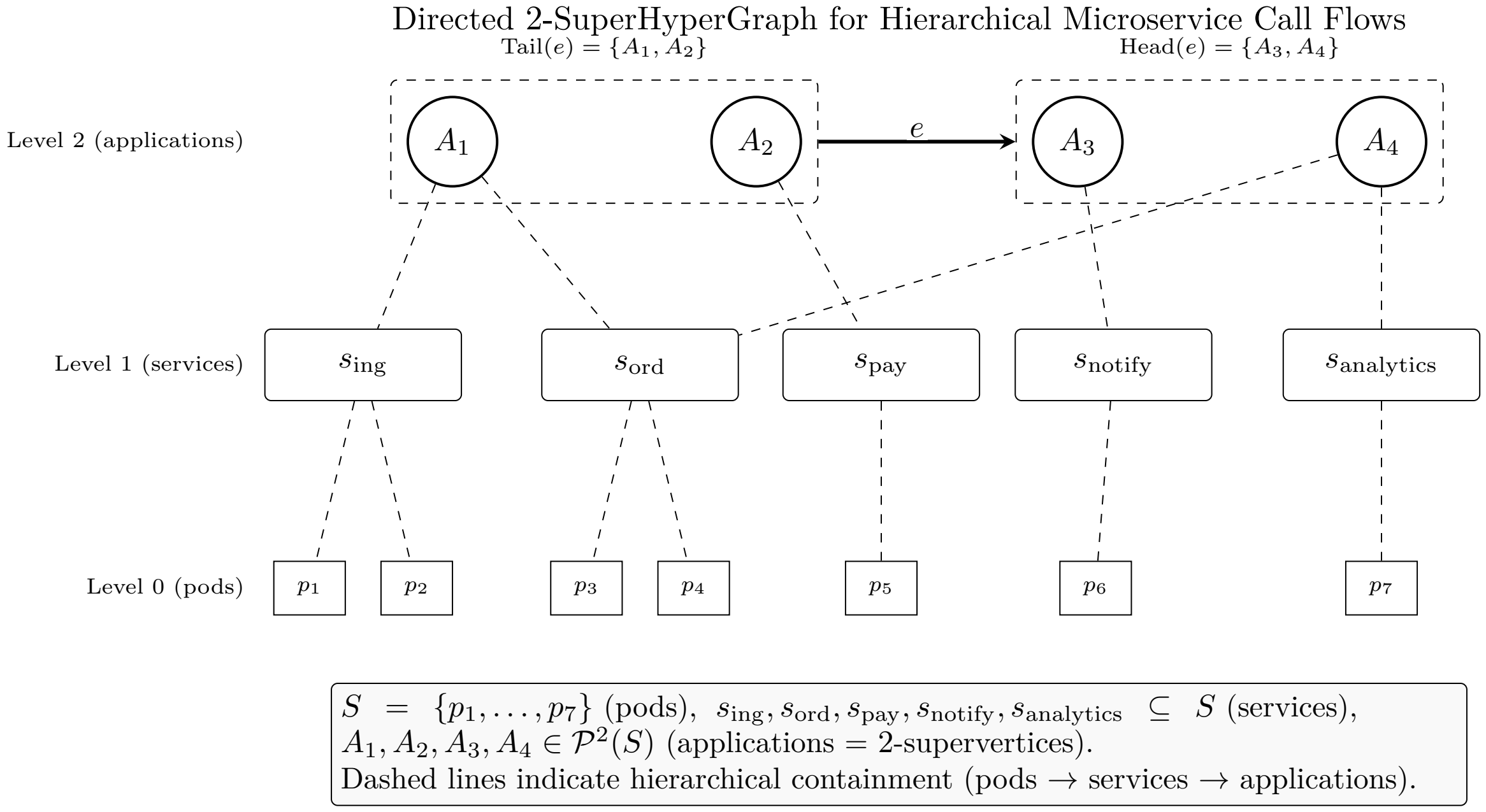

Figure 2.5: A directed 2-SuperHyperGraph for a Kubernetes-based microservice platform. The directed 2-superedge $e = (\text{Tail}(e), \text{Head}(e))$ models a fan-in/fan-out event flow between sets of applications, while preserving the internal hierarchy from pods to services to applications (Example 2.6.6).

# 3 Review and Results: SuperHypergraph Neural Network

In this section, we explore the SuperHyperGraph Neural Network.

## 3.1 HGNN: Hypergraph Neural Networks

Hypergraph neural networks (HGNNs) extend the graph neural network paradigm from pairwise relations to higher-order relations encoded by hyperedges [182, 183, 184]. They have been actively studied in a variety of architectures and learning settings [185, 186, 187]. Related concepts include molecular hypergraph neural networks [188, 189], heterogeneous hypergraph neural networks [190, 191, 182], hypergraph convolutional neural networks[192, 193, 194, 195], and fuzzy hypergraph neural networks[196]. We recall a standard spectral formulation below.

Definition 3.1.1 (Hypergraph neural network (HGNN)). [184] Let $G = (V, E, w)$ be a finite weighted hypergraph, where $V = \{v_1, \dots, v_n\}$ is the vertex set, $E = \{e_1, \dots, e_m\} \subseteq \mathcal{P}(V) \setminus \{\emptyset\}$ is the set of hyperedges, and $w : E \to (0, \infty)$ assigns a positive weight $w(e_j)$ to each hyperedge $e_j$. Let $W \in \mathbb{R}^{m\times m}$ be the diagonal hyperedge-weight matrix

$$W = \operatorname{diag}\big(w(e_1), \dots, w(e_m)\big).$$

Incidence matrix. The incidence matrix $H \in \{0, 1\}^{n\times m}$ is defined by

$$H_{ij} = \begin{cases} 1, & \text{if } v_i \in e_j, \\ 0, & \text{otherwise.} \end{cases}$$

Vertex and hyperedge degrees. The (weighted) degree of a vertex $v_i$ and the (cardinality) degree of a hyperedge $e_j$ are

$$d(v_i) = \sum_{j=1}^{m} H_{ij}\, w(e_j), \qquad \delta(e_j) = \sum_{i=1}^{n} H_{ij} = |e_j|.$$

Let $D_V \in \mathbb{R}^{n\times n}$ and $D_E \in \mathbb{R}^{m\times m}$ be the corresponding diagonal matrices:

$$(D_V)_{ii} = d(v_i), \qquad (D_E)_{jj} = \delta(e_j).$$

Normalized hypergraph Laplacian. (cf. [197, 198]) Define the normalized hypergraph adjacency (or propagation) operator

$$S := D_V^{-1/2}\, H\, W\, D_E^{-1}\, H^{\top}\, D_V^{-1/2} \in \mathbb{R}^{n\times n},$$

and the associated Laplacian

$$\Delta := I - S,$$

where $I$ is the $n \times n$ identity matrix.

HGNN layer (spectral-style propagation). (cf. [199, 200]) Let $X \in \mathbb{R}^{n\times d}$ be a node-feature matrix (row $i$ is the feature vector of $v_i$). An HGNN layer produces an output $Y \in \mathbb{R}^{n\times c}$ of the form

$$Y = \sigma\big(S\, X\, \Theta\big),$$

where $\Theta \in \mathbb{R}^{d\times c}$ is a learnable weight matrix and $\sigma$ is an element-wise nonlinearity (e.g., ReLU).

Multi-layer model and node classification. Given $X^{(0)}$ as the input features, a depth-$L$ HGNN is defined recursively by

$$X^{(l+1)} = \sigma\big(S\, X^{(l)}\Theta^{(l)}\big), \qquad l = 0, 1, \dots, L-1,$$

with trainable $\Theta^{(l)}$. For node classification with $c$ classes, the final predictions are obtained by applying row-wise softmax:

$$\widehat{Y} = \mathrm{softmax}\big(X^{(L)}\big) \in \mathbb{R}^{n\times c}.$$

Proposition 3.1.2. HGNNs strictly generalize (undirected) GNNs in the sense that every simple graph can be represented as a 2-uniform hypergraph, and the HGNN propagation operator reduces to the usual normalized adjacency used in GCN-style message passing.

Proof. Let $G_{\mathrm{gr}} = (V, E_{\mathrm{gr}})$ be a simple undirected graph with $V = \{v_1, \dots, v_n\}$. Form the 2-uniform hypergraph $G = (V, E, w)$ by taking

$$E := \big\{\{u, v\} \mid \{u, v\} \in E_{\mathrm{gr}}\big\}, \qquad w(e) \equiv 1.$$

Then $\delta(e) = 2$ for all $e \in E$, so $D_E = 2I_m$. Moreover, the matrix $HH^\top$ has diagonal entries equal to the (graph) degrees and off-diagonal entry $(i, j)$ equal to 1 precisely when $\{v_i, v_j\} \in E_{\mathrm{gr}}$. Consequently,

$$HD_E^{-1}H^\top = \tfrac{1}{2}\, HH^\top = \tfrac{1}{2}\,(D + A),$$

where $A$ is the adjacency matrix of $G_{\mathrm{gr}}$ and $D$ is the diagonal degree matrix. Since $D_V = D$ (because each edge has weight 1), the HGNN propagation operator becomes

$$S = D^{-1/2}\left(\tfrac{1}{2}(D + A)\right) D^{-1/2} = \tfrac{1}{2}I + \tfrac{1}{2}\, D^{-1/2}AD^{-1/2}.$$

Thus an HGNN layer $Y = \sigma(SX\Theta)$ reduces (up to the constant self-term $\frac{1}{2}I$) to the standard GCN-style normalized adjacency propagation on the original graph. Therefore, graph-based message passing is recovered as a special case of hypergraph message passing. □

## 3.2 1-SuperHypergraph Neural Network

In this subsection we formulate a 1-SuperHypergraph Neural Network (1-SHGNN), i.e., a neural message-passing scheme defined directly on a 1-SuperHypergraph (SuperHyperGraph). Conceptually, the model extends hypergraph neural networks by allowing the "nodes" of the learning architecture to be supervertices (nonempty subsets of a base vertex set) and the "hyperedges" to be superedges (nonempty sets of supervertices). We emphasize that the present development is purely theoretical.

Definition 3.2.1 (1-SuperHypergraph). Let $V_0$ be a finite nonempty set (the base vertices). A 1-SuperHypergraph over $V_0$ is an ordered pair $\mathsf{H} = (\mathcal{V}, \mathcal{E})$ such that

$$\mathcal{V} \subseteq \mathcal{P}^*(V_0), \qquad \mathcal{E} \subseteq \mathcal{P}^*(\mathcal{V}),$$

where $\mathcal{P}^*(X) = \mathcal{P}(X) \setminus \{\emptyset\}$. Elements of $\mathcal{V}$ are called supervertices and elements of $\mathcal{E}$ are called superedges.

Definition 3.2.2 (1-SuperHypergraph Neural Network (1-SHGNN)). Let $\mathsf{H} = (\mathcal{V}, \mathcal{E})$ be a 1-SuperHypergraph over $V_0 = \{x_1, \dots, x_n\}$, with $\mathcal{V} = \{S_1, \dots, S_p\}$ and $\mathcal{E} = \{\varepsilon_1, \dots, \varepsilon_q\}$. Assume base-vertex features

$$X \in \mathbb{R}^{n\times d} \qquad (\text{row } i \text{ corresponds to } x_i \in V_0).$$

A 1-SHGNN layer is defined by the following canonical operators.

(1) Base–supervertex membership. Define the membership matrix $B \in \{0,1\}^{n\times p}$ by

$$B_{ia} = \begin{cases} 1, & x_i \in S_a, \\ 0, & \text{otherwise.} \end{cases}$$

Let $D_S \in \mathbb{R}^{p\times p}$ and $D_0 \in \mathbb{R}^{n\times n}$ be diagonal matrices

$$(D_S)_{aa} = \sum_{i=1}^{n} B_{ia} = |S_a|, \qquad (D_0)_{ii} = \sum_{a=1}^{p} B_{ia}.$$

(When some diagonal entries vanish, one may replace $D_S^{-1}$ or $D_0^{-1}$ below by a pseudoinverse.)

(2) Lifting (base $\to$ supervertex features). Define supervertex features by normalized pooling:

$$\widetilde{X} \;:=\; D_S^{-1} B^\top X \in \mathbb{R}^{p\times d}.$$

Equivalently, the $a$-th row of $\widetilde{X}$ is the mean of the feature vectors of base vertices in $S_a$.

(3) Superedge incidence and normalized propagation on supervertices. Define the incidence matrix $H \in \{0,1\}^{p\times q}$ of $(\mathcal{V}, \mathcal{E})$ by

$$H_{aj} = \begin{cases} 1, & S_a \in \varepsilon_j, \\ 0, & \text{otherwise.} \end{cases}$$

Let $w : \mathcal{E} \to (0,\infty)$ be a superedge-weight function and set $W = \operatorname{diag}(w(\varepsilon_1), \dots, w(\varepsilon_q))$. Define diagonal degree matrices

$$(D_{\mathcal{V}})_{aa} = \sum_{j=1}^{q} H_{aj}\, w(\varepsilon_j), \qquad (D_{\mathcal{E}})_{jj} = \sum_{a=1}^{p} H_{aj} = |\varepsilon_j|.$$

Then the normalized propagation operator on supervertices is

$$S \;:=\; D_{\mathcal{V}}^{-1/2}\, H\, W\, D_{\mathcal{E}}^{-1}\, H^\top\, D_{\mathcal{V}}^{-1/2} \in \mathbb{R}^{p\times p}.$$

(4) Convolution on a 1-SuperHypergraph. Given learnable parameters $\Theta \in \mathbb{R}^{d\times c}$ and an activation $\sigma$, define the supervertex-level output

$$\widetilde{Y} \;=\; \sigma\Big(S\, \widetilde{X}\, \Theta\Big) \in \mathbb{R}^{p\times c}.$$

(5) Optional projection (supervertex $\to$ base). If a base-vertex output is required, project back via

$$Y \;:=\; D_0^{-1} B\, \widetilde{Y} \in \mathbb{R}^{n\times c}.$$

A multi-layer 1-SHGNN is obtained by iterating Step (4) with layer-wise parameters $\Theta^{(\ell)}$ and embeddings $\widetilde{X}^{(\ell)}$, optionally followed by Step (5) at the end.

Remark 3.2.3 (Relation to the "expanded hypergraph"). From a 1-SuperHypergraph $\mathsf{H} = (\mathcal{V}, \mathcal{E})$ over $V_0$, one may form a hypergraph on base vertices by expanding each superedge:

$$\varepsilon_j \in \mathcal{E} \quad\longmapsto\quad e'_j := \bigcup_{S\in\varepsilon_j} S \;\subseteq\; V_0.$$

This yields an "expanded" hypergraph $H' = (V_0, E')$ with $E' = \{e'_1, \dots, e'_q\}$. The 1-SHGNN above, however, performs propagation at the supervertex level (on $\mathcal{V}$), and only optionally maps embeddings back to $V_0$.

Example 3.2.4 (1-SHGNN on overlapping groups of base vertices). We present a concrete one-layer computation of a 1-SuperHypergraph Neural Network (Definition 3.2.2) on a small overlapping-group structure.

Let the base vertex set be

$$V_0 = \{x_1, x_2, x_3, x_4\},$$

and define the 1-supervertices

$$S_1 = \{x_1, x_2\}, \qquad S_2 = \{x_2, x_3\}, \qquad S_3 = \{x_3, x_4\}.$$

Hence

$$\mathcal{V} = \{S_1, S_2, S_3\}.$$

Let the superedge family be

$$\mathcal{E} = \{\varepsilon_1, \varepsilon_2\}, \qquad \varepsilon_1 = \{S_1, S_2\}, \quad \varepsilon_2 = \{S_2, S_3\}.$$

Thus $\mathsf{H} = (\mathcal{V}, \mathcal{E})$ is a 1-SuperHypergraph.

Assume the base-feature matrix $X \in \mathbb{R}^{4\times 2}$ is

$$X = \begin{bmatrix} 1 & 0 \\ 2 & 1 \\ 0 & 2 \\ 1 & 1 \end{bmatrix},$$

where row $i$ corresponds to $x_i$.

Step 1: Base–supervertex membership matrix $B$. By definition,

$$B_{ia} = 1 \iff x_i \in S_a,$$

so

$$B = \begin{bmatrix} 1 & 0 & 0 \\ 1 & 1 & 0 \\ 0 & 1 & 1 \\ 0 & 0 & 1 \end{bmatrix}.$$

The diagonal matrices $D_S$ and $D_0$ are

$$D_S = \operatorname{diag}(|S_1|, |S_2|, |S_3|) = \operatorname{diag}(2, 2, 2),$$

$$D_0 = \operatorname{diag}\left(\sum_{a=1}^{3} B_{1a}, \ldots, \sum_{a=1}^{3} B_{4a}\right) = \operatorname{diag}(1, 2, 2, 1).$$

Step 2: Lifting (base $\to$ supervertex features). The lifted feature matrix is

$$\widetilde{X} = D_S^{-1} B^\top X.$$

First,

$$B^\top X = \begin{bmatrix} 3 & 1 \\ 2 & 3 \\ 1 & 3 \end{bmatrix},$$

hence

$$\widetilde{X} = \frac{1}{2}\begin{bmatrix} 3 & 1 \\ 2 & 3 \\ 1 & 3 \end{bmatrix} = \begin{bmatrix} \frac{3}{2} & \frac{1}{2} \\ 1 & \frac{3}{2} \\ \frac{1}{2} & \frac{3}{2} \end{bmatrix}.$$

Thus each row of $\widetilde{X}$ is the mean feature vector of the corresponding supervertex.

Step 3: Superedge incidence and normalized propagation. The incidence matrix $H \in \{0,1\}^{3\times 2}$ of $(\mathcal{V}, \mathcal{E})$ is

$$H = \begin{bmatrix} 1 & 0 \\ 1 & 1 \\ 0 & 1 \end{bmatrix},$$

since $S_1, S_2 \in \varepsilon_1$ and $S_2, S_3 \in \varepsilon_2$.

Take unit superedge weights:

$$w(\varepsilon_1) = w(\varepsilon_2) = 1, \qquad W = \mathrm{diag}(1,1).$$

Then

$$D_{\mathcal{E}} = \mathrm{diag}(|\varepsilon_1|, |\varepsilon_2|) = \mathrm{diag}(2,2),$$

and

$$D_{\mathcal{V}} = \mathrm{diag}(1,2,1),$$

because $S_1$ belongs to one superedge, $S_2$ belongs to two, and $S_3$ belongs to one.

Therefore, the normalized propagation operator is

$$S = D_{\mathcal{V}}^{-1/2} H W D_{\mathcal{E}}^{-1} H^{\top} D_{\mathcal{V}}^{-1/2} = \begin{bmatrix} \frac{1}{2} & \frac{1}{2\sqrt{2}} & 0 \\ \frac{1}{2\sqrt{2}} & \frac{1}{2} & \frac{1}{2\sqrt{2}} \\ 0 & \frac{1}{2\sqrt{2}} & \frac{1}{2} \end{bmatrix}.$$

Step 4: One 1-SHGNN convolution layer. Choose

$$\Theta = I_2 \qquad \text{and} \qquad \sigma = \mathrm{ReLU}.$$

Since all entries below are nonnegative, ReLU acts as the identity. Thus

$$\widetilde{Y} = \sigma(S\widetilde{X}\Theta) = S\widetilde{X}.$$

Let $a := \frac{1}{2\sqrt{2}}$. A direct computation gives

$$\widetilde{Y} = \begin{bmatrix} \frac{3}{4} + a & \frac{1}{4} + \frac{3}{2}a \\ \frac{1}{2} + 2a & \frac{3}{4} + 2a \\ \frac{1}{4} + a & \frac{3}{4} + \frac{3}{2}a \end{bmatrix} \approx \begin{bmatrix} 1.1036 & 0.7803 \\ 1.2071 & 1.4571 \\ 0.6036 & 1.2803 \end{bmatrix}.$$

Step 5: Optional projection back to base vertices. The base-level output is

$$Y = D_0^{-1} B \widetilde{Y}.$$

Since $x_2$ and $x_3$ belong to two supervertices, their outputs are averaged:

$$Y = \begin{bmatrix} \widetilde{Y}_{1\bullet} \\ \frac{1}{2}(\widetilde{Y}_{1\bullet} + \widetilde{Y}_{2\bullet}) \\ \frac{1}{2}(\widetilde{Y}_{2\bullet} + \widetilde{Y}_{3\bullet}) \\ \widetilde{Y}_{3\bullet} \end{bmatrix} = \begin{bmatrix} \frac{3}{4} + a & \frac{1}{4} + \frac{3}{2}a \\ \frac{5}{8} + \frac{3}{2}a & \frac{1}{2} + \frac{7}{4}a \\ \frac{3}{8} + \frac{3}{2}a & \frac{3}{4} + \frac{7}{4}a \\ \frac{1}{4} + a & \frac{3}{4} + \frac{3}{2}a \end{bmatrix}$$

and numerically

$$Y \approx \begin{bmatrix} 1.1036 & 0.7803 \\ 1.1553 & 1.1187 \\ 0.9053 & 1.3687 \\ 0.6036 & 1.2803 \end{bmatrix}.$$

This example illustrates the two key mechanisms of a 1-SHGNN:

1. lifting base features to overlapping supervertices via normalized pooling, and

2. propagation on the superhypergraph structure before optionally projecting back.

In particular, the overlap vertices $x_2$ and $x_3$ receive information from two neighboring supervertices, which is precisely the intended higher-order message-passing behavior.

Theorem 3.2.5 (HGNN as a special case). The hypergraph neural network (HGNN) is a special case of the 1-SHGNN in Definition 3.2.2. More precisely, assume that the supervertex set consists of all singletons,

$$\mathcal{V} = \{\{x_i\} \mid x_i \in V_0\}.$$

Then, after identifying each singleton $\{x_i\}$ with $x_i$, the 1-SHGNN propagation in Definition 3.2.2 coincides with the standard HGNN propagation on the hypergraph $(V_0, E')$ where $E'$ corresponds to $\mathcal{E}$.

Proof. If $\mathcal{V} = \{\{x_i\} \mid i = 1, \ldots, n\}$, then $p = n$ and the membership matrix is $B = I_n$. Hence $\widetilde{X} = D_S^{-1} B^\top X = X$ because $|\{x_i\}| = 1$ implies $D_S = I_n$. Moreover, each superedge $\varepsilon_j \subseteq \mathcal{V}$ corresponds to the hyperedge $e'_j = \{x_i \mid \{x_i\} \in \varepsilon_j\} \subseteq V_0$, and the incidence matrix $H$ becomes the usual incidence matrix of the hypergraph $(V_0, E')$. Therefore the normalized operator $S = D_{\mathcal{V}}^{-1/2} H W D_{\mathcal{E}}^{-1} H^\top D_{\mathcal{V}}^{-1/2}$ and the layer map $\widetilde{Y} = \sigma(SX\Theta)$ are exactly the HGNN layer. □

Corollary 3.2.6 (GCN-type reduction under pairwise hyperedges). Under the assumptions of Theorem 3.2.5, if every hyperedge has size 2 (i.e., $E'$ is the edge set of an ordinary graph on $V_0$), then the HGNN/1-SHGNN propagation reduces to a graph-convolution operator on that graph, hence yields a GCN-type layer (up to the chosen normalization convention).

Proof. When all hyperedges have cardinality 2, the hypergraph $(V_0, E')$ is just an ordinary graph. In this case, the matrix $H W D_{\mathcal{E}}^{-1} H^\top$ is a weighted combination of pairwise endpoint contributions, and after symmetric degree normalization by $D_{\mathcal{V}}^{-1/2}$ it becomes a normalized adjacency-style propagation operator. Thus $\sigma(SX\Theta)$ is a standard graph-convolution update (possibly with a residual/self-loop term, depending on the specific normalization and weighting choices). □

## 3.3 Algorithm for 1-SuperHypergraph Neural Network (1-SHGNN)

This section presents an explicit algorithmic realization of a 1-SuperHypergraph Neural Network (1-SHGNN) layer as defined in Definition 3.2.2. Unlike the previous SHGNN draft (which first formed an "expanded" hypergraph on the base vertices), the 1-SHGNN performs message passing on supervertices and only optionally projects the learned representations back to base vertices.

Theorem 3.3.1 (Correctness of Algorithm 1-layer 1-SHGNN). Let $\mathsf{H} = (\mathcal{V}, \mathcal{E})$ be a 1-SuperHypergraph over $V_0$, and let $B, H, D_S, D_0, D_{\mathcal{V}}, D_{\mathcal{E}}, W$ be constructed as in the algorithm above. Define the normalized propagation operator on supervertices by

$$S \;:=\; D_{\mathcal{V}}^{-1/2}\, H\, W\, D_{\mathcal{E}}^{-1}\, H^\top\, D_{\mathcal{V}}^{-1/2} \in \mathbb{R}^{p\times p}.$$

Then the algorithm outputs

$$\widetilde{X} = D_S^{-1} B^\top X \qquad \text{and} \qquad \widetilde{Y} = \sigma\big(S\,\widetilde{X}\,\Theta\big).$$

If the optional base projection is executed, it additionally outputs

$$Y = D_0^{-1} B\, \widetilde{Y}.$$

Proof. Step (C) computes $\widetilde{X} = D_S^{-1} B^\top X$ by definition.

For the convolution, the algorithm performs the following sequence:

$$Z = D_{\mathcal{V}}^{-1/2} \widetilde{X}, \quad U = H^\top Z, \quad U \leftarrow D_{\mathcal{E}}^{-1} W U, \quad R = HU, \quad R \leftarrow D_{\mathcal{V}}^{-1/2} R.$$

Substituting and simplifying gives

$$R = D_{\mathcal{V}}^{-1/2} H\, W\, D_{\mathcal{E}}^{-1} H^{\top} D_{\mathcal{V}}^{-1/2} \widetilde{X} = S\, \widetilde{X}.$$

Finally the algorithm returns $\widetilde{Y} = \sigma(R\Theta) = \sigma(S\widetilde{X}\Theta)$. If the optional step is executed, it returns $Y = D_0^{-1} B \widetilde{Y}$ as stated. □

Theorem 3.3.2 (Time complexity of one 1-SHGNN layer). Let $n = |V_0|$, $p = |\mathcal{V}|$, $q = |\mathcal{E}|$, input dimension $d$, output dimension $c$. Let

$$\operatorname{nnz}(B) = \big|\{(i,a) : x_i \in S_a\}\big| = \sum_{a=1}^{p} |S_a|, \qquad \operatorname{nnz}(H) = \big|\{(a,j) : S_a \in \varepsilon_j\}\big| = \sum_{j=1}^{q} |\varepsilon_j|.$$

Then the total time to compute one layer $\widetilde{Y}$ (and optionally $Y$) is

$$O(\operatorname{nnz}(B) \cdot d \; + \; \operatorname{nnz}(H) \cdot d \; + \; p \cdot d \cdot c \; + \; \mathbf{1}_{\text{proj}} \cdot \operatorname{nnz}(B) \cdot c)\,,$$

where $\mathbf{1}_{\text{proj}} = 1$ if the optional base projection is computed and 0 otherwise.

Proof. We bound each block of the algorithm:

- Constructing $B$: proportional to listing memberships $x_i \in S_a$, i.e., $O(\operatorname{nnz}(B))$. Computing $D_S, D_0$: column/row sums, $O(\operatorname{nnz}(B))$.
- Lifting $\widetilde{X} = D_S^{-1} B^{\top} X$: sparse–dense multiplication $B^{\top} X$ costs $O(\operatorname{nnz}(B) \cdot d)$, followed by diagonal scaling $D_S^{-1}$ costing $O(p \cdot d)$.
- Constructing $H$: $O(\operatorname{nnz}(H))$. Computing $D_{\mathcal{V}}, D_{\mathcal{E}}$: summations over nonzeros, $O(\operatorname{nnz}(H))$.
- Convolution: $Z = D_{\mathcal{V}}^{-1/2} \widetilde{X}$ costs $O(p \cdot d)$; $U = H^{\top} Z$ costs $O(\operatorname{nnz}(H) \cdot d)$; diagonal scalings by $W$ and $D_{\mathcal{E}}^{-1}$ cost $O(q \cdot d)$; $R = HU$ costs $O(\operatorname{nnz}(H) \cdot d)$; final scaling by $D_{\mathcal{V}}^{-1/2}$ costs $O(p \cdot d)$; dense multiplication $R\Theta$ costs $O(p \cdot d \cdot c)$; activation costs $O(p \cdot c)$.
- Optional projection $Y = D_0^{-1} B \widetilde{Y}$: sparse–dense multiplication $B\widetilde{Y}$ costs $O(\operatorname{nnz}(B) \cdot c)$, plus diagonal scaling $O(n \cdot c)$.

Collecting dominant terms yields the stated bound. □

Theorem 3.3.3 (Space complexity of one 1-SHGNN layer). With the notation of Theorem 3.3.2, the space usage (beyond input storage conventions) is

$$O(\operatorname{nnz}(B) + \operatorname{nnz}(H) \; + \; n \cdot d \; + \; p \cdot (d + c) \; + \; q \cdot d \; + \; d \cdot c \; + \; \mathbf{1}_{\text{proj}} \cdot n \cdot c)\,.$$

Proof. We store the sparse matrices $B$ and $H$, contributing $O(\operatorname{nnz}(B) + \operatorname{nnz}(H))$. We store the dense feature matrices and intermediates: $X \in \mathbb{R}^{n \times d}$, $\widetilde{X}, Z, R \in \mathbb{R}^{p \times d}$, $U \in \mathbb{R}^{q \times d}$, $\widetilde{Y} \in \mathbb{R}^{p \times c}$, and $\Theta \in \mathbb{R}^{d \times c}$, as well as diagonal degree/weight vectors of sizes $n, p, q$ (lower order). If projection is requested, we additionally store $Y \in \mathbb{R}^{n \times c}$. Summing these yields the stated bound. □

Theorem 3.3.4 (Near-linear behavior under sparsity). Assume $\operatorname{nnz}(B) = O(n)$ and $\operatorname{nnz}(H) = O(p)$, and that $d, c$ are treated as constants. Then a single 1-SHGNN layer runs in $O(n + p)$ time and uses $O(n + p)$ additional space (up to constant factors depending on $d, c$).

Proof. Under the assumptions, the dominant terms in Theorems 3.3.2 and 3.3.3 reduce to $O(n + p)$ when $d, c$ are constants. □

## 3.4 $n$-SuperHyperGraph Neural Network

A SuperHyperGraph can be extended to an $n$-SuperHyperGraph by allowing supervertices to be nested set-theoretic objects obtained from iterated powersets.

Definition 3.4.1 ((Recall) $n$-SuperHyperGraph). (cf. [17]) Let $V_0$ be a finite set of base vertices. For $k \geq 0$, define the iterated powersets by

$$\mathcal{P}^0(V_0) := V_0, \qquad \mathcal{P}^{k+1}(V_0) := \mathcal{P}\big(\mathcal{P}^k(V_0)\big),$$

and set the powerset tower up to level $n$ as

$$\mathcal{P}^{\leq n}(V_0) \ := \ \bigcup_{k=0}^{n} \mathcal{P}^k(V_0).$$

An $n$-SuperHyperGraph is an ordered pair $H = (V, E)$ such that

- $V \subseteq \mathcal{P}^{\leq n}(V_0)$ is a finite set of $n$-supervertices;
- $E \subseteq \mathcal{P}^*(V)$ is a finite set of $n$-superedges, where $\mathcal{P}^*(V) := \mathcal{P}(V) \setminus \{\emptyset\}$.

Thus, each superedge $\varepsilon \in E$ is a nonempty set of supervertices, i.e., it connects elements of $V$.

If one wishes to incorporate uncertainty (e.g., fuzziness) into the model, it is mathematically cleaner to label supervertices and/or superedges by uncertainty degrees (for instance, via membership functions) rather than identifying a supervertex itself with a fuzzy set.

Theorem 3.4.2 ([201]). An $n$-SuperHyperGraph generalizes a (classical) SuperHyperGraph.

Proof. A SuperHyperGraph corresponds to the case $n = 1$, where supervertices lie in $\mathcal{P}(V_0)$. Since $\mathcal{P}(V_0) \subseteq \mathcal{P}^{\leq n}(V_0)$ for every $n \geq 1$, any SuperHyperGraph is an $n$-SuperHyperGraph by the same vertex/edge specification. □

Corollary 3.4.3. An $n$-SuperHyperGraph generalizes both hypergraphs and classical graphs.

Proof. A hypergraph is obtained from a SuperHyperGraph by restricting supervertices to singletons of $V_0$, and a graph is obtained by additionally restricting hyperedges to size 2. Since SuperHyperGraphs are included as a special case of $n$-SuperHyperGraphs, the claim follows. □

Definition 3.4.4 (Expanded Hypergraph of an $n$-SuperHyperGraph). Let $H = (V, E)$ be an $n$-SuperHyperGraph over $V_0$. Define the base-expansion map Expand : $\mathcal{P}^{\leq n}(V_0) \to \mathcal{P}(V_0)$ recursively by

$$\mathrm{Expand}(x) = \{x\} \ (x \in V_0), \qquad \mathrm{Expand}(\emptyset) = \emptyset, \qquad \mathrm{Expand}(S) = \bigcup_{u \in S} \mathrm{Expand}(u) \ (S \notin V_0).$$

For each superedge $\varepsilon \in E$, define the corresponding (expanded) hyperedge on base vertices by

$$\xi(\varepsilon) \ := \ \bigcup_{v \in \varepsilon} \mathrm{Expand}(v) \ \in \ \mathcal{P}(V_0).$$

The Expanded Hypergraph of $H$ is the hypergraph

$$H^{\uparrow} \ = \ (V_0, E^{\uparrow}),$$

where $E^{\uparrow}$ is the indexed family $\big(\xi(\varepsilon)\big)_{\varepsilon \in E}$ (viewed as a multiset if duplicates occur).

Theorem 3.4.5. The expanded-hypergraph construction for $n$-SuperHyperGraphs extends the usual expanded-hypergraph construction for SuperHyperGraphs.

Proof. When $n = 1$, every supervertex $v \in V \subseteq \mathcal{P}^{\leq 1}(V_0)$ is either a base vertex $x \in V_0$ or a subset $S \subseteq V_0$. In this case, $\mathrm{Expand}(x) = \{x\}$ and $\mathrm{Expand}(S) = \bigcup_{u \in S}\{u\} = S$, so $\xi(\varepsilon) = \bigcup_{v \in \varepsilon} v$, which is precisely the standard expansion rule. For $n > 1$, the same recursion expands nested set-objects down to $V_0$, hence the construction is a genuine extension. □

We now define an $n$-SuperHyperGraph neural layer.

Definition 3.4.6 ($n$-SuperHyperGraph Neural Network ($n$-SHGNN) layer). Let $H = (V, E)$ be an $n$-SuperHyperGraph over $V_0$. Fix enumerations $V = \{S_1, \ldots, S_p\}$ and $E = \{\varepsilon_1, \ldots, \varepsilon_q\}$. Let $X \in \mathbb{R}^{|V_0| \times d}$ be the feature matrix on base vertices.

(1) Lift base features to supervertices. Define $B \in \{0, 1\}^{|V_0| \times p}$ by

$$B_{ia} := \begin{cases} 1, & \text{if } x_i \in \mathrm{Expand}(S_a), \\ 0, & \text{otherwise}, \end{cases} \qquad (x_i \in V_0).$$

Let $D_S := \mathrm{diag}(|\mathrm{Expand}(S_1)|, \ldots, |\mathrm{Expand}(S_p)|)$, and set

$$\widetilde{X} \;:=\; D_S^{-1} B^{\top} X \in \mathbb{R}^{p \times d},$$

using a pseudoinverse of $D_S$ if needed.

(2) Incidence on the supervertex layer. Define the incidence matrix $\mathbf{H} \in \{0, 1\}^{p \times q}$ by

$$\mathbf{H}_{aj} := \begin{cases} 1, & \text{if } S_a \in \varepsilon_j, \\ 0, & \text{otherwise}. \end{cases}$$

Let $W := \mathrm{diag}(w(\varepsilon_1), \ldots, w(\varepsilon_q))$ with $w(\varepsilon_j) > 0$, and define

$$(D_V)_{aa} := \sum_{j=1}^{q} \mathbf{H}_{aj}\, w(\varepsilon_j), \qquad (D_E)_{jj} := \sum_{a=1}^{p} \mathbf{H}_{aj} = |\varepsilon_j|.$$

(3) Convolution (supervertex output). Given learnable $\Theta \in \mathbb{R}^{d \times c}$ and an activation $\sigma$, define

$$\widetilde{Y} \;:=\; \sigma\Big(D_V^{-1/2}\, \mathbf{H}\, W\, D_E^{-1}\, \mathbf{H}^{\top}\, D_V^{-1/2}\, \widetilde{X}\, \Theta\Big) \in \mathbb{R}^{p \times c},$$

using pseudoinverses for $D_V$ and $D_E$ if necessary.

(4) Optional projection back to base vertices. Let $D_0 := \mathrm{diag}\big(\sum_{a=1}^{p} B_{1a}, \ldots, \sum_{a=1}^{p} B_{|V_0|\,a}\big)$. A base-level output can be defined by

$$Y \;:=\; D_0^{-1} B\, \widetilde{Y} \in \mathbb{R}^{|V_0| \times c},$$

again using a pseudoinverse if required.

Theorem 3.4.7. The SuperHyperGraph Neural Network (SHGNN) is a special case of the $n$-SHGNN when $n = 1$.

Proof. For $n = 1$, we have $V \subseteq \mathcal{P}^{\leq 1}(V_0)$, so every supervertex is either a base vertex or a subset of $V_0$. The recursion defining Expand yields $\mathrm{Expand}(x) = \{x\}$ for $x \in V_0$ and $\mathrm{Expand}(S) = S$ for $S \subseteq V_0$. Hence the lift matrix $B$ coincides with the usual base–supervertex membership matrix, and the incidence/degree operators on $(V, E)$ are unchanged. Therefore the layer formula in Definition 3.4.6 reduces exactly to the 1-SHGNN construction. □

As basic procedures for $n$-SuperHyperGraphs, we consider (i) constructing the expanded hypergraph on the base vertex set, and (ii) performing one $n$-SHGNN convolution layer on that expanded hypergraph.

Algorithm 2: Construction of the Expanded Hypergraph $H^{\uparrow}$

Input: An $n$-SuperHyperGraph $H = (V, E)$ over the base vertex set $V_0$.
Output: The expanded hypergraph $H^{\uparrow} = (V_0, E^{\uparrow})$.

```
Initialize E↑ ← [ ] as an empty list (to preserve multiplicities);
foreach superedge ε ∈ E do
    ξ ← Expand(ε) ; // ξ ⊆ V0
    Append ξ to E↑;
end
return H↑ = (V0, E↑);
```

**Algorithm 3:** One-layer $n$-SHGNN Convolution on the Expanded Hypergraph

**Input:**

- Base-vertex features $X \in \mathbb{R}^{N\times d}$ with $N := |V_0|$.
- Expanded hypergraph $H^{\uparrow} = (V_0, E^{\uparrow})$ with $E^{\uparrow} = [\xi_1, \ldots, \xi_m]$.
- Hyperedge weights $w_1, \ldots, w_m > 0$ and $W := \mathrm{diag}(w_1, \ldots, w_m)$.
- Learnable weights $\Theta \in \mathbb{R}^{d\times c}$.
- Activation function $\sigma$.

**Output:** Output features $Y \in \mathbb{R}^{N\times c}$.

Incidence matrix:;
Construct $\mathbf{H} \in \{0,1\}^{N\times m}$ by

$$\mathbf{H}_{ij} := \begin{cases} 1, & \text{if } v_i \in \xi_j, \\ 0, & \text{otherwise.} \end{cases}$$

;
Degree matrices:;
Compute

$$(D_V)_{ii} := \sum_{j=1}^{m} \mathbf{H}_{ij}\, w_j, \qquad (D_E)_{jj} := \sum_{i=1}^{N} \mathbf{H}_{ij} = |\xi_j|,$$

and set $D_V := \mathrm{diag}((D_V)_{11}, \ldots, (D_V)_{NN})$, $D_E := \mathrm{diag}((D_E)_{11}, \ldots, (D_E)_{mm})$;
Normalization:;
Compute $\widehat{\mathbf{H}} := D_V^{-1/2}\mathbf{H}WD_E^{-1}$;
Convolution:;
Compute

$$Y \leftarrow \sigma\Big(\widehat{\mathbf{H}}\,\mathbf{H}^{\top} D_V^{-1/2} X\,\Theta\Big) \;=\; \sigma\Big(D_V^{-1/2}\mathbf{H}WD_E^{-1}\mathbf{H}^{\top} D_V^{-1/2} X\,\Theta\Big).$$

;
return $Y$;

**Theorem 3.4.8** (Correctness). The $n$-SHGNN convolution algorithm outputs the matrix

$$Y = \sigma\Big(D_V^{-1/2}\mathbf{H}WD_E^{-1}\mathbf{H}^{\top} D_V^{-1/2} X\,\Theta\Big),$$

i.e., it implements exactly the convolution operator defined on the expanded hypergraph $H^{\uparrow}$.

*Proof.* By construction, $\mathbf{H}$ is the incidence matrix of $H^{\uparrow}$, and $D_V, D_E$ are computed from their defining formulas. The algorithm then forms the normalized operator $\widehat{\mathbf{H}} = D_V^{-1/2}\mathbf{H}WD_E^{-1}$ and applies $\widehat{\mathbf{H}}\mathbf{H}^{\top} D_V^{-1/2}$ to $X\Theta$, followed by $\sigma$. This is precisely the stated convolution expression. □

**Theorem 3.4.9** (Time complexity). Let $N := |V_0|$, $m := |E^{\uparrow}|$, and let $\mathrm{nnz}(\mathbf{H})$ be the number of nonzero entries in the incidence matrix $\mathbf{H}$ (equivalently, $\mathrm{nnz}(\mathbf{H}) = \sum_{j=1}^{m} |\xi_j|$). Let $T_{\mathrm{exp}}$ denote the total cost of constructing $E^{\uparrow}$ via Algorithm 1. Then one convolution layer (Algorithm 2) runs in

$$O(T_{\mathrm{exp}} + \mathrm{nnz}(\mathbf{H})\,(d+1) + N\,d\,c)$$

time under standard sparse-matrix arithmetic.

*Proof.* Algorithm 1 contributes $T_{\mathrm{exp}}$. In Algorithm 2, constructing $\mathbf{H}$ and computing $D_V, D_E$ are $O(\mathrm{nnz}(\mathbf{H}))$. Applying $\mathbf{H}^{\top}$ and $\mathbf{H}$ to dense feature matrices costs $O(\mathrm{nnz}(\mathbf{H})\,d)$. The final dense multiplication by $\Theta$ costs $O(Ndc)$, and the remaining diagonal scalings and activation are lower order. Summing yields the stated bound. □

Theorem 3.4.10 (Space complexity). With the same notation as above, the memory usage of one convolution layer is

$$O(\mathrm{nnz}(\mathbf{H}) \;+\; N(d+c) \;+\; md \;+\; dc),$$

assuming $\mathbf{H}$ is stored in a sparse format.

Proof. Storing $\mathbf{H}$ costs $O(\mathrm{nnz}(\mathbf{H}))$. The feature matrices $X$ and $Y$ cost $O(Nd)$ and $O(Nc)$, respectively. A typical sparse implementation forms an intermediate $m \times d$ matrix (e.g., $\mathbf{H}^\top(\cdot)$), costing $O(md)$. Finally, $\Theta$ costs $O(dc)$. Adding these terms gives the stated bound. □

Remark 3.4.11 (A coarse bound in terms of nesting parameters). If each nested object has branching factor at most $k$ and the nesting depth is at most $n$, then $|\xi_j| = |\,\mathrm{Expand}(\varepsilon_j)| \leq k^n$, hence $\mathrm{nnz}(\mathbf{H}) = \sum_{j=1}^{m} |\xi_j| \leq mk^n$. In this regime, the time bound simplifies to $O(T_{\exp} + mk^n(d+1) + Ndc)$, and the space bound to $O(mk^n + N(d+c) + md + dc)$.

## 3.5 Dynamic $n$-SuperHypergraph Neural Network

In this subsection we introduce a Dynamic $n$-SuperHypergraph Neural Network, motivated by the Dynamic Hypergraph Neural Network (DHGNN) [202]. A DHGNN models evolving relations by rebuilding the (hyper)edge set from the current embeddings at each layer, and then applying a hypergraph convolution. Similar ideas have also been studied in dynamic (graph) neural networks and their hypergraph extensions (cf. [203, 204, 205, 206]).

Definition 3.5.1 (Dynamic hypergraph [202]). A dynamic hypergraph at layer $l$ is a pair $\mathcal{H}_l = (V, \mathcal{E}_l)$, where $V$ is a fixed vertex set and $\mathcal{E}_l$ is a hyperedge family constructed from the layer-$l$ vertex embeddings $X_l$. Typically, $\mathcal{E}_l$ is obtained by a deterministic or learned rule (e.g., $k$-NN or clustering) applied to $X_l$.

Definition 3.5.2 (Dynamic Hypergraph Neural Network (DHGNN) [202]). A DHGNN is a layered architecture in which, at each layer $l$, one (i) constructs a dynamic hypergraph $\mathcal{H}_l = (V, \mathcal{E}_l)$ from the current embeddings $X_l$, and then (ii) applies a hypergraph convolution to produce updated embeddings $X_{l+1}$. Abstractly,

$$X_{l+1} = \sigma\big(X_l W_l \;+\; \mathrm{HGC}(\mathcal{H}_l, X_l)\big),$$

where $W_l$ is learnable and $\sigma$ is an activation function.

Definition 3.5.3 (Dynamic $n$-SuperHypergraph). Fix a finite base vertex set $V_0$ and an integer $n \in \mathbb{N}_0$. A dynamic $n$-SuperHypergraph is a finite sequence

$$\{\; H^{(l)} = (V^{(l)}, E^{(l)}) \;\}_{l=0}^{L},$$

such that for every layer $l$:

- $V^{(l)} \subseteq \mathcal{P}^n(V_0)$ is a finite set of $n$-supervertices;
- $E^{(l)} \subseteq \mathcal{P}(V^{(l)})$ is a finite family of superedges (each superedge is a subset of $V^{(l)}$).

The evolution $H^{(l)} \mapsto H^{(l+1)}$ may depend on the current embeddings (or other external signals).

Example 3.5.4 (Dynamic 2-SuperHypergraph for evolving service bundles). Let the finite base vertex set be

$$V_0 = \{s_1, s_2, s_3, s_4\},$$

where each $s_i$ represents a basic service component. We consider $n = 2$, so each 2-supervertex is an element of $\mathcal{P}^2(V_0) = \mathcal{P}(\mathcal{P}(V_0))$, i.e., a family of subsets of $V_0$.

Define four candidate 2-supervertices:

$$\begin{aligned} U_1 &:= \{\{s_1, s_2\}, \{s_3\}\}, \\ U_2 &:= \{\{s_2\}, \{s_4\}\}, \\ U_3 &:= \{\{s_1\}, \{s_3, s_4\}\}, \\ U_4 &:= \{\{s_1, s_4\}\}. \end{aligned}$$

Each $U_i$ is a set of subsets of $V_0$, hence $U_i \in \mathcal{P}^2(V_0)$.

Now define a dynamic 2-SuperHyperGraph

$$\{H^{(l)} = (V^{(l)}, E^{(l)})\}_{l=0}^{2}$$

as follows.

Layer $l = 0$ (initial deployment).

$$V^{(0)} = \{U_1, U_2, U_3\}, \qquad E^{(0)} = \{\{U_1, U_2\}\}.$$

This means that, at the initial stage, there is one active higher-order interaction between the supervertices $U_1$ and $U_2$.

Layer $l = 1$ (traffic increase / reconfiguration).

$$V^{(1)} = \{U_1, U_2, U_3\}, \qquad E^{(1)} = \{\{U_1, U_2\}, \{U_2, U_3\}\}.$$

A second superedge $\{U_2, U_3\}$ appears, indicating a new interaction pattern triggered by load balancing or event-routing changes.

Layer $l = 2$ (replacement of a bundle).

$$V^{(2)} = \{U_1, U_3, U_4\}, \qquad E^{(2)} = \{\{U_1, U_3\}, \{U_3, U_4\}\}.$$

Here $U_2$ is removed and a new supervertex $U_4$ is introduced (e.g., after a deployment update). The active superedges are also updated accordingly.

Thus $\{H^{(l)}\}_{l=0}^{2}$ is a concrete dynamic 2-SuperHyperGraph over $V_0$, illustrating how both the supervertex set and the superedge family may evolve over time. As a reference, a concrete example is shown in Fig. 3.1.

Theorem 3.5.5 (Static reduction). A single $n$-SuperHyperGraph $H = (V, E)$ is a special case of a dynamic $n$-SuperHypergraph, obtained by the constant sequence $H^{(l)} = H$ for all $l$.

Proof. If $H^{(l)} = H$ for every $l$, then each layer is an $n$-SuperHyperGraph by definition and no evolution occurs. □

Theorem 3.5.6 (Dynamic hypergraphs as a special case). A dynamic hypergraph $\{\mathcal{H}_l = (V_0, \mathcal{E}_l)\}_{l=0}^{L}$ is a special case of a dynamic $n$-SuperHypergraph (with $n = 0$).

Proof. For $n = 0$, one has $\mathcal{P}^0(V_0) = V_0$. Set $V^{(l)} := V_0$ and $E^{(l)} := \mathcal{E}_l \subseteq \mathcal{P}(V_0) = \mathcal{P}(V^{(l)})$. Then $\{(V^{(l)}, E^{(l)})\}_{l=0}^{L}$ is a dynamic 0-SuperHypergraph whose layers are exactly the given hypergraphs. □

Definition 3.5.7 (Dynamic $n$-SuperHypergraph Neural Network (D$n$-SHGNN)). A Dynamic $n$-SuperHypergraph Neural Network (abbrev. D$n$-SHGNN) is a layered model that, at each layer $l$, constructs an $n$-SuperHyperGraph $H^{(l)} = (V^{(l)}, E^{(l)})$ from the current base-vertex embeddings $X^{(l)} \in \mathbb{R}^{|V_0| \times d_l}$, expands it to a hypergraph on $V_0$, and then applies an $n$-SHGNN-type convolution.

More precisely, for each layer $l$:

Dynamic 2-SuperHyperGraph: Time-Evolving Supervertices and Superedges

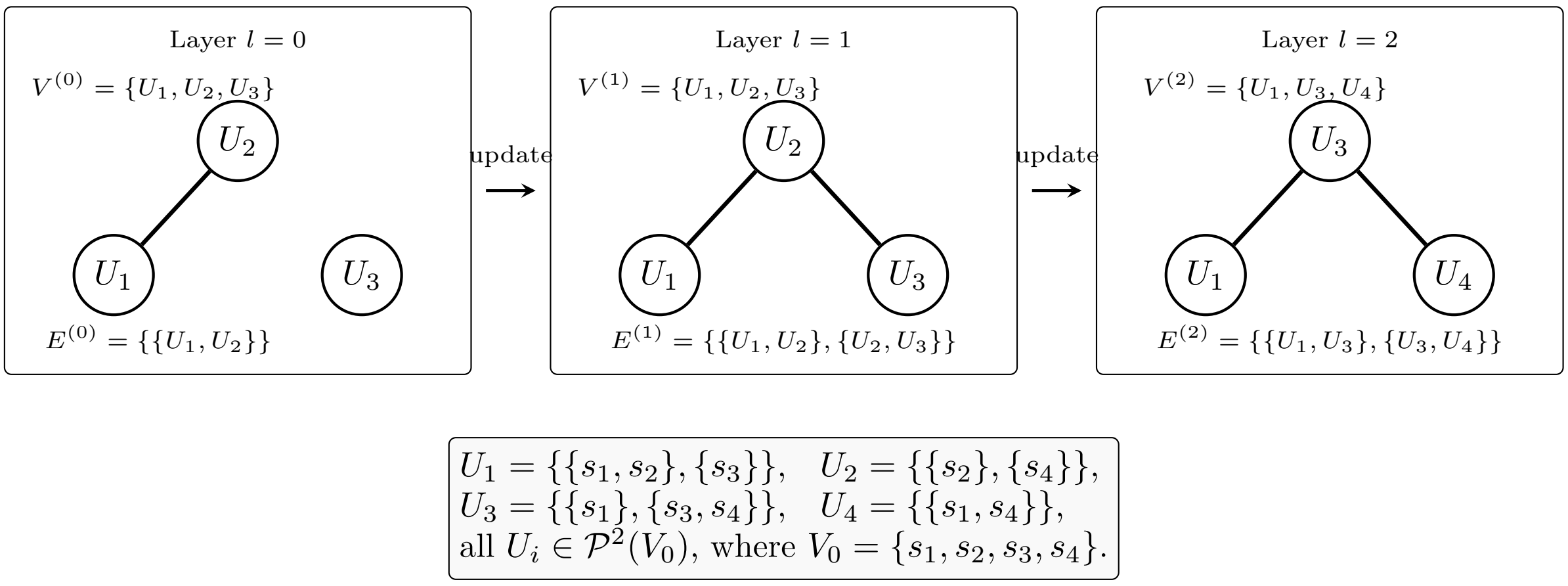

Figure 3.1: A dynamic 2-SuperHyperGraph over a fixed base set $V_0$. The sequence $H^{(0)} \to H^{(1)} \to H^{(2)}$ illustrates how both the supervertex set and the superedge family can evolve across discrete time layers.

1. (Dynamic $n$-SuperHypergraph construction) build $H^{(l)} = (V^{(l)}, E^{(l)})$ by a (possibly learned) rule

$$\mathsf{Build}_n : \mathbb{R}^{|V_0| \times d_l} \longrightarrow \big\{(V, E) : V \subseteq \mathcal{P}^n(V_0),\ E \subseteq \mathcal{P}(V)\big\}, \qquad H^{(l)} := \mathsf{Build}_n(X^{(l)}).$$

2. (Expansion) form the expanded hypergraph $H^{\uparrow(l)} = (V_0, E^{\uparrow(l)})$ by expanding each superedge $\varepsilon \in E^{(l)}$ to a base-level hyperedge $\mathrm{Expand}(\varepsilon) \subseteq V_0$, and collecting these expansions into $E^{\uparrow(l)}$.

3. (Convolution on the expansion) let $\mathbf{H}^{(l)} \in \{0,1\}^{|V_0| \times |E^{\uparrow(l)}|}$ be the incidence matrix of $H^{\uparrow(l)}$, let $W^{(l)}$ be a diagonal hyperedge-weight matrix, and define degree matrices $D_V^{(l)}$ and $D_E^{(l)}$ in the standard way from $\mathbf{H}^{(l)}$ and $W^{(l)}$. Then the layer update is

$$X^{(l+1)} = \sigma\Big(D_V^{(l)\,-1/2}\, \mathbf{H}^{(l)} W^{(l)} D_E^{(l)\,-1}\, \mathbf{H}^{(l)\top}\, D_V^{(l)\,-1/2}\, X^{(l)}\, \Theta^{(l)}\Big),$$

where $\Theta^{(l)}$ is learnable and $\sigma$ is an activation function.

Theorem 3.5.8 (Induced dynamic $n$-SuperHypergraph structure). A D$n$-SHGNN canonically determines a dynamic $n$-SuperHypergraph $\{H^{(l)}\}_{l=0}^{L}$ via its layerwise construction step.

Proof. By definition of a D$n$-SHGNN, at each layer $l$ the construction map $\mathsf{Build}_n$ outputs an $n$-SuperHyperGraph $H^{(l)} = (V^{(l)}, E^{(l)})$ with $V^{(l)} \subseteq \mathcal{P}^n(V_0)$ and $E^{(l)} \subseteq \mathcal{P}(V^{(l)})$. Collecting these outputs over $l = 0, \dots, L$ yields a dynamic $n$-SuperHypergraph sequence. □

We next give a simple procedure that dynamically constructs a 1-superhypergraph at layer $l$ from the current base-vertex embeddings $X^{(l)}$. The output is a superhypergraph $H^{(l)} = (V^{(l)}, E^{(l)})$ whose supervertices are subsets of $V_0$ and whose superedges are subsets of $V^{(l)}$.

Algorithm 4: Dynamic SuperHypergraph Construction (DSHC) at layer $l$

Input:

- Base-vertex embeddings $X^{(l)} = [x_1^{(l)}; \dots; x_n^{(l)}] \in \mathbb{R}^{n\times d}$, where $n := |V_0|$ and $x_i^{(l)} \in \mathbb{R}^d$ corresponds to $v_i \in V_0$.
- Parameters: number of supervertices $s \in \mathbb{N}$, max supervertex size $k \in \mathbb{N}$, number of superedges $t \in \mathbb{N}$, max superedge size $m \in \mathbb{N}$.
- A clustering routine $\mathsf{Cluster}(\cdot; \#\text{clusters})$ and a distance $\text{dist}(\cdot, \cdot)$ on $\mathbb{R}^d$.
- An aggregation operator $\mathsf{Agg}$ (e.g., mean) to obtain supervertex embeddings from base-vertex embeddings.

Output: A dynamically constructed 1-superhypergraph $H^{(l)} = (V^{(l)}, E^{(l)})$.

Step 1: Construct supervertices (subsets of $V_0$).;
Run $\mathsf{Cluster}(X^{(l)}; \#\text{clusters} = s)$ to obtain an assignment $a : \{1, \dots, n\} \to \{1, \dots, s\}$ and centroids $(\mu_r)_{r=1}^{s} \subset \mathbb{R}^d$;
for $r \leftarrow 1$ to $s$ do // form a bounded-size subset for each cluster
  $C_r \leftarrow \{\, i \in \{1, \dots, n\} \mid a(i) = r \,\}$;
  if $C_r = \emptyset$ then
    continue;
  end
  Order indices in $C_r$ by increasing $\text{dist}(x_i^{(l)}, \mu_r)$;
  $S_r \leftarrow$ the first $\min\{k, |C_r|\}$ indices in this order;
  Define the supervertex $v_r^{(l)} \leftarrow \{\, v_i \in V_0 \mid i \in S_r \,\} \subseteq V_0$;
end
$V^{(l)} \leftarrow \{\, v_r^{(l)} \neq \emptyset \mid 1 \le r \le s \,\}$;
Step 2: Construct superedges (subsets of $V^{(l)}$).;
For each $v \in V^{(l)}$, define a supervertex embedding

$$\bar{x}_v^{(l)} \;\leftarrow\; \mathsf{Agg}\big(\{\, x_i^{(l)} \mid v_i \in v \,\}\big) \in \mathbb{R}^d;$$

Run $\mathsf{Cluster}\big(\{\bar{x}_v^{(l)} : v \in V^{(l)}\}; \#\text{clusters} = t\big)$ to obtain a partition $(G_q)_{q=1}^{t}$ of $V^{(l)}$ into (possibly empty) groups and group centroids $(\nu_q)_{q=1}^{t}$;
for $q \leftarrow 1$ to $t$ do // cap each group size by $m$
  if $G_q = \emptyset$ then
    continue;
  end
  Order $v \in G_q$ by increasing $\text{dist}(\bar{x}_v^{(l)}, \nu_q)$;
  $e_q^{(l)} \leftarrow$ the first $\min\{m, |G_q|\}$ supervertices in this order;
  Define the superedge $e_q^{(l)} \subseteq V^{(l)}$ as that selected subset;
end
$E^{(l)} \leftarrow \{\, e_q^{(l)} \neq \emptyset \mid 1 \le q \le t \,\} \subseteq \mathcal{P}(V^{(l)})$;
return $H^{(l)} = (V^{(l)}, E^{(l)})$;

Theorem 3.5.9 (Layerwise correctness of DSHGNN update). Assume that, at layer $l$, a DSHGNN first constructs $H^{(l)} = (V^{(l)}, E^{(l)})$ using Algorithm DSHC, then builds the expanded hypergraph on $V_0$, forms the corresponding incidence/degree matrices, and finally applies the DSHGNN convolution operator of the form

$$X^{(l+1)} \;=\; \sigma\Big(D_V^{(l)\,-1/2}\,\mathbf{H}^{(l)} W^{(l)} D_E^{(l)\,-1}\,\mathbf{H}^{(l)\top}\,D_V^{(l)\,-1/2}\,X^{(l)}\,\Theta^{(l)}\Big).$$

Then the computed embedding matrix $X^{(l+1)}$ is exactly the output of that convolution operator applied to the dynamically constructed superhypergraph at layer $l$.

Proof. Algorithm DSHC outputs $H^{(l)} = (V^{(l)}, E^{(l)})$ by construction. The subsequent expansion step yields an expanded hypergraph on $V_0$, hence an incidence matrix $\mathbf{H}^{(l)}$, and the degree matrices $D_V^{(l)}$, $D_E^{(l)}$ and weight matrix $W^{(l)}$ are computed from these objects by their definitions. Finally, the layer update applies the displayed formula, so the resulting $X^{(l+1)}$ coincides with the convolution operator output. □

Theorem 3.5.10 (Per-layer time complexity). Let $n := |V_0|$, $d$ be the input embedding dimension, $c$ the output dimension, $s_l := |V^{(l)}| \leq s$, $t_l := |E^{(l)}| \leq t$, and assume $|v| \leq k$ for all $v \in V^{(l)}$ and $|e| \leq m$ for all $e \in E^{(l)}$. Let $I_1$ and $I_2$ be the iteration counts of the two clustering calls in Algorithm DSHC. Then the total per-layer runtime of (i) construction + (ii) expansion + (iii) one convolution is

$$O\Big(I_1\, ns\, d \;+\; I_2\, st\, d \;+\; t_l\, mk \;+\; \mathrm{nnz}(\mathbf{H}^{(l)})\, d \;+\; ndc\Big),$$

where $\mathrm{nnz}(\mathbf{H}^{(l)}) \leq \sum_{e\in E^{\uparrow(l)}} |e| \leq t_l\, mk$. In particular, one may bound $\mathrm{nnz}(\mathbf{H}^{(l)})\, d$ by $O(t_l\, mk\, d)$.

Proof. Algorithm DSHC performs (a) clustering $n$ points in $\mathbb{R}^d$ into $s$ clusters, costing $O(I_1 nsd)$, and (b) clustering $s_l \leq s$ supervertex embeddings into $t$ groups, costing $O(I_2 std)$. Selecting at most $k$ base vertices per cluster and at most $m$ supervertices per group adds only linear overhead. Expansion forms each expanded hyperedge as a union of at most $m$ supervertices of size at most $k$, hence costs $O(t_l mk)$. The convolution can be implemented with sparse products whose cost is $O(\mathrm{nnz}(\mathbf{H}^{(l)})\, d)$, followed by a dense multiplication by $\Theta^{(l)}$ costing $O(ndc)$. Combining these bounds yields the stated complexity. □

Theorem 3.5.11 (Per-layer space complexity). Under the same notation, the space required to store the layer-$l$ objects (embeddings, constructed superhypergraph, and one convolution) is

$$O\Big(nd \;+\; sd \;+\; \mathrm{nnz}(\mathbf{H}^{(l)}) \;+\; dc \;+\; nc\Big),$$

where $\mathrm{nnz}(\mathbf{H}^{(l)}) \leq t_l mk$.

Proof. We store the input and output embeddings $X^{(l)}, X^{(l+1)}$ (cost $O(nd + nc)$), supervertex embeddings and bookkeeping for $V^{(l)}$ (cost $O(sd)$), the sparse incidence matrix $\mathbf{H}^{(l)}$ (cost $O(\mathrm{nnz}(\mathbf{H}^{(l)}))$), and the weight matrix $\Theta^{(l)}$ (cost $O(dc)$). Summing these contributions gives the bound. □

Theorem 3.5.12 (DHGNN as a special case of DSHGNN). The Dynamic Hypergraph Neural Network (DHGNN) is a special case of the Dynamic SuperHypergraph Neural Network (DSHGNN). More precisely, if at every layer $l$ the constructed supervertex family consists only of singletons,

$$V^{(l)} = \big\{\{v_1\}, \ldots, \{v_n\}\big\},$$

then each superedge $e \in E^{(l)} \subseteq \mathcal{P}(V^{(l)})$ expands to the base-level hyperedge $\bigcup_{u\in e} u \subseteq V_0$, and the DSHGNN layer update reduces to the DHGNN update on the resulting dynamic hypergraph.

Proof. If every supervertex is a singleton, the expansion map sends each superedge $e \subseteq V^{(l)}$ to the hyperedge $e^{\uparrow} = \bigcup_{u\in e} u \subseteq V_0$, so the expanded hypergraph is exactly a (dynamic) hypergraph on $V_0$. With this identification, the incidence and degree matrices coincide with those used in a DHGNN layer, and the DSHGNN convolution formula becomes the usual hypergraph convolution on the dynamically reconstructed hypergraph. □

## 3.6 Multi-Graph Neural Networks and Their Generalization

Multi-graph neural networks have been proposed in recent years (e.g., [207, 208, 209]). In this section we clarify, in a mathematically precise way, how multi-graphs and MGNN-type message passing fit naturally into the framework of $n$-SuperHyperGraph neural networks.

Definition 3.6.1 (Multi-graph in incidence form (undirected)). (cf. [210]) A (finite, undirected) multi-graph is a triple

$$G = (V, E, \varphi),$$

where $V$ is a finite vertex set, $E$ is a finite edge-index set, and

$$\varphi : E \longrightarrow \mathrm{MS}_2(V)$$

is an endpoint map into the set $\mathrm{MS}_2(V)$ of 2-element multisets on $V$. Thus, $\varphi(e) = \{u, v\}$ (as a multiset) allows $u = v$, in which case $\varphi(e) = \{v, v\}$ represents a loop. Distinct edges $e_1 \neq e_2$ may satisfy $\varphi(e_1) = \varphi(e_2)$, which models parallel edges.

Remark 3.6.2. The incidence-form definition is the standard way to treat parallel edges without ambiguity: parallelism is represented by distinct indices in $E$ that map to the same endpoints under $\varphi$. (If one prefers directed multi-graphs, replace $\mathrm{MS}_2(V)$ by $V \times V$.)

Theorem 3.6.3. Every multi-graph can be encoded as a 1-SuperHyperGraph (at the base level) once superedges are treated as indexed objects (equivalently, via an incidence map).

Proof. Let $G = (V, E, \varphi)$ be a multi-graph. Set $V_0 := V$ and define the singleton supervertex family

$$V^\star \;:=\; \big\{\{v\} \subseteq V_0 \;\big|\; v \in V_0\big\} \;\subseteq\; \mathcal{P}(V_0).$$

Define an incidence map $\iota : E \to \mathcal{P}(V^\star)$ by

$$\iota(e) \;:=\; \begin{cases} \big\{\{u\}, \{v\}\big\}, & \text{if } \varphi(e) = \{u, v\} \text{ with } u \neq v, \\ \big\{\{v\}\big\}, & \text{if } \varphi(e) = \{v, v\} \text{ (a loop).} \end{cases}$$

Then $\iota(e)$ is a (super)edge connecting singleton supervertices, and distinct edges in $E$ remain distinct even if they have the same endpoints, since they remain distinct as indices. Hence the incidence-encoded object $\big(V^\star, E, \iota\big)$ is a 1-level superhypergraph representation of $G$, and it reproduces both loops and parallel edges. □

Definition 3.6.4 (Typed multi-graph). A typed multi-graph is a tuple $(V, E, \varphi, \tau)$, where $(V, E, \varphi)$ is a multi-graph and $\tau : E \to T$ assigns an edge type to each edge, for some finite type set $T$. We write $E_t := \tau^{-1}(\{t\})$ for the set of edges of type $t$.

Definition 3.6.5 (Multi-Graph Neural Network (MGNN) in message-passing form). (cf. [209]) Let $(V, E, \varphi, \tau)$ be a typed multi-graph. An MGNN layer updates node embeddings $\{h_v^{(\ell)} \in \mathbb{R}^{d_\ell}\}_{v \in V}$ to $\{h_v^{(\ell+1)}\}_{v \in V}$ by aggregating type-dependent messages. A standard template is

$$h_v^{(\ell+1)} \;=\; \Phi\left( h_v^{(\ell)},\; \bigoplus_{t \in T} \quad \bigoplus_{\substack{e \in E_t \\ v \in \mathrm{supp}(\varphi(e))}} \quad \Psi_t\Big(h_u^{(\ell)}, h_v^{(\ell)},\, x_e\Big) \right),$$

where:

- $\mathrm{supp}(\varphi(e)) = \{u, v\}$ (as a set) is the set of incident endpoints of $e$ (for loops, $\mathrm{supp}(\varphi(e)) = \{v\}$);
- $u$ denotes the other endpoint of $e$ when $e$ is not a loop (for loops one may set $u = v$);
- $x_e$ is an optional edge feature;
- $\Psi_t$ is a type-specific message function, $\Phi$ is a node update function, and $\bigoplus$ is an aggregation operator (e.g., sum/mean/max).

Theorem 3.6.6. An $n$-SuperHyperGraph neural network can generalize an MGNN in the following precise sense: every MGNN layer on a typed multi-graph $(V, E, \varphi, \tau)$ admits an equivalent realization as a 1-SuperHyperGraph neural construction on an incidence structure obtained from $(V, E, \varphi, \tau)$, provided the $n$-SHGNN layer is allowed to use edge (hyperedge) attributes to parameterize its messages.

Proof. Let $(V, E, \varphi, \tau)$ be a typed multi-graph. By Theorem 3.6.3, we may represent it as a 1-level superhypergraph incidence object $(V^{\star}, E, \iota)$ on the base set $V_0 = V$, where each $e \in E$ is an indexed superedge $\iota(e) \subseteq V^{\star}$.

Now form the expanded hypergraph on the base vertices $V_0$ by setting, for each indexed edge $e \in E$,

$$e^{\uparrow} := \bigcup_{S \in \iota(e)} S \subseteq V_0, \qquad E^{\uparrow} := \{ e^{\uparrow} : e \in E \},$$

where $E^{\uparrow}$ is understood as an indexed family (so parallel edges yield repeated hyperedges with distinct indices). Attach to each expanded hyperedge $e^{\uparrow}$ the attributes $(\tau(e), x_e)$. A 1-SHGNN/$n$-SHGNN layer that performs hyperedge-indexed (hence edge-indexed) aggregation with type-dependent kernels can then reproduce the MGNN update by choosing its hyperedge message kernel to match $\Psi_t$ and its vertex update to match $\Phi$ (with the same aggregation operator $\bigoplus$). Because the hyperedges are indexed by $E$, parallel edges contribute separately, exactly as in the MGNN. Therefore, the MGNN layer is realized as a special case of an $n$-SHGNN layer (already with $n = 1$). □

Remark 3.6.7. The key technical point is that parallel edges require an indexed hyperedge family (equivalently, an incidence map), so that two distinct edges with identical endpoints remain distinct objects and can contribute separately during aggregation. This is the same convention used implicitly whenever an incidence matrix is allowed to have repeated columns.

## 3.7 Directed Hypergraph Neural Networks

Directed graph neural networks (DGNNs) extend standard message passing by treating incoming and outgoing adjacency separately, so that edge orientation is explicitly encoded in the aggregation and update steps (cf. [44, 45, 211, 212]). Building on this idea, directed hypergraph neural networks distinguish between incoming and outgoing hyperarcs—each hyperarc sends information from a nonempty tail set of sources to a nonempty head set of targets—thereby enabling directional high-order message passing (cf. [213, 214, 215]). We formalize both models below.

Definition 3.7.1 (Directed Graph Neural Network (DGNN)). (cf.[44, 45]) Let $G = (V, E)$ be a finite directed graph with $E \subseteq V \times V$. For each node $v \in V$, let $x_v \in \mathbb{R}^F$ be an input feature, and set $h_v^{(0)} := x_v$. Optionally, each arc $(u, v) \in E$ carries an edge feature $x_{uv} \in \mathbb{R}^{F_e}$.

For $\ell = 1, \ldots, L$, define the in/out neighborhoods

$$\mathcal{N}^{-}(v) := \{u \in V : (u, v) \in E\}, \qquad \mathcal{N}^{+}(v) := \{u \in V : (v, u) \in E\},$$

and compute messages by

$$m_{u \to v}^{(\ell)} = \psi^{(\ell)}\big(h_u^{(\ell-1)}, h_v^{(\ell-1)}, x_{uv}\big) \qquad ((u, v) \in E),$$

where $\psi^{(\ell)}$ is a learnable function (e.g. an MLP), and $x_{uv}$ may be omitted if edge features are not used. Aggregate incoming and outgoing messages separately via permutation-invariant operators $\mathrm{AGG}^{-}, \mathrm{AGG}^{+}$:

$$\bar{m}_v^{-(\ell)} = \mathrm{AGG}^{-}\big(\{ m_{u \to v}^{(\ell)} : u \in \mathcal{N}^{-}(v) \}\big), \qquad \bar{m}_v^{+(\ell)} = \mathrm{AGG}^{+}\big(\{ m_{v \to u}^{(\ell)} : u \in \mathcal{N}^{+}(v) \}\big),$$

and update node embeddings by

$$h_v^{(\ell)} = \phi^{(\ell)}\big(h_v^{(\ell-1)}, \bar{m}_v^{-(\ell)}, \bar{m}_v^{+(\ell)}\big),$$

where $\phi^{(\ell)}$ is learnable (e.g. MLP/GRU). A graph-level representation may be obtained by

$$h_G = R\big(\{ h_v^{(L)} : v \in V \}\big),$$

with permutation-invariant readout $R$ (e.g. sum/mean/max).

Definition 3.7.2 (Directed Hypergraph Neural Network (DHGNN)). (cf.[213, 214]) Let $H = (V, \mathcal{E})$ be a finite directed hypergraph, where each hyperarc $e \in \mathcal{E}$ is an ordered pair $e = (T(e), H(e))$ with $\emptyset \neq T(e) \subseteq V$ (tail) and $\emptyset \neq H(e) \subseteq V$ (head). Assign node features $x_v \in \mathbb{R}^F$ for $v \in V$ and hyperarc features $x_e \in \mathbb{R}^{F_e}$ for $e \in \mathcal{E}$, and set $h_v^{(0)} := x_v$.

For $\ell = 1, \ldots, L$, define incoming/outgoing hyperarc sets at a node $v \in V$ by

$$\mathrm{In}(v) := \{\, e \in \mathcal{E} : v \in H(e) \,\}, \qquad \mathrm{Out}(v) := \{\, e \in \mathcal{E} : v \in T(e) \,\}.$$

Let $\rho_T^{(\ell)}$ and $\rho_H^{(\ell)}$ be permutation-invariant aggregators over tail and head node embeddings (e.g. sum/mean), and let $\psi_e^{(\ell)}$, $\phi_v^{(\ell)}$ be learnable update functions. Compute:

$$\begin{aligned}
s_T^{(\ell)}(e) &:= \rho_T^{(\ell)}\big(\{\, h_u^{(\ell-1)} : u \in T(e) \,\}\big),\\
s_H^{(\ell)}(e) &:= \rho_H^{(\ell)}\big(\{\, h_w^{(\ell-1)} : w \in H(e) \,\}\big),\\
m_e^{(\ell)} &:= \psi_e^{(\ell)}\big(x_e,\, s_T^{(\ell)}(e),\, s_H^{(\ell)}(e)\big),\\
\bar{m}_v^{-(\ell)} &:= \mathrm{AGG}^{-}\big(\{\, m_e^{(\ell)} : e \in \mathrm{In}(v) \,\}\big),\\
\bar{m}_v^{+(\ell)} &:= \mathrm{AGG}^{+}\big(\{\, m_e^{(\ell)} : e \in \mathrm{Out}(v) \,\}\big),\\
h_v^{(\ell)} &:= \phi_v^{(\ell)}\big(h_v^{(\ell-1)},\, \bar{m}_v^{-(\ell)},\, \bar{m}_v^{+(\ell)}\big).
\end{aligned}$$

Finally, a hypergraph-level vector may be obtained by a permutation-invariant readout $h_H := R(\{h_v^{(L)} : v \in V\})$.

## 3.8 Directed SuperHyperGraph Neural Networks

A directed $n$-SuperHyperGraph Neural Network propagates information over directed $n$-superedges, distinguishing tail-to-head flow at the supervertex level, and thereby modeling nested, high-order directional interactions via message passing.

Definition 3.8.1 (Directed $n$-SuperHyperGraph Neural Network (DnSHGNN)). Let $\mathrm{DSuHG}^{(n)} = (V, \mathcal{E})$ be a finite directed $n$-superhypergraph, where $V \subseteq \mathcal{P}^n(S)$ is the set of $n$-supervertices and each directed $n$-superedge is an ordered pair

$$e = \big(\mathrm{Tail}(e), \mathrm{Head}(e)\big) \in \mathcal{E} \subseteq \big(\mathcal{P}(V) \setminus \{\emptyset\}\big) \times \big(\mathcal{P}(V) \setminus \{\emptyset\}\big),$$

with $\emptyset \neq \mathrm{Tail}(e) \subseteq V$ and $\emptyset \neq \mathrm{Head}(e) \subseteq V$. Assign supervertex features $x_v \in \mathbb{R}^F$ for $v \in V$ and superedge features $x_e \in \mathbb{R}^{F_e}$ for $e \in \mathcal{E}$, and set $h_v^{(0)} := x_v$.

For $\ell = 1, \ldots, L$, define

$$\mathrm{In}(v) := \{\, e \in \mathcal{E} : v \in \mathrm{Head}(e) \,\}, \qquad \mathrm{Out}(v) := \{\, e \in \mathcal{E} : v \in \mathrm{Tail}(e) \,\}.$$

Let $\rho_{\mathrm{Tail}}^{(\ell)}$, $\rho_{\mathrm{Head}}^{(\ell)}$, $\mathrm{AGG}^{-}$, $\mathrm{AGG}^{+}$ be permutation-invariant aggregators, and let $\psi_e^{(\ell)}$, $\phi_v^{(\ell)}$ be learnable functions. Compute:

$$\begin{aligned}
s_{\mathrm{Tail}}^{(\ell)}(e) &:= \rho_{\mathrm{Tail}}^{(\ell)}\big(\{\, h_u^{(\ell-1)} : u \in \mathrm{Tail}(e) \,\}\big),\\
s_{\mathrm{Head}}^{(\ell)}(e) &:= \rho_{\mathrm{Head}}^{(\ell)}\big(\{\, h_w^{(\ell-1)} : w \in \mathrm{Head}(e) \,\}\big),\\
m_e^{(\ell)} &:= \psi_e^{(\ell)}\big(x_e,\, s_{\mathrm{Tail}}^{(\ell)}(e),\, s_{\mathrm{Head}}^{(\ell)}(e)\big),\\
\bar{m}_v^{-(\ell)} &:= \mathrm{AGG}^{-}\big(\{\, m_e^{(\ell)} : e \in \mathrm{In}(v) \,\}\big),\\
\bar{m}_v^{+(\ell)} &:= \mathrm{AGG}^{+}\big(\{\, m_e^{(\ell)} : e \in \mathrm{Out}(v) \,\}\big),\\
h_v^{(\ell)} &:= \phi_v^{(\ell)}\big(h_v^{(\ell-1)},\, \bar{m}_v^{-(\ell)},\, \bar{m}_v^{+(\ell)}\big).
\end{aligned}$$

A global readout may then produce a representation of the directed $n$-superhypergraph:

$$h_{\mathrm{DSuHG}^{(n)}} \;:=\; R\big(\{\, h_v^{(L)} : v \in V \,\}\big),$$

where $R$ is permutation-invariant.

## 3.9 Recursive SuperHyperGraph Neural Networks

In this section, we define a Recursive SuperHyperGraph Neural Network (RSHGNN), which combines:

- the hierarchical supervertex structure (via iterated powersets over a base set $V_0$), and
- the recursive superhyperedge structure (via bounded-depth recursive set formation over the supervertex set $V$).

We then prove that this framework strictly generalizes the (ordinary) SuperHyperGraph Neural Network (SHGNN).

Let $V_0 = \{x_1, \dots, x_N\}$ be a finite nonempty base set, and let $n \in \mathbb{N}_0$. Define

$$\mathcal{P}^0(V_0) := V_0, \qquad \mathcal{P}^{m+1}(V_0) := \mathcal{P}\big(\mathcal{P}^m(V_0)\big) \quad (m \geq 0),$$

and the hierarchical universe

$$\mathcal{U}_n(V_0) := \bigcup_{m=0}^{n} \Big(\mathcal{P}^m(V_0) \setminus \{\emptyset\}\Big).$$

Let $V = \{v_1, \dots, v_p\} \subseteq \mathcal{U}_n(V_0)$ be a finite set of $n$-supervertices. We use two support maps.

Definition 3.9.1 (Base support of a supervertex). The base support map

$$\mathrm{Supp}_0 : \mathcal{U}_n(V_0) \longrightarrow \mathcal{P}(V_0) \setminus \{\emptyset\}$$

is defined recursively by

$$\mathrm{Supp}_0(x) := \{x\} \quad (x \in V_0), \qquad \mathrm{Supp}_0(X) := \bigcup_{y \in X} \mathrm{Supp}_0(y) \quad (X \in \mathcal{U}_n(V_0) \setminus V_0).$$

Thus $\mathrm{Supp}_0(v_a) \subseteq V_0$ is the set of base vertices contained (recursively) in $v_a$.

Next, let $k \in \mathbb{N}_0$, and define the depth-$k$ recursive universe over $V$ by

$$S_0 := V, \qquad S_r := \mathcal{P}\Big(\bigcup_{t=0}^{r-1} S_t\Big) \quad (r \geq 1), \qquad U_{V,k} := \bigcup_{r=0}^{k} S_r.$$

Definition 3.9.2 (Supervertex support of a recursive edge object). The supervertex support map

$$\mathrm{Supp}_V : U_{V,k} \longrightarrow \mathcal{P}(V)$$

is defined recursively by

$$\mathrm{Supp}_V(v) := \{v\} \quad (v \in V), \qquad \mathrm{Supp}_V(Z) := \bigcup_{y \in Z} \mathrm{Supp}_V(y) \quad (Z \in U_{V,k} \setminus V).$$

For a nonempty recursive superedge $e \subseteq U_{V,k}$, define its effective supervertex support by

$$\rho(e) := \bigcup_{z \in e} \mathrm{Supp}_V(z) \subseteq V.$$

Remark 3.9.3. The set $e$ may contain nested objects (e.g., sets of supervertices, or sets of sets of supervertices). The map $\rho(e)$ "flattens" this recursive structure to the supervertex domain $V$, which is the domain on which message passing is performed.

We now define the neural layer on an $(n, k)$-recursive SuperHyperGraph.

Definition 3.9.4 (Recursive SuperHyperGraph Neural Network layer (RSHGNN layer)). Let

$$\mathbb{H} = \mathrm{RSHG}^{(n,k)} = (V, E)$$

be an $(n, k)$-recursive SuperHyperGraph over $V_0$, where

$$V = \{v_1, \dots, v_p\} \subseteq \mathcal{U}_n(V_0), \qquad E = \{e_1, \dots, e_q\}, \qquad e_j \subseteq U_{V,k} \ \ (1 \le j \le q).$$

Assume the effective support nondegeneracy condition

$$\rho(e_j) \neq \emptyset \qquad (j = 1, \dots, q),$$

where $\rho(e_j)$ is as in Definition 3.9.2.

Let base-vertex features be given by

$$X \in \mathbb{R}^{N \times d}, \qquad \text{with row } i \text{ corresponding to } x_i \in V_0.$$

A RSHGNN layer is defined by the following operators.

(1) Base-to-supervertex membership and lifting. Define the membership matrix $B \in \{0,1\}^{N \times p}$ by

$$B_{ia} := \begin{cases} 1, & x_i \in \mathrm{Supp}_0(v_a), \\ 0, & \text{otherwise.} \end{cases}$$

Define diagonal matrices

$$(D_{\mathrm{sv}})_{aa} := \sum_{i=1}^{N} B_{ia} = |\,\mathrm{Supp}_0(v_a)|, \qquad (D_0)_{ii} := \sum_{a=1}^{p} B_{ia}.$$

(If some diagonal entries are zero, use the Moore–Penrose pseudoinverse in place of $D_{\mathrm{sv}}^{-1}$ or $D_0^{-1}$.)

Lift base features to supervertex features by

$$\widetilde{X} := D_{\mathrm{sv}}^{-1} B^{\top} X \in \mathbb{R}^{p \times d}.$$

(2) Recursive superedge incidence (flattened to $V$). Define the effective incidence matrix

$$\Pi \in \{0,1\}^{p \times q}$$

by

$$\Pi_{aj} := \begin{cases} 1, & v_a \in \rho(e_j), \\ 0, & \text{otherwise.} \end{cases}$$

Let $w : E \to (0, \infty)$ be a recursive superedge-weight function, and define

$$W := \mathrm{diag}\big(w(e_1), \dots, w(e_q)\big).$$

Define the diagonal degree matrices

$$(D_V^{\mathrm{rec}})_{aa} := \sum_{j=1}^{q} \Pi_{aj}\, w(e_j), \qquad (D_E^{\mathrm{rec}})_{jj} := \sum_{a=1}^{p} \Pi_{aj} = |\rho(e_j)|.$$

(3) Normalized propagation on supervertices. Define the recursive propagation operator

$$S_{\mathrm{rec}} := \big(D_V^{\mathrm{rec}}\big)^{-1/2}\, \Pi\, W\, \big(D_E^{\mathrm{rec}}\big)^{-1}\, \Pi^{\top}\, \big(D_V^{\mathrm{rec}}\big)^{-1/2} \in \mathbb{R}^{p \times p}.$$

(4) Convolution on the recursive SuperHyperGraph. Given learnable parameters $\Theta \in \mathbb{R}^{d \times c}$ and an activation $\sigma$, define

$$\widetilde{Y} := \sigma\Big(S_{\mathrm{rec}}\, \widetilde{X}\, \Theta\Big) \in \mathbb{R}^{p \times c}.$$

(5) Optional projection back to the base level. If a base-vertex output is required, define

$$Y := D_0^{-1} B \widetilde{Y} \in \mathbb{R}^{N\times c}.$$

A multi-layer RSHGNN is obtained by iterating Step (4) with layer-wise parameters $\Theta^{(\ell)}$, using $\widetilde{X}^{(0)} := \widetilde{X}$ and

$$\widetilde{X}^{(\ell+1)} = \sigma\Big(S_{\mathrm{rec}}\, \widetilde{X}^{(\ell)}\, \Theta^{(\ell)}\Big).$$

We now prove that RSHGNN generalizes the usual SHGNN (i.e., the case of non-recursive superedges).

Theorem 3.9.5 (RSHGNN strictly generalizes SHGNN). Fix $n \in \mathbb{N}_0$. Let

$$\mathbb{H} = (V, E)$$

be an $n$-SuperHyperGraph (equivalently, a hierarchical SuperHyperGraph of height $n$ with ordinary superedges), so that

$$V \subseteq \mathcal{U}_n(V_0), \qquad E = \{e_1, \dots, e_q\}, \qquad e_j \subseteq V \quad (1 \le j \le q).$$

Let $X$, $\Theta$, $\sigma$, and the edge weights $w(e_j)$ be fixed.

Then the SHGNN layer on $\mathbb{H}$ is exactly the $(n, 0)$-RSHGNN layer (Definition 3.9.4) on the same data. In particular,

$$S_{\mathrm{rec}} = S_{\mathrm{shg}}, \qquad \widetilde{Y}_{\mathrm{rec}} = \widetilde{Y}_{\mathrm{shg}}, \qquad Y_{\mathrm{rec}} = Y_{\mathrm{shg}}$$

whenever the optional projection is used.

Proof. For $k = 0$, by construction of the recursive universe over $V$,

$$U_{V,0} = V.$$

Hence every recursive superedge $e_j$ is simply an ordinary nonempty subset of $V$:

$$e_j \subseteq U_{V,0} = V.$$

Now consider the supervertex support map $\mathrm{Supp}_V$ from Definition 3.9.2. Since every element of $e_j$ is already in $V$, one has

$$\mathrm{Supp}_V(v) = \{v\} \qquad (v \in V),$$

and therefore

$$\rho(e_j) = \bigcup_{z\in e_j} \mathrm{Supp}_V(z) = \bigcup_{z\in e_j} \{z\} = e_j.$$

Thus the effective support of each recursive edge coincides exactly with the ordinary superedge itself.

Consequently, the effective incidence matrix $\Pi$ satisfies

$$\Pi_{aj} = 1 \iff v_a \in \rho(e_j) \iff v_a \in e_j.$$

But the right-hand condition is precisely the usual SHGNN incidence relation, so

$$\Pi = H,$$

where $H$ is the standard superedge incidence matrix of the SHGNN.

It follows immediately that the degree matrices also coincide:

$$D_V^{\mathrm{rec}} = D_{\mathcal{V}}, \qquad D_E^{\mathrm{rec}} = D_{\mathcal{E}},$$

because $D_E^{\mathrm{rec}}$ uses $|\rho(e_j)| = |e_j|$, and $D_V^{\mathrm{rec}}$ uses the same weighted incidence counts.

Therefore,

$$S_{\text{rec}} = \big(D_V^{\text{rec}}\big)^{-1/2} \Pi W \big(D_E^{\text{rec}}\big)^{-1} \Pi^\top \big(D_V^{\text{rec}}\big)^{-1/2} = D_{\mathcal{V}}^{-1/2} H W D_{\mathcal{E}}^{-1} H^\top D_{\mathcal{V}}^{-1/2} = S_{\text{shg}}.$$

The lifting operator $B$ and the lifted features $\widetilde{X} = D_{\text{sv}}^{-1} B^\top X$ depend only on the supervertex set $V$ (through $\text{Supp}_0$), and are independent of $k$. Hence they are identical in both models. Using the same $\Theta$ and $\sigma$, we obtain

$$\widetilde{Y}_{\text{rec}} = \sigma\Big(S_{\text{rec}} \widetilde{X} \Theta\Big) = \sigma\Big(S_{\text{shg}} \widetilde{X} \Theta\Big) = \widetilde{Y}_{\text{shg}}.$$

If the optional projection is applied, then $B$ and $D_0$ are again the same in both models, so

$$Y_{\text{rec}} = D_0^{-1} B \widetilde{Y}_{\text{rec}} = D_0^{-1} B \widetilde{Y}_{\text{shg}} = Y_{\text{shg}}.$$

This proves the claim. □

Corollary 3.9.6 (Recovery of the 1-SHGNN). If $n = 1$ and $k = 0$, then the RSHGNN layer reduces to the canonical 1-SHGNN layer.

More precisely, if

$$V = \{S_1, \ldots, S_p\} \subseteq \mathcal{P}(V_0) \setminus \{\emptyset\},$$

then $\text{Supp}_0(S_a) = S_a$, and the membership matrix $B$ becomes exactly

$$B_{ia} = 1 \iff x_i \in S_a,$$

which is the usual 1-SHGNN base–supervertex incidence. Together with Theorem 3.9.5, this yields the standard 1-SHGNN propagation rule.

Remark 3.9.7 (Why the recursive model is genuinely more expressive). The case $k > 0$ allows a superedge $e$ to contain nested objects such as

$$e = \big\{\, v_1,\ \{v_2, v_3\},\ \{\{v_4\}\}\, \big\},$$

which is not an ordinary superedge (since it is not a subset of $V$). The flattening map $\rho(e)$ still induces a well-defined propagation neighborhood on $V$, while the recursive representation preserves additional structural information that can be used in more advanced architectures (e.g., recursive edge encoders, attention over nesting depth, or typed nested relations).

## 3.10 Hierarchical SuperHyperGraph Neural Networks

We now define a Hierarchical SuperHyperGraph Neural Network (HSHGNN), which extends the usual SuperHyperGraph Neural Network (SHGNN) by allowing supervertices from multiple iterated-powerset levels to coexist in a single model.

Let $V_0 = \{x_1, \ldots, x_N\}$ be a finite nonempty base set, and let $r \in \mathbb{N}_0$. Define

$$\mathcal{P}^0(V_0) := V_0, \qquad \mathcal{P}^{m+1}(V_0) := \mathcal{P}\big(\mathcal{P}^m(V_0)\big) \quad (m \ge 0),$$

and the height-$r$ hierarchical universe

$$\mathcal{U}_r(V_0) := \bigcup_{m=0}^{r} \Big(\mathcal{P}^m(V_0) \setminus \{\emptyset\}\Big).$$

For $z \in \mathcal{U}_r(V_0)$, define its (minimal) level by

$$\ell(z) := \min\{\, m \in \{0, \ldots, r\} : z \in \mathcal{P}^m(V_0)\,\}.$$

Definition 3.10.1 (Hierarchical support map). The base support map

$$\mathrm{Supp} : \mathcal{U}_r(V_0) \longrightarrow \mathcal{P}(V_0) \setminus \{\emptyset\}$$

is defined recursively by

$$\mathrm{Supp}(x) := \{x\} \quad (x \in V_0), \qquad \mathrm{Supp}(X) := \bigcup_{y \in X} \mathrm{Supp}(y) \quad (X \in \mathcal{U}_r(V_0) \setminus V_0).$$

Thus $\mathrm{Supp}(z)$ is the set of base vertices contained in $z$, recursively.

We define the neural layer on a mixed-level supervertex domain. (If the downward-closure coherence condition is additionally imposed, this becomes the coherent hierarchical case.)

Definition 3.10.2 (Hierarchical SuperHyperGraph Neural Network layer (HSHGNN layer)). Let

$$\mathbb{H}^{\langle r \rangle} = (V, E)$$

be a height-$r$ hierarchical superhypergraph domain over $V_0$, where

$$V = \{v_1, \dots, v_p\} \subseteq \mathcal{U}_r(V_0), \qquad E = \{\varepsilon_1, \dots, \varepsilon_q\}, \qquad \varepsilon_j \in \mathcal{P}(V) \setminus \{\emptyset\}.$$

Assume base-vertex features

$$X \in \mathbb{R}^{N \times d}, \qquad \text{row } i \text{ corresponds to } x_i \in V_0.$$

An HSHGNN layer is defined by the following operators.

(1) Base–supervertex membership (hierarchical lifting map). Define the membership matrix $B \in \{0,1\}^{N \times p}$ by

$$B_{ia} := \begin{cases} 1, & x_i \in \mathrm{Supp}(v_a), \\ 0, & \text{otherwise.} \end{cases}$$

Define diagonal matrices

$$(D_{\mathrm{sup}})_{aa} := \sum_{i=1}^{N} B_{ia} = |\,\mathrm{Supp}(v_a)|, \qquad (D_0)_{ii} := \sum_{a=1}^{p} B_{ia}.$$

(If some diagonal entries vanish, use the Moore–Penrose pseudoinverse in place of $D_{\mathrm{sup}}^{-1}$ or $D_0^{-1}$.)

(2) Lifting (base → hierarchical supervertices). Define the lifted supervertex feature matrix by normalized pooling:

$$\widetilde{X} \;:=\; D_{\mathrm{sup}}^{-1} B^{\top} X \;\in\; \mathbb{R}^{p \times d}.$$

(3) Superedge incidence and normalized propagation on hierarchical supervertices. Define the incidence matrix $H \in \{0,1\}^{p \times q}$ by

$$H_{aj} := \begin{cases} 1, & v_a \in \varepsilon_j, \\ 0, & \text{otherwise.} \end{cases}$$

Let $w : E \to (0, \infty)$ be a superedge-weight function, and define

$$W := \mathrm{diag}\big(w(\varepsilon_1), \dots, w(\varepsilon_q)\big).$$

Define the degree matrices

$$(D_{\mathcal{V}})_{aa} := \sum_{j=1}^{q} H_{aj}\, w(\varepsilon_j), \qquad (D_{\mathcal{E}})_{jj} := \sum_{a=1}^{p} H_{aj} = |\varepsilon_j|.$$

The normalized propagation operator is

$$S_{\mathrm{H}} := D_{\mathcal{V}}^{-1/2}\, H\, W\, D_{\mathcal{E}}^{-1}\, H^{\top}\, D_{\mathcal{V}}^{-1/2} \;\in\; \mathbb{R}^{p \times p}.$$

(4) Convolution on the hierarchical superhypergraph domain. Given learnable parameters $\Theta \in \mathbb{R}^{d\times c}$ and an activation $\sigma$, define

$$\widetilde{Y} \;:=\; \sigma\Big(S_{\mathrm{H}}\,\widetilde{X}\,\Theta\Big) \;\in\; \mathbb{R}^{p\times c}.$$

(5) Optional projection (hierarchical supervertices $\to$ base). If a base-level output is required, define

$$Y \;:=\; D_0^{-1} B\,\widetilde{Y} \;\in\; \mathbb{R}^{N\times c}.$$

A multi-layer HSHGNN is obtained by iterating Step (4) with layer-wise parameters $\Theta^{(\ell)}$, optionally followed by Step (5).

Remark 3.10.3 (Coherent hierarchical case). If, in addition, $(V, E)$ satisfies the downward-closure condition

$$X \in V,\ \ell(X) \geq 1 \implies X \subseteq V,$$

then $\mathbb{H}^{\langle r\rangle}$ is a (coherent) hierarchical SuperHyperGraph in the usual sense. Definition 3.10.2 remains unchanged.

We now show that HSHGNN contains the usual SHGNN as a special case.

Theorem 3.10.4 (HSHGNN generalizes SHGNN). Fix $n \in \mathbb{N}_0$, and let

$$\mathsf{H}^{(n)} = (V, E)$$

be an $n$-SuperHyperGraph over $V_0$, with

$$V = \{v_1, \ldots, v_p\} \subseteq \mathcal{P}^n(V_0) \setminus \{\emptyset\}, \qquad E = \{\varepsilon_1, \ldots, \varepsilon_q\}, \qquad \varepsilon_j \in \mathcal{P}(V) \setminus \{\emptyset\}.$$

Consider the canonical SHGNN layer on $\mathsf{H}^{(n)}$, defined by:

- base–supervertex lifting via the recursive support map on $\mathcal{P}^n(V_0)$, and
- normalized hypergraph propagation on the supervertex domain $V$.

Then this SHGNN layer is exactly the HSHGNN layer (Definition 3.10.2) with height $r = n$, when $\mathsf{H}^{(n)}$ is viewed as a height-$n$ hierarchical domain (i.e., $V \subseteq \mathcal{U}_n(V_0)$). More precisely, all matrices and outputs coincide:

$$B_{\mathrm{HSHGNN}} = B_{\mathrm{SHGNN}}, \qquad S_{\mathrm{H}} = S_{\mathrm{SHG}}, \qquad \widetilde{Y}_{\mathrm{HSHGNN}} = \widetilde{Y}_{\mathrm{SHGNN}},$$

and, if Step (5) is used,

$$Y_{\mathrm{HSHGNN}} = Y_{\mathrm{SHGNN}}.$$

Proof. Since $V \subseteq \mathcal{P}^n(V_0) \setminus \{\emptyset\}$, we also have

$$V \subseteq \mathcal{U}_n(V_0) = \bigcup_{m=0}^{n} \big(\mathcal{P}^m(V_0) \setminus \{\emptyset\}\big).$$

Hence the same pair $(V, E)$ is admissible as a height-$n$ hierarchical superhypergraph domain.

The HSHGNN membership matrix is defined by

$$(B_{\mathrm{HSHGNN}})_{ia} = 1 \iff x_i \in \mathrm{Supp}(v_a),$$

where Supp is the recursive support map on $\mathcal{U}_n(V_0)$. Restricted to $V \subseteq \mathcal{P}^n(V_0)$, this is exactly the same recursive support rule used in the canonical SHGNN. Therefore,

$$B_{\mathrm{HSHGNN}} = B_{\mathrm{SHGNN}}.$$

Consequently, the lifting operators coincide:

$$\widetilde{X}_{\mathrm{HSHGNN}} = D_{\sup}^{-1} B^{\top} X = \widetilde{X}_{\mathrm{SHGNN}}.$$

Next, both models use the same supervertex set $V$, the same superedge family $E$, and the same superedge weights. Hence their incidence matrices are identical:

$$H_{\text{HSHGNN}} = H_{\text{SHGNN}}.$$

Therefore the corresponding degree matrices also coincide:

$$D_{\mathcal{V}}^{(\text{HSHGNN})} = D_{\mathcal{V}}^{(\text{SHGNN})}, \qquad D_{\mathcal{E}}^{(\text{HSHGNN})} = D_{\mathcal{E}}^{(\text{SHGNN})}.$$

It follows that the normalized propagation operators are equal:

$$S_{\text{H}} = D_{\mathcal{V}}^{-1/2} H W D_{\mathcal{E}}^{-1} H^{\top} D_{\mathcal{V}}^{-1/2} = S_{\text{SHG}}.$$

Using the same learnable matrix $\Theta$ and activation $\sigma$, we obtain

$$\widetilde{Y}_{\text{HSHGNN}} = \sigma\Big(S_{\text{H}} \widetilde{X} \Theta\Big) = \sigma\Big(S_{\text{SHG}} \widetilde{X} \Theta\Big) = \widetilde{Y}_{\text{SHGNN}}.$$

Finally, if the optional base projection is applied, it uses the same matrices $B$ and $D_0$, so

$$Y_{\text{HSHGNN}} = D_0^{-1} B \widetilde{Y}_{\text{HSHGNN}} = D_0^{-1} B \widetilde{Y}_{\text{SHGNN}} = Y_{\text{SHGNN}}.$$

Thus the SHGNN layer is a special case of the HSHGNN layer. □

Theorem 3.10.5 (Strictness of the generalization). HSHGNN is a strict generalization of SHGNN (for fixed-level supervertex models).

Proof. We exhibit a valid hierarchical-domain instance that cannot be represented as an $n$-SHGNN for any fixed $n$.

Let $V_0 = \{x_1, x_2\}$, and consider

$$V = \{x_1,\ x_2,\ \{x_1, x_2\}\}.$$

Then $x_1, x_2 \in \mathcal{P}^0(V_0)$ (level 0), while $\{x_1, x_2\} \in \mathcal{P}^1(V_0)$ (level 1). Hence

$$V \subseteq \mathcal{U}_1(V_0),$$

so $V$ is a valid mixed-level hierarchical supervertex set. Let

$$E = \big\{\{x_1, \{x_1, x_2\}\}\big\},$$

which is a valid superedge family on $V$. Therefore, Definition 3.10.2 yields a valid HSHGNN layer.

Suppose, for contradiction, that this same vertex domain $V$ could be represented as an $n$-SHGNN for some fixed $n$. Then one would require

$$V \subseteq \mathcal{P}^n(V_0).$$

But $x_1 \in \mathcal{P}^0(V_0)$ and $\{x_1, x_2\} \in \mathcal{P}^1(V_0)$, so $V$ contains vertices from two distinct levels. Hence no single $n$ can satisfy $V \subseteq \mathcal{P}^n(V_0)$. This contradiction shows that the instance is not representable in the fixed-level SHGNN framework.

Therefore, HSHGNN strictly generalizes SHGNN. □

**Algorithm 1:** One-layer 1-SuperHypergraph Neural Network (1-SHGNN) convolution

Input: Base vertex set $V_0 = \{x_1, \ldots, x_n\}$;
1-SuperHypergraph $\mathsf{H} = (\mathcal{V}, \mathcal{E})$ over $V_0$, where $\mathcal{V} = \{S_1, \ldots, S_p\} \subseteq \mathcal{P}^*(V_0)$ and $\mathcal{E} = \{\varepsilon_1, \ldots, \varepsilon_q\} \subseteq \mathcal{P}^*(\mathcal{V})$;
Base feature matrix $X \in \mathbb{R}^{n\times d}$;
Superedge weights $w(\varepsilon_j) > 0$ for $j = 1, \ldots, q$;
Learnable weight matrix $\Theta \in \mathbb{R}^{d\times c}$; activation $\sigma$;
(Optional) a flag indicating whether to return base-vertex outputs.
Output: Supervertex output $\widetilde{Y} \in \mathbb{R}^{p\times c}$, and optionally base-vertex output $Y \in \mathbb{R}^{n\times c}$.

(A) Construct membership matrix $B \in \{0,1\}^{n\times p}$
Initialize sparse $B \leftarrow 0$
/* For each supervertex, mark its base members. */
for $a \leftarrow 1$ to $p$ do
  foreach $x_i \in S_a$ do
    $B_{ia} \leftarrow 1$
  end
end

(B) Compute diagonal normalizers $D_S$ and $D_0$
for $a \leftarrow 1$ to $p$ do
  $(D_S)_{aa} \leftarrow \sum_{i=1}^{n} B_{ia}$ // $= |S_a|$
end
for $i \leftarrow 1$ to $n$ do
  $(D_0)_{ii} \leftarrow \sum_{a=1}^{p} B_{ia}$
end
/* If some diagonal entries are zero, use pseudoinverses below. */

(C) Lift base features to supervertex features
$\widetilde{X} \leftarrow D_S^{-1} B^\top X$

(D) Construct incidence matrix $H \in \{0,1\}^{p\times q}$
Initialize sparse $H \leftarrow 0$
/* For each superedge, mark its incident supervertices. */
for $j \leftarrow 1$ to $q$ do
  foreach $S_a \in \varepsilon_j$ do
    $H_{aj} \leftarrow 1$
  end
end

(E) Compute degree matrices and weights
$W \leftarrow \mathrm{diag}\big(w(\varepsilon_1), \ldots, w(\varepsilon_q)\big)$
for $a \leftarrow 1$ to $p$ do
  $(D_{\mathcal{V}})_{aa} \leftarrow \sum_{j=1}^{q} H_{aj}\, w(\varepsilon_j)$
end
for $j \leftarrow 1$ to $q$ do
  $(D_{\mathcal{E}})_{jj} \leftarrow \sum_{a=1}^{p} H_{aj}$ // $= |\varepsilon_j|$
end

(F) 1-SHGNN convolution on supervertices (no explicit $S$)
$Z \leftarrow D_{\mathcal{V}}^{-1/2}\, \widetilde{X}$
$U \leftarrow H^\top Z$ // sparse–dense
$U \leftarrow D_{\mathcal{E}}^{-1}\, W\, U$ // diagonal scales
$R \leftarrow H\, U$ // sparse–dense
$R \leftarrow D_{\mathcal{V}}^{-1/2} R$
$\widetilde{Y} \leftarrow \sigma\big(R\,\Theta\big)$

(G) Optional projection back to base vertices
if base output is required then
  $Y \leftarrow D_0^{-1} B\, \widetilde{Y}$
  return $(\widetilde{Y}, Y)$
end
else
  return $\widetilde{Y}$
end

# 4 Review and Results: Revisiting Definitions for SHGNN

This chapter revisits several basic objects that are routinely used in the analysis of SuperHyperGraph Neural Networks (SHGNNs). In particular, we formulate the SuperHyperGraph Laplacian, SuperHyperGraph convolution, SuperHyperGraph clustering, and SuperHyperGraph degree centrality in a way that is (i) mathematically unambiguous and (ii) explicitly reduces to the corresponding hypergraph notions under the natural "no-superstructure" specialization. We also include a precise attention mechanism for $n$-SuperHyperGraphs.

## 4.1 SuperHyperGraph Laplacian

For orientation, recall that a graph Laplacian is a matrix representation of a graph that encodes connectivity and supports spectral analysis (cf. [216, 217]). We begin with the standard normalized hypergraph Laplacian and then define its superhypergraph analogue via expansion to the base vertex set.

Definition 4.1.1 (Hypergraph Laplacian). (cf. [198, 197]) Let $\mathcal{H} = (\mathcal{V}, \mathcal{E})$ be a finite weighted hypergraph with $|\mathcal{V}| = n$ and $|\mathcal{E}| = m$. Let $w : \mathcal{E} \to \mathbb{R}_{>0}$ be a hyperedge-weight function, and let $W \in \mathbb{R}^{m\times m}$ be the diagonal matrix with $W_{jj} = w(e_j)$.

Define the incidence matrix $B \in \mathbb{R}^{n\times m}$ by

$$B_{ij} = \begin{cases} 1, & v_i \in e_j, \\ 0, & \text{otherwise.} \end{cases}$$

Define the diagonal vertex-degree matrix $D_v \in \mathbb{R}^{n\times n}$ and hyperedge-degree matrix $D_e \in \mathbb{R}^{m\times m}$ by

$$(D_v)_{ii} = d_v(v_i) := \sum_{j=1}^{m} B_{ij}\, w(e_j), \qquad (D_e)_{jj} = d_e(e_j) := \sum_{i=1}^{n} B_{ij} = |e_j|.$$

Assume $d_v(v_i) > 0$ for all $i$ (otherwise replace $D_v^{-1/2}$ by a generalized inverse). The (normalized) hypergraph Laplacian is

$$L_{\mathcal{H}} \;:=\; I_n - D_v^{-1/2}\, B\, W\, D_e^{-1}\, B^\top\, D_v^{-1/2}.$$

Definition 4.1.2 (Expanded hypergraph of a SuperHyperGraph). Let $H = (V, E)$ be a SuperHyperGraph with base vertex set $V_0$, where each supervertex $v \in V$ is a (nonempty) subset of $V_0$. Define the expanded hypergraph $\mathcal{H}^{\uparrow} = (V_0, E^{\uparrow})$ as follows.

- The vertex set is $V_0$.
- The hyperedge family $E^{\uparrow}$ is indexed by $E$: for each superedge $e \in E$, define the expanded hyperedge

  $$e^{\uparrow} \;:=\; \bigcup_{v\in e} v \;\subseteq\; V_0,$$

  and include one copy of $e^{\uparrow}$ for each $e \in E$. (Thus distinct $e_1 \neq e_2$ may yield identical subsets $e_1^{\uparrow} = e_2^{\uparrow}$ but are still treated as distinct indexed hyperedges.)

Remark 4.1.3 (Two expanded models: multiset vs. merged-weight). Definition 4.1.2 yields an indexed (multiset) expanded hypergraph: different superhyperedges may induce the same expanded subset of $V_0$, but are kept as distinct hyperedges because they carry different indices. For some constructions (e.g. partitioning or compression), it is convenient to merge duplicate expanded hyperedges and store their multiplicity/strength as a weight. We formalize this merged-weight variant below and relate it to the indexed model by a quotient construction.

Definition 4.1.4 (Merged-weight expanded hypergraph). Let $V_0$ be a finite nonempty base set and let $\mathbb{H}^{(n)} = (V_n, E_n)$ be an $n$-superhypergraph on $V_0$. Let $\mathbb{H}^{\mathrm{ms}}_{\mathrm{exp}} = (V_0, E^{\mathrm{ms}}_{\mathrm{exp}})$ be the indexed (multiset) expanded hypergraph given by Definition 4.1.2, and let

$$u: \; E^{\mathrm{ms}}_{\mathrm{exp}} \longrightarrow \mathcal{P}^{*}(V_0)$$

be the underlying-subset map sending each indexed expanded hyperedge $e^{\uparrow} \in E^{\mathrm{ms}}_{\mathrm{exp}}$ to its underlying subset $u(e^{\uparrow}) \subseteq V_0$ (i.e. the expansion of the corresponding superhyperedge).

Assume moreover that the indexed expanded hyperedges are equipped with a weight

$$w^{\mathrm{ms}}: \; E^{\mathrm{ms}}_{\mathrm{exp}} \longrightarrow \mathbb{R}_{>0},$$

where one may take $w^{\mathrm{ms}} \equiv 1$ if no additional information is intended.

Define the merged edge set

$$\overline{E}_{\mathrm{exp}} \;:=\; u\big(E^{\mathrm{ms}}_{\mathrm{exp}}\big) \;\subseteq\; \mathcal{P}^{*}(V_0),$$

and the merged weight $\overline{w} : \overline{E}_{\mathrm{exp}} \to \mathbb{R}_{>0}$ by the pushforward rule

$$\overline{w}(S) \;:=\; \sum_{\substack{e^{\uparrow} \in E^{\mathrm{ms}}_{\mathrm{exp}} \\ u(e^{\uparrow}) = S}} w^{\mathrm{ms}}(e^{\uparrow}) \qquad (S \in \overline{E}_{\mathrm{exp}}).$$

The merged-weight expanded hypergraph is the weighted hypergraph

$$\mathbb{H}^{\mathrm{mw}}_{\mathrm{exp}} \;:=\; (V_0, \overline{E}_{\mathrm{exp}}, \overline{w}).$$

Proposition 4.1.5 (Quotient relation between the two expansions). With notation as in Definition 4.1.4, define an equivalence relation $\sim$ on $E^{\mathrm{ms}}_{\mathrm{exp}}$ by

$$e^{\uparrow} \sim f^{\uparrow} \quad \Longleftrightarrow \quad u(e^{\uparrow}) = u(f^{\uparrow}).$$

Then:

(i) The canonical map

$$\pi: \; E^{\mathrm{ms}}_{\mathrm{exp}} \longrightarrow \overline{E}_{\mathrm{exp}}, \qquad \pi(e^{\uparrow}) = u(e^{\uparrow}),$$

is surjective and induces a bijection

$$E^{\mathrm{ms}}_{\mathrm{exp}}/\!\sim \;\cong\; \overline{E}_{\mathrm{exp}}.$$

(ii) The merged weight $\overline{w}$ is the pushforward of $w^{\mathrm{ms}}$ along $\pi$ in the sense that, for every $S \in \overline{E}_{\mathrm{exp}}$,

$$\overline{w}(S) \;=\; \sum_{e^{\uparrow} \in \pi^{-1}(S)} w^{\mathrm{ms}}(e^{\uparrow}).$$

In particular, if $w^{\mathrm{ms}} \equiv 1$, then $\overline{w}(S) = \#\pi^{-1}(S)$ equals the multiplicity of $S$ in the indexed (multiset) model.

Proof. (i) By definition, $\overline{E}_{\mathrm{exp}} = u(E^{\mathrm{ms}}_{\mathrm{exp}})$, so $\pi = u$ is surjective onto $\overline{E}_{\mathrm{exp}}$. Moreover, $\sim$ is precisely the kernel relation of $\pi$, i.e. $e^{\uparrow} \sim f^{\uparrow}$ iff $\pi(e^{\uparrow}) = \pi(f^{\uparrow})$. Hence $\pi$ identifies exactly the equivalence classes of $\sim$, and the universal property of quotients yields a bijection $E^{\mathrm{ms}}_{\mathrm{exp}}/\sim \;\cong\; \overline{E}_{\mathrm{exp}}$.

(ii) This is immediate from the definition of $\overline{w}$ as the sum of weights over each fiber $\pi^{-1}(S)$. □

Definition 4.1.6 (SuperHyperGraph Laplacian). Let $H = (V, E)$ be a SuperHyperGraph with base vertex set $V_0$, and let $\mathcal{H}^{\uparrow} = (V_0, E^{\uparrow})$ be its expanded hypergraph (Definition 4.1.2). Fix a weight function $w : E \to \mathbb{R}_{>0}$ on superedges and transfer it to expanded hyperedges by $w(e^{\uparrow}) := w(e)$ for the copy indexed by $e \in E$. Let $m := |E|$ (counted with indices).

Let $B^{\uparrow} \in \mathbb{R}^{|V_0| \times m}$ be the incidence matrix of $\mathcal{H}^{\uparrow}$,

$$(B^{\uparrow})_{ij} = \begin{cases} 1, & v_i \in e_j^{\uparrow}, \\ 0, & \text{otherwise}, \end{cases}$$

and define the diagonal matrices $D_V$ and $D_E$ by

$$(D_V)_{ii} := \sum_{j=1}^{m} (B^{\uparrow})_{ij}\, w(e_j^{\uparrow}), \qquad (D_E)_{jj} := \sum_{i=1}^{|V_0|} (B^{\uparrow})_{ij} = |e_j^{\uparrow}|.$$

Let $W \in \mathbb{R}^{m \times m}$ be diagonal with $W_{jj} = w(e_j^{\uparrow})$, and assume $D_V$ is invertible (or use a generalized inverse). The SuperHyperGraph Laplacian is

$$L_H \ := \ I_{|V_0|} - D_V^{-1/2}\, B^{\uparrow}\, W\, D_E^{-1}\, (B^{\uparrow})^{\top}\, D_V^{-1/2}.$$

Example 4.1.7 (SuperHyperGraph Laplacian for a small 1-SuperHyperGraph). Let the base vertex set be

$$V_0 = \{x_1, x_2, x_3\},$$

and consider the 1-SuperHyperGraph $H = (V, E)$ with supervertices

$$v_1 = \{x_1, x_2\}, \qquad v_2 = \{x_2, x_3\}, \qquad v_3 = \{x_3\},$$

so that

$$V = \{v_1, v_2, v_3\}.$$

Define two superedges:

$$e_1 = \{v_1, v_2\}, \qquad e_2 = \{v_2, v_3\}, \qquad E = \{e_1, e_2\}.$$

By expansion to the base level (Definition 4.1.2), each superedge $e_j$ induces a hyperedge $e_j^{\uparrow} \subseteq V_0$ obtained by taking the union of the base supports of the supervertices contained in $e_j$. Hence

$$e_1^{\uparrow} = v_1 \cup v_2 = \{x_1, x_2, x_3\}, \qquad e_2^{\uparrow} = v_2 \cup v_3 = \{x_2, x_3\}.$$

Thus the expanded hypergraph is

$$\mathcal{H}^{\uparrow} = (V_0, E^{\uparrow}), \qquad E^{\uparrow} = \{e_1^{\uparrow}, e_2^{\uparrow}\}.$$

Step 1: Choose superedge weights. Let

$$w(e_1) = 2, \qquad w(e_2) = 1.$$

By transfer of weights,

$$w(e_1^{\uparrow}) = 2, \qquad w(e_2^{\uparrow}) = 1.$$

Step 2: Incidence matrix of the expanded hypergraph. Ordering rows by $(x_1, x_2, x_3)$ and columns by $(e_1^{\uparrow}, e_2^{\uparrow})$, we obtain

$$B^{\uparrow} = \begin{bmatrix} 1 & 0 \\ 1 & 1 \\ 1 & 1 \end{bmatrix},$$

since $x_1 \in e_1^{\uparrow}$ only, while $x_2, x_3 \in e_1^{\uparrow} \cap e_2^{\uparrow}$.

Step 3: Degree matrices $D_V$, $D_E$, and weight matrix $W$. The hyperedge-size matrix is

$$D_E = \mathrm{diag}\big(|e_1^{\uparrow}|, |e_2^{\uparrow}|\big) = \mathrm{diag}(3, 2).$$

The weight matrix is

$$W = \mathrm{diag}(2, 1).$$

The weighted base-vertex degrees are

$$(D_V)_{ii} = \sum_{j=1}^{2} (B^{\uparrow})_{ij}\, w(e_j^{\uparrow}),$$

hence

$$D_V = \mathrm{diag}(2, 3, 3),$$

because

$$\deg(x_1) = 2, \qquad \deg(x_2) = 2 + 1 = 3, \qquad \deg(x_3) = 2 + 1 = 3.$$

Step 4: Normalized adjacency term. Set

$$P := D_V^{-1/2} B^{\uparrow} W D_E^{-1} (B^{\uparrow})^{\top} D_V^{-1/2}.$$

First,

$$W D_E^{-1} = \mathrm{diag}\left(\frac{2}{3}, \frac{1}{2}\right).$$

Therefore,

$$B^{\uparrow} W D_E^{-1} (B^{\uparrow})^{\top} = \begin{bmatrix} \frac{2}{3} & \frac{2}{3} & \frac{2}{3} \\ \frac{2}{3} & \frac{7}{6} & \frac{7}{6} \\ \frac{2}{3} & \frac{7}{6} & \frac{7}{6} \end{bmatrix}.$$

Since

$$D_V^{-1/2} = \mathrm{diag}\left(\frac{1}{\sqrt{2}}, \frac{1}{\sqrt{3}}, \frac{1}{\sqrt{3}}\right),$$

we obtain

$$P = \begin{bmatrix} \frac{1}{3} & \frac{2}{3\sqrt{6}} & \frac{2}{3\sqrt{6}} \\ \frac{2}{3\sqrt{6}} & \frac{7}{18} & \frac{7}{18} \\ \frac{2}{3\sqrt{6}} & \frac{7}{18} & \frac{7}{18} \end{bmatrix}.$$

Step 5: SuperHyperGraph Laplacian. By Definition 4.1.6,

$$L_H = I_3 - P.$$

Hence

$$L_H = \begin{bmatrix} \frac{2}{3} & -\frac{2}{3\sqrt{6}} & -\frac{2}{3\sqrt{6}} \\ -\frac{2}{3\sqrt{6}} & \frac{11}{18} & -\frac{7}{18} \\ -\frac{2}{3\sqrt{6}} & -\frac{7}{18} & \frac{11}{18} \end{bmatrix}.$$

This matrix is the normalized Laplacian of the expanded hypergraph induced by the SuperHyperGraph $H$, and it can be used for spectral analysis, diffusion, and Laplacian-based learning on the base vertex set $V_0$.

Theorem 4.1.8 (Reduction to the hypergraph Laplacian). The SuperHyperGraph Laplacian $L_H$ reduces to the hypergraph Laplacian $L_{\mathcal{H}}$ when $H$ encodes an ordinary hypergraph on $V_0$ (i.e., all supervertices are singletons and each superedge is a set of such singletons).

Proof. Assume $V = \{\{v\} : v \in V_0\}$ and each superedge $e \in E$ is a set of singleton supervertices. Then for every $e \in E$,

$$e^{\uparrow} = \bigcup_{v \in e} v = \bigcup_{\{u\} \in e} \{u\} = \{u \in V_0 : \{u\} \in e\},$$

so the expanded hypergraph $\mathcal{H}^{\uparrow} = (V_0, E^{\uparrow})$ is precisely the underlying hypergraph on $V_0$ (with the same edge indices and weights). Hence $B^{\uparrow}$ equals the usual incidence matrix $B$, and $D_V, D_E, W$ coincide with $D_v, D_e, W$. Substituting into Definition 4.1.6 yields $L_H = L_{\mathcal{H}}$. □

## 4.2 SuperHyperGraph Convolution

Graph convolution aggregates node features through the graph structure (cf. [218, 219, 220]). We treat hypergraph and superhypergraph convolution in a unified way through the expanded hypergraph.

Definition 4.2.1 (Hypergraph convolution). (cf. [199, 200]) Let $\mathcal{H} = (\mathcal{V}, \mathcal{E})$ be a weighted hypergraph with incidence matrix $B$, degree matrices $D_v, D_e$, and diagonal weight matrix $W$ as in the hypergraph Laplacian definition. Given input features $X \in \mathbb{R}^{|\mathcal{V}| \times d}$, parameters $\Theta \in \mathbb{R}^{d \times c}$, and an activation $\sigma$, the (normalized) hypergraph convolution is

$$Y \;=\; \sigma\Big(D_v^{-1/2}\, B\, W\, D_e^{-1}\, B^{\top}\, D_v^{-1/2}\, X\, \Theta\Big).$$

Definition 4.2.2 (SuperHyperGraph convolution). Let $H = (V, E)$ be a SuperHyperGraph with base vertex set $V_0$, and let $\mathcal{H}^{\uparrow} = (V_0, E^{\uparrow})$ be its expanded hypergraph. Let $B^{\uparrow}, D_V, D_E, W$ be as in Definition 4.1.6. Given base-vertex features $X \in \mathbb{R}^{|V_0| \times d}$, parameters $\Theta \in \mathbb{R}^{d \times c}$, and an activation $\sigma$, the SHG convolution is

$$Y \;=\; \sigma\Big(D_V^{-1/2}\, B^{\uparrow}\, W\, D_E^{-1}\, (B^{\uparrow})^{\top}\, D_V^{-1/2}\, X\, \Theta\Big).$$

Example 4.2.3 (SuperHyperGraph convolution on a small 1-SuperHyperGraph). We give a concrete computation of the SuperHyperGraph convolution from Definition 4.2.2.

Let the base vertex set be

$$V_0 = \{x_1, x_2, x_3\},$$

and consider the 1-SuperHyperGraph $H = (V, E)$ with supervertices

$$v_1 = \{x_1, x_2\}, \qquad v_2 = \{x_2, x_3\}, \qquad v_3 = \{x_3\},$$

and superedges

$$e_1 = \{v_1, v_2\}, \qquad e_2 = \{v_2, v_3\}.$$

Step 1: Expanded hypergraph $\mathcal{H}^{\uparrow} = (V_0, E^{\uparrow})$. The expanded hyperedges are

$$e_1^{\uparrow} = v_1 \cup v_2 = \{x_1, x_2, x_3\}, \qquad e_2^{\uparrow} = v_2 \cup v_3 = \{x_2, x_3\}.$$

Thus

$$E^{\uparrow} = \{e_1^{\uparrow}, e_2^{\uparrow}\}.$$

Step 2: Incidence and degree data. Choose superedge weights

$$w(e_1) = 2, \qquad w(e_2) = 1,$$

and transfer them to $E^{\uparrow}$.

Ordering rows by $(x_1, x_2, x_3)$ and columns by $(e_1^{\uparrow}, e_2^{\uparrow})$, the incidence matrix is

$$B^{\uparrow} = \begin{bmatrix} 1 & 0 \\ 1 & 1 \\ 1 & 1 \end{bmatrix}.$$

Hence

$$W = \mathrm{diag}(2,1), \qquad D_E = \mathrm{diag}(3,2), \qquad D_V = \mathrm{diag}(2,3,3).$$

Therefore

$$D_V^{-1/2} = \mathrm{diag}\Big(\frac{1}{\sqrt{2}}, \frac{1}{\sqrt{3}}, \frac{1}{\sqrt{3}}\Big).$$

Step 3: Normalized propagation matrix. Set

$$P_H := D_V^{-1/2}\, B^{\uparrow}\, W\, D_E^{-1}\, (B^{\uparrow})^{\top}\, D_V^{-1/2}.$$

A direct computation gives

$$P_H = \begin{bmatrix} \frac{1}{3} & \frac{2}{3\sqrt{6}} & \frac{2}{3\sqrt{6}} \\ \frac{2}{3\sqrt{6}} & \frac{7}{18} & \frac{7}{18} \\ \frac{2}{3\sqrt{6}} & \frac{7}{18} & \frac{7}{18} \end{bmatrix}.$$

Step 4: Input features and trainable parameters. Let the base-vertex feature matrix be

$$X = \begin{bmatrix} 1 & 0 \\ 0 & 1 \\ 1 & 1 \end{bmatrix} \in \mathbb{R}^{3\times 2},$$

and choose

$$\Theta = \begin{bmatrix} 1 & -1 \\ 0 & 1 \end{bmatrix} \in \mathbb{R}^{2\times 2}, \qquad \sigma = \mathrm{ReLU}.$$

Step 5: Convolution output. By Definition 4.2.2,

$$Y = \sigma(P_H X \Theta).$$

First,

$$P_H X = \begin{bmatrix} \frac{1}{3} + \frac{2}{3\sqrt{6}} & \frac{4}{3\sqrt{6}} \\ \frac{2}{3\sqrt{6}} + \frac{7}{18} & \frac{7}{9} \\ \frac{2}{3\sqrt{6}} + \frac{7}{18} & \frac{7}{9} \end{bmatrix}.$$

Multiplying by $\Theta$, we obtain

$$P_H X \Theta = \begin{bmatrix} \frac{1}{3} + \frac{2}{3\sqrt{6}} & \frac{2}{3\sqrt{6}} - \frac{1}{3} \\ \frac{2}{3\sqrt{6}} + \frac{7}{18} & \frac{7}{18} - \frac{2}{3\sqrt{6}} \\ \frac{2}{3\sqrt{6}} + \frac{7}{18} & \frac{7}{18} - \frac{2}{3\sqrt{6}} \end{bmatrix}.$$

Since

$$\frac{2}{3\sqrt{6}} - \frac{1}{3} < 0 \qquad \text{and} \qquad \frac{7}{18} - \frac{2}{3\sqrt{6}} > 0,$$

the ReLU activation yields

$$Y = \begin{bmatrix} \frac{1}{3} + \frac{2}{3\sqrt{6}} & 0 \\ \frac{2}{3\sqrt{6}} + \frac{7}{18} & \frac{7}{18} - \frac{2}{3\sqrt{6}} \\ \frac{2}{3\sqrt{6}} + \frac{7}{18} & \frac{7}{18} - \frac{2}{3\sqrt{6}} \end{bmatrix}.$$

Numerically,

$$Y \approx \begin{bmatrix} 0.6055 & 0 \\ 0.6611 & 0.1167 \\ 0.6611 & 0.1167 \end{bmatrix}.$$

This example illustrates how the SHG convolution performs weighted, normalized diffusion on the expanded hypergraph $\mathcal{H}^{\uparrow}$, while still originating from the higher-order SuperHyperGraph structure $H = (V, E)$.

Theorem 4.2.4 (Reduction to hypergraph convolution). The SuperHyperGraph convolution in Definition 4.2.2 reduces to the hypergraph convolution when $H$ encodes an ordinary hypergraph on $V_0$.

Proof. Under the same singleton encoding as in Theorem 4.1.8, the expanded hypergraph equals the original hypergraph, so $B^{\uparrow} = B$ and $D_V = D_v, D_E = D_e$. Substituting yields the standard hypergraph convolution formula. □

## 4.3 SuperHyperGraph Clustering

Graph clustering partitions a vertex set into groups with strong within-group connectivity (cf. [221, 222, 223]). Hypergraph clustering extends this idea by measuring cuts induced by hyperedges (cf. [224, 225, 226]). For superhypergraphs we cluster the base vertices $V_0$ using the expanded hypergraph.

Definition 4.3.1 (Graph clustering). (cf. [227, 228]) Let $G = (V, E, w)$ be a finite weighted graph with $w : E \to \mathbb{R}_{>0}$. A clustering is a partition $C = \{C_1, \dots, C_k\}$ of $V$ into nonempty disjoint blocks: $\bigcup_i C_i = V$ and $C_i \cap C_j = \emptyset$ for $i \neq j$.

Definition 4.3.2 (Hypergraph normalized cut objective). (cf. [224, 225, 226]) Let $\mathcal{H} = (\mathcal{V}, \mathcal{E})$ be a weighted hypergraph with vertex degrees $d_v(\cdot)$. For a subset $S \subseteq \mathcal{V}$, define

$$\mathrm{vol}(S) := \sum_{v \in S} d_v(v), \qquad \mathrm{cut}(S, \overline{S}) := \sum_{e \in \mathcal{E}} w(e) \frac{|e \cap S| \cdot |e \cap \overline{S}|}{|e|},$$

where $\overline{S} := \mathcal{V} \setminus S$. For a partition $\mathcal{C} = \{C_1, \dots, C_k\}$, the normalized cut is

$$\mathrm{NCut}(\mathcal{C}) := \sum_{i=1}^{k} \frac{\mathrm{cut}(C_i, \overline{C_i})}{\mathrm{vol}(C_i)}.$$

Example 4.3.3 (Concrete real-world example of the hypergraph normalized cut). Consider a grocery retailer that wants to segment customers by basket co-occurrence. Let the vertices be customers

$$\mathcal{V} = \{v_1, v_2, v_3, v_4, v_5\},$$

and let hyperedges represent frequently observed joint-basket groups (customers whose baskets are highly similar within a campaign/week), with weights proportional to the number of co-occurring baskets:

$$\mathcal{E} = \{e_1, e_2, e_3\}, \quad e_1 = \{v_1, v_2, v_3\}, \ w(e_1) = 2,$$

$$e_2 = \{v_3, v_4\}, \ w(e_2) = 1, \quad e_3 = \{v_1, v_4, v_5\}, \ w(e_3) = 1.5.$$

Use the common degree choice $d_v(v) := \sum_{e \ni v} w(e)$. Then

$$d_v(v_1) = 2 + 1.5 = 3.5, \quad d_v(v_2) = 2, \quad d_v(v_3) = 2 + 1 = 3,$$

$$d_v(v_4) = 1 + 1.5 = 2.5, \quad d_v(v_5) = 1.5.$$

Suppose we test the 2-way partition $S = \{v_1, v_2, v_3\}$ and $\overline{S} = \{v_4, v_5\}$. Its volume is

$$\mathrm{vol}(S) = 3.5 + 2 + 3 = 8.5, \qquad \mathrm{vol}(\overline{S}) = 2.5 + 1.5 = 4.$$

The cut is

$$\mathrm{cut}(S, \overline{S}) = \sum_{e \in \mathcal{E}} w(e) \frac{|e \cap S| \cdot |e \cap \overline{S}|}{|e|}.$$

Compute each term:

$$e_1 : \ |e_1 \cap S| = 3, \ |e_1 \cap \overline{S}| = 0 \ \Rightarrow \ 0,$$

$$e_2 : \ |e_2 \cap S| = 1, \ |e_2 \cap \overline{S}| = 1, \ |e_2| = 2 \ \Rightarrow \ 1 \cdot \frac{1 \cdot 1}{2} = 0.5,$$

$$e_3 : \ |e_3 \cap S| = 1, \ |e_3 \cap \overline{S}| = 2, \ |e_3| = 3 \ \Rightarrow \ 1.5 \cdot \frac{1 \cdot 2}{3} = 1.0.$$

Hence $\mathrm{cut}(S, \overline{S}) = 0.5 + 1.0 = 1.5$. For the partition $\mathcal{C} = \{S, \overline{S}\}$, the normalized cut equals

$$\mathrm{NCut}(\mathcal{C}) = \frac{\mathrm{cut}(S, \overline{S})}{\mathrm{vol}(S)} + \frac{\mathrm{cut}(\overline{S}, S)}{\mathrm{vol}(\overline{S})} = \frac{1.5}{8.5} + \frac{1.5}{4} \approx 0.5515.$$

Interpreting this value: the only "crossing" co-occurrences are mediated by $e_2$ and $e_3$, while $e_1$ is entirely inside $S$; therefore the partition separates a relatively coherent customer group $\{v_1, v_2, v_3\}$ from $\{v_4, v_5\}$ with limited weighted cross-basket interaction.

Definition 4.3.4 (SuperHyperGraph clustering). Let $H = (V, E)$ be a SuperHyperGraph with base vertex set $V_0$, and let $\mathcal{H}^{\uparrow} = (V_0, E^{\uparrow})$ be its expanded hypergraph (Definition 4.1.2) with induced weights. A SuperHyperGraph clustering is a partition $\mathcal{C} = \{C_1, \dots, C_k\}$ of the base vertex set $V_0$. Its quality can be measured by applying any hypergraph clustering criterion to $\mathcal{H}^{\uparrow}$, for example the normalized cut $\mathrm{NCut}(\mathcal{C})$ defined above (with $\mathcal{V} = V_0$ and $\mathcal{E} = E^{\uparrow}$).

In particular, spectral SHG clustering is obtained by computing eigenvectors of the SuperHyperGraph Laplacian $L_H$ (Definition 4.1.6) and then clustering the resulting spectral embeddings of vertices in $V_0$.

Example 4.3.5 (Concrete real-world example of SuperHyperGraph clustering). Consider a cybersecurity SOC that wants to cluster IP addresses into attack campaigns. Let the base vertex set be

$$V_0 = \{p_1, p_2, p_3, p_4, p_5, p_6\},$$

where each $p_i$ is a suspicious IP observed in alerts. In daily operations, analysts observe multiple alert-level hyperedges (co-occurring IPs within the same incident), and also higher-order "campaign" objects that group several alert-hyperedges together. This motivates a depth-1 SuperHyperGraph $H = (V, E)$ in which each superhyperedge is a set of hyperedges on $V_0$.

For instance, suppose the SOC has three alert-hyperedges

$$e_a = \{p_1, p_2, p_3\}, \qquad e_b = \{p_3, p_4\}, \qquad e_c = \{p_4, p_5, p_6\},$$

and two campaign-level superhyperedges (each collecting the alert-hyperedges attributed to the same campaign):

$$E_1 = \{e_a, e_b\} \quad \text{(campaign A)}, \qquad E_2 = \{e_c\} \quad \text{(campaign B)}.$$

Assign campaign weights representing severity or frequency, e.g.,

$$w_H(E_1) = 5, \qquad w_H(E_2) = 3.$$

The expanded hypergraph $\mathcal{H}^{\uparrow} = (V_0, E^{\uparrow})$ (Definition 4.1.2) contains the induced base-level hyperedges

$$E^{\uparrow} = \{e_a, e_b, e_c\},$$

with induced weights, for example by summing campaign weights over all superhyperedges containing the given alert-hyperedge:

$$w^{\uparrow}(e) := \sum_{E \in E: \ e \in E} w_H(E), \quad \text{so that} \quad w^{\uparrow}(e_a) = 5, \ w^{\uparrow}(e_b) = 5, \ w^{\uparrow}(e_c) = 3.$$

A SuperHyperGraph clustering is a partition $\mathcal{C} = \{C_1, C_2\}$ of $V_0$. For example, one candidate partition is

$$C_1 = \{p_1, p_2, p_3, p_4\}, \qquad C_2 = \{p_5, p_6\},$$

which hypothesizes that $p_1, p_2, p_3, p_4$ belong to one campaign (A) and $p_5, p_6$ to another (B). Its quality can be evaluated by applying a hypergraph objective (e.g. NCut) to $\mathcal{H}^{\uparrow}$: intuitively, $e_a$ and $e_b$ lie inside $C_1$, while $e_c$ lies inside $C_2 \cup \{p_4\}$ and crosses the cut only through $p_4$, so the only "bridge" evidence between campaigns is the shared IP $p_4$.

In spectral SHG clustering, one computes eigenvectors of the SuperHyperGraph Laplacian $L_H$ (Definition 4.1.6) to obtain a low-dimensional embedding $z(p_i)$ for each base vertex $p_i \in V_0$, and then runs a standard clustering method (e.g. $k$-means) on $\{z(p_i)\}_{i=1}^{6}$. Operationally, this produces IP clusters that align with multi-alert, campaign-level structure rather than relying on pairwise similarity alone, which is useful when attacks manifest as groups of co-occurrences grouped into higher-order campaigns.

Theorem 4.3.6 (Reduction to hypergraph clustering). If $H$ encodes an ordinary hypergraph on $V_0$, then SuperHyperGraph clustering (Definition 4.3.4) reduces to standard hypergraph clustering on that hypergraph.

Proof. Under the singleton encoding, the expanded hypergraph $\mathcal{H}^{\uparrow}$ equals the original hypergraph on $V_0$, and the Laplacian $L_H$ equals the hypergraph Laplacian. Hence both objective-based and spectral procedures coincide. □

## 4.4 Degree centrality in SuperHyperGraphs

Degree centrality measures vertex importance by counting (weighted) incident connections (cf. [229, 230]). In hypergraphs and superhypergraphs it is natural to use the weighted vertex degree induced by the incidence structure.

Definition 4.4.1 (Hypergraph degree centrality). [231, 232, 233] Let $\mathcal{H} = (\mathcal{V}, \mathcal{E})$ be a weighted hypergraph with incidence matrix $B$ and hyperedge weights $w$. The degree centrality of $v_i \in \mathcal{V}$ is

$$C_{\mathcal{H}}(v_i) \;:=\; d_v(v_i) \;=\; \sum_{j=1}^{|\mathcal{E}|} B_{ij}\, w(e_j).$$

Definition 4.4.2 (SuperHyperGraph degree centrality). Let $H = (V, E)$ be a SuperHyperGraph with base vertex set $V_0$, and let $\mathcal{H}^{\uparrow} = (V_0, E^{\uparrow})$ be its expanded hypergraph with incidence matrix $B^{\uparrow}$ and weights $w(e_j^{\uparrow})$. The degree centrality of a base vertex $v_i \in V_0$ is

$$C_H(v_i) \;:=\; d_V(v_i) \;=\; \sum_{j=1}^{|E|} (B^{\uparrow})_{ij}\, w(e_j^{\uparrow}).$$

Example 4.4.3 (SuperHyperGraph degree centrality on a small 1-SuperHyperGraph). We compute the SuperHyperGraph degree centrality (Definition 4.4.2) for a simple 1-SuperHyperGraph.

Let the base vertex set be

$$V_0 = \{x_1, x_2, x_3, x_4\}.$$

Define the 1-supervertices

$$v_1 = \{x_1, x_2\}, \qquad v_2 = \{x_2, x_3\}, \qquad v_3 = \{x_3, x_4\},$$

and the superedge family

$$E = \{e_1, e_2\}, \qquad e_1 = \{v_1, v_2\}, \quad e_2 = \{v_2, v_3\}.$$

Thus $H = (V, E)$ is a 1-SuperHyperGraph, where $V = \{v_1, v_2, v_3\}$.

Step 1: Expanded hypergraph $\mathcal{H}^{\uparrow} = (V_0, E^{\uparrow})$. By the expansion rule, each superedge induces a hyperedge on $V_0$:

$$e_1^{\uparrow} = v_1 \cup v_2 = \{x_1, x_2, x_3\}, \qquad e_2^{\uparrow} = v_2 \cup v_3 = \{x_2, x_3, x_4\}.$$

Hence

$$E^{\uparrow} = \{e_1^{\uparrow}, e_2^{\uparrow}\}.$$

Step 2: Assign weights. Let the superedge weights be

$$w(e_1) = 2, \qquad w(e_2) = 1.$$

These are transferred to the expanded hyperedges:

$$w(e_1^{\uparrow}) = 2, \qquad w(e_2^{\uparrow}) = 1.$$

Step 3: Incidence matrix $B^{\uparrow}$. Ordering rows by $(x_1, x_2, x_3, x_4)$ and columns by $(e_1^{\uparrow}, e_2^{\uparrow})$, the incidence matrix is

$$B^{\uparrow} = \begin{bmatrix} 1 & 0 \\ 1 & 1 \\ 1 & 1 \\ 0 & 1 \end{bmatrix}.$$

Step 4: Degree centrality values. By Definition 4.4.2,

$$C_H(x_i) = \sum_{j=1}^{2} (B^{\uparrow})_{ij}\, w(e_j^{\uparrow}).$$

Therefore,

$$\begin{aligned} C_H(x_1) &= 1 \cdot 2 + 0 \cdot 1 = 2, \\ C_H(x_2) &= 1 \cdot 2 + 1 \cdot 1 = 3, \\ C_H(x_3) &= 1 \cdot 2 + 1 \cdot 1 = 3, \\ C_H(x_4) &= 0 \cdot 2 + 1 \cdot 1 = 1. \end{aligned}$$

Interpretation. Thus the weighted degree-centrality ranking is

$$x_2 \sim x_3 \;>\; x_1 \;>\; x_4.$$

The vertices $x_2$ and $x_3$ are the most central because they belong to both expanded hyperedges, while $x_1$ and $x_4$ appear in only one expanded hyperedge each. Moreover, $x_1$ has larger centrality than $x_4$ because it belongs to the heavier hyperedge $e_1^{\uparrow}$ (weight 2).

This example shows how SuperHyperGraph degree centrality on the base vertex set $V_0$ naturally combines:

- the higher-order structure of superedges (through expansion), and
- the importance of superedges (through the weights $w$).

Theorem 4.4.4 (Reduction to hypergraph degree). The SuperHyperGraph degree centrality in Definition 4.4.2 reduces to hypergraph degree centrality when $H$ encodes an ordinary hypergraph on $V_0$.

Proof. In the singleton encoding, the expanded incidence matrix $B^{\uparrow}$ equals the hypergraph incidence matrix $B$, and weights match. Hence the formulas coincide. □

## 4.5 $n$-SuperHyperGraph Attention

Graph attention dynamically weights neighborhood information during message passing (cf. [234, 235]). We present a standard hypergraph-attention template and then define $n$-SuperHyperGraph attention by applying attention to the expanded hypergraph on the base vertex set.

Definition 4.5.1 (Hypergraph attention (vertex-to-hyperedge incidence weighting)). [236, 200, 237] Let $\mathcal{H} = (\mathcal{V}, \mathcal{E})$ be a weighted hypergraph with incidence matrix $B$. Let $X \in \mathbb{R}^{|\mathcal{V}| \times d}$ be node features and $\Theta \in \mathbb{R}^{d\times d'}$ a shared linear map. Write $x'_i := x_i\Theta$. For each hyperedge $e_j$, define a hyperedge embedding (one common choice) by

$$u'_j \;:=\; \frac{1}{|e_j|}\sum_{v_k\in e_j} x'_k.$$

Fix a learnable vector $a \in \mathbb{R}^{2d'}$ and a nonlinearity $\eta$ (e.g., LeakyReLU). For each incidence pair $(v_i, e_j)$ with $B_{ij} = 1$, define the unnormalized score

$$s_{ij} \;:=\; \eta\big(a^\top [\, x'_i \,\|\, u'_j \,]\big),$$

and normalize over hyperedges incident to $v_i$:

$$\alpha_{ij} \;:=\; \frac{\exp(s_{ij})}{\sum_{k:\, B_{ik}=1}\exp(s_{ik})}.$$

Define an attention-weighted incidence matrix $\widetilde{B}$ by $\widetilde{B}_{ij} := \alpha_{ij}$ if $B_{ij} = 1$ and $\widetilde{B}_{ij} := 0$ otherwise. An attention-based hypergraph propagation step can then be written in the normalized form

$$X^{\text{out}} = \sigma\Big(D_v^{-1/2}\,\widetilde{B}\,W\,D_e^{-1}\,\widetilde{B}^\top\,D_v^{-1/2}\,X\,\Theta_0\Big),$$

with a learnable $\Theta_0$ and activation $\sigma$.

Definition 4.5.2 ($n$-SuperHyperGraph attention). Let $H^{(n)} = (V^{(n)}, E^{(n)})$ be an $n$-SuperHyperGraph on a base vertex set $V_0$, and let $\mathcal{H}^{(n)\uparrow} = (V_0, E^{(n)\uparrow})$ be its expanded hypergraph (with an incidence matrix $B^{(n)\uparrow}$ whose columns are indexed by superedges, as in Definition 4.1.2). Define $n$-SuperHyperGraph attention to be the hypergraph-attention construction applied to $\mathcal{H}^{(n)\uparrow}$: namely, attention coefficients $\alpha_{ij}$ are computed between base vertices $v_i \in V_0$ and expanded hyperedges $e_j^{(n)\uparrow} \in E^{(n)\uparrow}$ whenever $(B^{(n)\uparrow})_{ij} = 1$, producing an attention-weighted incidence matrix $\widetilde{B}^{(n)}$. The resulting attention-based propagation has the form

$$X^{\text{out}} = \sigma\Big((D_V^{(n)})^{-1/2}\,\widetilde{B}^{(n)}\,W^{(n)}\,(D_E^{(n)})^{-1}\,\widetilde{B}^{(n)\top}\,(D_V^{(n)})^{-1/2}\,X\,\Theta_0\Big),$$

where $D_V^{(n)}, D_E^{(n)}, W^{(n)}$ are computed from $\mathcal{H}^{(n)\uparrow}$.

Example 4.5.3 (2-SuperHyperGraph attention on a small hierarchical interaction model). We present a concrete example of $n$-SuperHyperGraph attention (Definition 4.5.2) with $n = 2$.

Let the base vertex set be

$$V_0 = \{x_1, x_2, x_3\}.$$

Define the following 2-supervertices (each is an element of $\mathcal{P}^2(V_0) = \mathcal{P}(\mathcal{P}(V_0))$):

$$u_1 := \big\{\{x_1, x_2\}\big\}, \qquad u_2 := \big\{\{x_2, x_3\}\big\}, \qquad u_3 := \big\{\{x_3\}\big\}.$$

Hence

$$V^{(2)} = \{u_1, u_2, u_3\}.$$

Let the 2-superedge family be

$$E^{(2)} = \{\varepsilon_1, \varepsilon_2\}, \qquad \varepsilon_1 = \{u_1, u_2\}, \quad \varepsilon_2 = \{u_2, u_3\}.$$

Step 1: Expanded hypergraph $\mathcal{H}^{(2)\uparrow}$. Using the recursive base-support map, we have

$$\mathrm{Supp}(u_1) = \{x_1, x_2\}, \qquad \mathrm{Supp}(u_2) = \{x_2, x_3\}, \qquad \mathrm{Supp}(u_3) = \{x_3\}.$$

Therefore, the expanded hyperedges are

$$e_1^{(2)\uparrow} = \mathrm{Supp}(u_1) \cup \mathrm{Supp}(u_2) = \{x_1, x_2, x_3\}, \qquad e_2^{(2)\uparrow} = \mathrm{Supp}(u_2) \cup \mathrm{Supp}(u_3) = \{x_2, x_3\}.$$

Thus

$$\mathcal{H}^{(2)\uparrow} = (V_0, E^{(2)\uparrow}), \qquad E^{(2)\uparrow} = \{e_1^{(2)\uparrow}, e_2^{(2)\uparrow}\}.$$

Ordering rows by $(x_1, x_2, x_3)$ and columns by $(e_1^{(2)\uparrow}, e_2^{(2)\uparrow})$, the expanded incidence matrix is

$$B^{(2)\uparrow} = \begin{bmatrix} 1 & 0 \\ 1 & 1 \\ 1 & 1 \end{bmatrix}.$$

Step 2: Attention coefficients and attention-weighted incidence. Assume an attention module (e.g., score + softmax) produces the following coefficients on incident pairs $(x_i, e_j^{(2)\uparrow})$:

$$\alpha_{11} = 0.2, \qquad \alpha_{21} = 0.5, \qquad \alpha_{31} = 0.3,$$

for $e_1^{(2)\uparrow}$, and

$$\alpha_{22} = 0.6, \qquad \alpha_{32} = 0.4,$$

for $e_2^{(2)\uparrow}$. (These are zero on non-incident pairs.) Hence the attention-weighted incidence matrix is

$$\widetilde{B}^{(2)} = \begin{bmatrix} 0.2 & 0 \\ 0.5 & 0.6 \\ 0.3 & 0.4 \end{bmatrix}.$$

Step 3: Degree matrices and weights. Choose expanded-hyperedge weights

$$w\left(e_1^{(2)\uparrow}\right) = 2, \qquad w\left(e_2^{(2)\uparrow}\right) = 1.$$

Then

$$W^{(2)} = \mathrm{diag}(2, 1).$$

Using the expanded hypergraph (as in Definition 4.1.6), we get

$$D_E^{(2)} = \mathrm{diag}\big(|e_1^{(2)\uparrow}|,\ |e_2^{(2)\uparrow}|\big) = \mathrm{diag}(3, 2),$$

and

$$D_V^{(2)} = \mathrm{diag}(2, 3, 3),$$

since

$$d(x_1) = 2, \qquad d(x_2) = 2 + 1 = 3, \qquad d(x_3) = 2 + 1 = 3.$$

Step 4: Attention-based propagation matrix. Define

$$P_{\mathrm{att}} := (D_V^{(2)})^{-1/2}\, \widetilde{B}^{(2)}\, W^{(2)}\, (D_E^{(2)})^{-1}\, \widetilde{B}^{(2)\top}\, (D_V^{(2)})^{-1/2}.$$

A direct computation yields

$$P_{\mathrm{att}} = \begin{bmatrix} \frac{1}{75} & \frac{\sqrt{6}}{90} & \frac{\sqrt{6}}{150} \\ \frac{\sqrt{6}}{90} & \frac{26}{225} & \frac{11}{150} \\ \frac{\sqrt{6}}{150} & \frac{11}{150} & \frac{7}{150} \end{bmatrix} \approx \begin{bmatrix} 0.0133 & 0.0272 & 0.0163 \\ 0.0272 & 0.1156 & 0.0733 \\ 0.0163 & 0.0733 & 0.0467 \end{bmatrix}.$$

Step 5: Input features and output. Let the base-vertex feature matrix be

$$X = \begin{bmatrix} 1 & 0 \\ 0 & 1 \\ 1 & 1 \end{bmatrix} \in \mathbb{R}^{3\times 2},$$

and choose

$$\Theta_0 = I_2, \qquad \sigma = \mathrm{ReLU}.$$

Then

$$X^{\mathrm{out}} = \sigma(P_{\mathrm{att}} X \Theta_0) = \sigma(P_{\mathrm{att}} X).$$

Since all entries are nonnegative in this example, ReLU acts as the identity, and

$$X^{\mathrm{out}} = \begin{bmatrix} \frac{1}{75} + \frac{\sqrt{6}}{150} & \frac{4\sqrt{6}}{225} \\ \frac{\sqrt{6}}{90} + \frac{11}{150} & \frac{17}{90} \\ \frac{\sqrt{6}}{150} + \frac{7}{150} & \frac{3}{25} \end{bmatrix} \approx \begin{bmatrix} 0.0297 & 0.0435 \\ 0.1005 & 0.1889 \\ 0.0630 & 0.1200 \end{bmatrix}.$$

This example shows how 2-SuperHyperGraph attention is implemented by:

1. expanding the 2-SuperHyperGraph to a base-level hypergraph,
2. replacing the binary incidence $B^{(2)\uparrow}$ with an attention-weighted incidence $\widetilde{B}^{(2)}$, and
3. applying normalized hypergraph-style propagation on the base features.

The attention coefficients allow the model to emphasize more influential base vertices within each expanded hyperedge.

Theorem 4.5.4 (Reduction to hypergraph attention). If $n = 1$ and the 1-SuperHyperGraph $H^{(1)}$ encodes an ordinary hypergraph on $V_0$ (via singleton supervertices, as in Theorem 4.1.8), then $n$-SuperHyperGraph attention (Definition 4.5.2) coincides with standard hypergraph attention.

Proof. Under the singleton encoding, the expanded hypergraph $\mathcal{H}^{(1)\uparrow}$ is exactly the original hypergraph on $V_0$, and $B^{(1)\uparrow}$ equals its incidence matrix. Since Definition 4.5.2 applies the hypergraph-attention construction to the expanded hypergraph, the resulting coefficients, attention-weighted incidence matrix, and propagation operator match those of hypergraph attention. □

# 5 Review and Results: Uncertain Graph Neural Networks

In this section, we explore uncertain graph networks, including Fuzzy Graph Neural Networks, Neutrosophic Graph Neural Networks, and Plithogenic Graph Neural Networks.

## 5.1 Fuzzy Graph Neural Network (F-GNN)

In this section, we introduce the concept of the Fuzzy Graph Neural Network (F-GNN). A Fuzzy Graph Neural Network (F-GNN) is a graph-based inference model that combines the principles of fuzzy logic with graph neural networks (GNNs). It is specifically designed to handle fuzzy and uncertain data in graph-structured information (cf. [238, 239, 240]). Related concepts, such as Fuzzy Neural Networks, are also well known [241, 242]. Below, we present a formal definition of F-GNN.

Definition 5.1.1 (Fuzzy Graph Neural Network (F-GNN)). [243] An F-GNN is defined as a quintuple:

$$\text{F-GNN} = (G, \mathcal{F}_V, \mathcal{F}_E, \mathcal{R}, \mathcal{D}),$$

where:

- $G = (V, E)$ is a graph where $V$ represents the set of vertices and $E$ represents the set of edges.
- $\mathcal{F}_V$ and $\mathcal{F}_E$ are the fuzzification functions for vertices and edges, respectively. These functions map vertex and edge attributes to fuzzy membership values:

  $$\mathcal{F}_V : \mathcal{X}_V \to [0,1]^M, \quad \mathcal{F}_E : \mathcal{X}_E \to [0,1]^M,$$

  where $M$ is the number of fuzzy subsets, and $\mathcal{X}_V$ and $\mathcal{X}_E$ denote the attribute spaces for vertices and edges.
- $\mathcal{R}$ represents the rule layer, which encodes fuzzy rules of the form:

  $$\text{IF } \bigwedge_{i=1}^{N} \text{vertex } v_i \text{ satisfies } \mathcal{F}_V(v_i) \text{ THEN } \mathcal{D}(v_i) \text{ outputs the prediction,}$$

  where $\mathcal{D}$ is the defuzzification layer.
- $\mathcal{D}$ is the defuzzification function, which aggregates the outputs of the rule layer to produce a crisp output for each vertex or edge.

Definition 5.1.2. [243] Given an input graph $G = (V, E)$ with vertex features $X_V$ and edge features $X_E$, F-GNN operates as follows:

1. Fuzzification Layer: Each vertex $v \in V$ and edge $e \in E$ is fuzzified using membership functions:

   $$\mathcal{F}_V(v) = [\mu_1(v), \mu_2(v), \dots, \mu_M(v)], \quad \mathcal{F}_E(e) = [\mu_1(e), \mu_2(e), \dots, \mu_M(e)].$$

2. Rule Layer: A set of fuzzy rules is defined to aggregate neighborhood information. For example:

   $$\text{IF } v \in A_m \text{ AND } u \in A_n \text{ THEN } y_k = f_k(x_v, x_u),$$

   where $A_m, A_n$ are fuzzy subsets, $x_v, x_u$ are vertex features, and $f_k$ is a trainable function.

3. Normalization Layer: The firing strength of each rule is normalized:

$$\hat{r}_k = \frac{r_k}{\sum_{j=1}^{K} r_j},$$

where $r_k$ is the firing strength of the $k$-th rule.

4. Defuzzification Layer: The normalized rule outputs are aggregated to produce crisp predictions:

$$y = \sum_{k=1}^{K} \hat{r}_k \cdot f_k(x).$$

Definition 5.1.3. [243] For a multi-layer F-GNN, the $l$-th layer is defined as:

$$H^{(l)} = \sigma\left(f_\theta\left(H^{(l-1)}, A\right) + H^{(l-1)}\right),$$

where:

- $H^{(l)}$ is the output of the $l$-th layer.
- $\sigma$ is a non-linear activation function (e.g., ReLU).
- $A$ is the adjacency matrix of the graph.
- $f_\theta$ is a trainable function.

The final output of the F-GNN is:

$$Y = \text{Softmax}\left(H^{(L)}\right),$$

where $L$ is the number of layers in the F-GNN.

Theorem 5.1.4. A Fuzzy Graph Neural Network (F-GNN) generalizes a Graph Neural Network (GNN).

Proof. To prove this, we show that the definition of an F-GNN encompasses the definition of a GNN as a special case.

1. Graph Structure: Both GNNs and F-GNNs operate on a graph $G = (V, E)$, where $V$ is the set of vertices, and $E \subseteq V \times V$ is the set of edges. While GNNs use crisp edge connections, F-GNNs extend this by assigning fuzzy membership values to vertices and edges through the fuzzification functions $\mathcal{F}_V$ and $\mathcal{F}_E$:

$$\mathcal{F}_V : \mathcal{X}_V \to [0,1]^M, \quad \mathcal{F}_E : \mathcal{X}_E \to [0,1]^M.$$

When $M = 1$ and membership values are restricted to binary $\{0, 1\}$, the F-GNN reduces to a standard GNN, where $\mathcal{F}_V$ and $\mathcal{F}_E$ represent crisp vertices and edges.

2. Message Passing: In a GNN, messages between nodes are exchanged using functions $\phi_m$ and aggregated at each node $v_i$ as:

$$\mathbf{m}_i^{(t+1)} = \sum_{v_j \in \mathcal{N}(i)} \phi_m(\mathbf{h}_i^{(t)}, \mathbf{h}_j^{(t)}, \mathbf{e}_{ij}),$$

where $\mathcal{N}(i)$ is the set of neighbors of $v_i$.

In an F-GNN, the message passing incorporates fuzzy membership values through the rule layer $\mathcal{R}$, which defines fuzzy rules such as:

$$\text{IF } v_i \in A_m \text{ AND } v_j \in A_n \text{ THEN } f_k(\mathbf{h}_i, \mathbf{h}_j, \mathbf{e}_{ij}),$$

where $A_m$ and $A_n$ are fuzzy subsets, and $f_k$ is a trainable function. If fuzzy subsets $A_m$ and $A_n$ are crisp (e.g., $A_m = A_n = \{1\}$), the F-GNN reduces to the standard message passing mechanism of a GNN.

3. Node Updates: In a GNN, node updates are defined as:

$$\mathbf{h}_i^{(t+1)} = \phi_u(\mathbf{h}_i^{(t)}, \mathbf{m}_i^{(t+1)}),$$

where $\phi_u$ is a node update function.

In an F-GNN, node updates are governed by fuzzy rules and defuzzification, aggregating over normalized firing strengths:

$$y = \sum_{k=1}^{K} \hat{r}_k \cdot f_k(\mathbf{h}_i),$$

where $\hat{r}_k$ is the normalized firing strength of the $k$-th fuzzy rule. If there is only one rule ($K = 1$) and no fuzzification is applied, the F-GNN node update simplifies to the standard GNN node update.

4. Generalization: The fuzzification and defuzzification layers in an F-GNN extend the crisp operations of a GNN by introducing degrees of membership, enabling the model to handle uncertainty and imprecision. When these additional features are disabled (e.g., by setting $M = 1$ and $K = 1$), the F-GNN reduces exactly to a GNN.

Since every operation in a GNN is a special case of the corresponding operation in an F-GNN, we conclude that the F-GNN generalizes the GNN. □

## 5.2 Neutrosophic Graph Neural Network (N-GNN)

In this section we formalize a Neutrosophic Graph Neural Network (N-GNN) as a message-passing model whose propagation is modulated by neutrosophic (truth–indeterminacy–falsity) assessments on vertices and/or edges [244, 245]. Related concepts include Neutrosophic Neural Networks [246, 247] and Neutrosophic Machine Learning [248, 249, 250]. We also show that a fuzzy GNN is obtained as a natural specialization.

Let $\mathbb{N}_{[0,1]} := [0,1]^3$. Elements $(t, i, f) \in \mathbb{N}_{[0,1]}$ are interpreted as degrees of truth, indeterminacy, and falsity, respectively.

Definition 5.2.1 ((Recall) Single-valued neutrosophic graph). [168] A single-valued neutrosophic graph is a tuple

$$G_N = (V, E, \nu_V, \nu_E),$$

where $G = (V, E)$ is a (simple) graph and

$$\nu_V : V \to \mathbb{N}_{[0,1]}, \qquad \nu_E : E \to \mathbb{N}_{[0,1]}$$

assign neutrosophic triplets to vertices and edges. Writing $\nu_V(v) = (T_V(v), I_V(v), F_V(v))$ and $\nu_E(e) = (T_E(e), I_E(e), F_E(e))$, we only require $T_*, I_*, F_* \in [0,1]$ pointwise.

(Optional consistency conditions.) Some applications additionally impose constraints relating edge and endpoint degrees, e.g., for $e = \{u, v\} \in E$,

$$T_E(e) \le \min\{T_V(u), T_V(v)\}, \quad I_E(e) \le \max\{I_V(u), I_V(v)\}, \quad F_E(e) \le \max\{F_V(u), F_V(v)\}.$$

When such constraints are assumed, we call $G_N$ consistent.

We now define an N-GNN in a way that is compatible with standard GNN notation while keeping the neutrosophic components explicit.

Definition 5.2.2 (Neutrosophic Graph Neural Network (N-GNN)). Fix a graph $G = (V, E)$ with $|V| = n$. Let $X \in \mathbb{R}^{n \times d}$ be the input node-feature matrix, where the $v$-th row $x_v^\top$ is the feature of $v \in V$.

An N-GNN of depth $L$ is specified by the data

$$\text{N-GNN} = \Big(G,\ \nu_V, \nu_E,\ \{\Phi^{(\ell)}\}_{\ell=0}^{L-1},\ \sigma,\ \text{Readout}\Big),$$

where:

- $\nu_V : V \to [0,1]^3$ and $\nu_E : E \to [0,1]^3$ define a single-valued neutrosophic graph $G_N = (V, E, \nu_V, \nu_E)$ (Definition 5.2.1). These can be given or learned from attributes via fuzzification maps $\mathcal{N}_V, \mathcal{N}_E$ (see below).
- Each layer operator $\Phi^{(\ell)}$ maps node embeddings to node embeddings, $\Phi^{(\ell)} : \mathbb{R}^{n\times d_\ell} \to \mathbb{R}^{n\times d_{\ell+1}}$, and is allowed to depend on $\nu_V, \nu_E$.
- $\sigma$ is an activation function applied entrywise.
- Readout is a task-specific map (e.g., linear classifier + softmax for node classification, or a pooling operator for graph classification). In particular, Readout may include a defuzzification step that maps neutrosophic quantities to crisp scores.

The forward propagation is $H^{(0)} := X$ and

$$H^{(\ell+1)} \;:=\; \sigma\big(\Phi^{(\ell)}(H^{(\ell)})\big), \qquad \ell = 0, 1, \ldots, L-1,$$

followed by $Y := \mathrm{Readout}(H^{(L)})$.

Remark 5.2.3 (Learning neutrosophic degrees). A common choice is to obtain $\nu_V, \nu_E$ from attributes by neutrosophic fuzzification:

$$\mathcal{N}_V : \mathbb{R}^d \to [0,1]^3, \qquad \mathcal{N}_E : \mathbb{R}^{d_e} \to [0,1]^3,$$

and then set $\nu_V(v) = \mathcal{N}_V(x_v)$ and $\nu_E(e) = \mathcal{N}_E(x_e)$ (if edge features $x_e$ exist), possibly with a projection $\Pi : [0,1]^3 \to [0,1]^3$ to enforce optional consistency constraints.

To make the definition operational, we present a standard matrix-form layer whose adjacency weights are derived from neutrosophic edge degrees.

Definition 5.2.4 (Neutrosophic-weighted propagation layer). Let $G_N = (V, E, \nu_V, \nu_E)$ be a single-valued neutrosophic graph with $|V| = n$. Fix any edge weighting functional

$$g : [0,1]^3 \to \mathbb{R}_{\ge 0}.$$

Define the neutrosophic weighted adjacency $\widetilde{A} \in \mathbb{R}^{n\times n}_{\ge 0}$ by

$$\widetilde{A}_{uv} = \begin{cases} g\big(\nu_E(\{u,v\})\big), & \{u,v\} \in E, \\ 0, & \text{otherwise}, \end{cases}$$

and the corresponding degree matrix $\widetilde{D}$ by $\widetilde{D}_{uu} := \sum_{v=1}^{n} \widetilde{A}_{uv}$. (Optionally add self-loops by replacing $\widetilde{A} \leftarrow \widetilde{A} + I_n$.)

Given embeddings $H \in \mathbb{R}^{n\times d_\ell}$ and parameters $\Theta^{(\ell)} \in \mathbb{R}^{d_\ell \times d_{\ell+1}}$, define

$$\Phi^{(\ell)}(H) \;:=\; \widetilde{D}^{-1/2}\, \widetilde{A}\, \widetilde{D}^{-1/2}\, H\, \Theta^{(\ell)}.$$

Then $H^{(\ell+1)} = \sigma(\Phi^{(\ell)}(H^{(\ell)}))$ is a valid N-GNN layer in the sense of Definition 5.2.2.

We formalize a minimal fuzzy-GNN model as a specialization where each edge carries a single membership weight.

Definition 5.2.5 (Fuzzy-weighted GNN (F-GNN), minimal form). Let $G = (V, E)$ be a graph and let $\mu_E : E \to [0,1]$ be a fuzzy edge-membership function. Define a weighted adjacency $A^\mu$ by $A^\mu_{uv} = \mu_E(\{u,v\})$ if $\{u,v\} \in E$ and 0 otherwise, and let $D^\mu$ be its degree matrix. A (normalized) fuzzy propagation layer is

$$\Phi_{\mathrm{F}}(H) \;:=\; (D^\mu)^{-1/2}\, A^\mu\, (D^\mu)^{-1/2}\, H\, \Theta, \qquad H^+ = \sigma(\Phi_{\mathrm{F}}(H)).$$

A network obtained by stacking such layers is called an F-GNN.

Theorem 5.2.6 (N-GNN generalizes F-GNN). Every F-GNN (Definition 5.2.5) is a special case of an N-GNN (Definition 5.2.2).

Proof. Let an F-GNN be given with fuzzy edge-membership $\mu_E : E \to [0,1]$. Define a neutrosophic edge map $\nu_E : E \to [0,1]^3$ by the standard embedding of a fuzzy degree into a single-valued neutrosophic triple, for example

$$\nu_E(e) \;:=\; \big(\mu_E(e),\, 0,\, 1-\mu_E(e)\big) \qquad (e \in E),$$

and choose any $\nu_V$ (it is irrelevant for the layer in Definition 5.2.4). Now define $g : [0,1]^3 \to \mathbb{R}_{\geq 0}$ by $g(t,i,f) := t$. Then the neutrosophic weighted adjacency $\widetilde{A}$ from Definition 5.2.4 satisfies $\widetilde{A}_{uv} = \mu_E(\{u,v\})$, hence $\widetilde{A} = A^\mu$ and $\widetilde{D} = D^\mu$. Therefore the N-GNN layer

$$\widetilde{D}^{-1/2}\,\widetilde{A}\,\widetilde{D}^{-1/2}\,H\,\Theta$$

coincides with the fuzzy layer $(D^\mu)^{-1/2}A^\mu(D^\mu)^{-1/2}H\Theta$. Stacking layers preserves equality, so the entire F-GNN is realized as an N-GNN. □

The next statement clarifies in what precise sense an N-GNN "has the structure" of a neutrosophic graph.

Theorem 5.2.7 (An N-GNN operates on a neutrosophic graph). Let N-GNN $= (G, \nu_V, \nu_E, \ldots)$ be as in Definition 5.2.2. Then $(V, E, \nu_V, \nu_E)$ is a single-valued neutrosophic graph (Definition 5.2.1), and each layer operator $\Phi^{(\ell)}$ may be viewed as a propagation rule on this neutrosophic graph. If, in addition, $(V, E, \nu_V, \nu_E)$ is assumed consistent (Definition 5.2.1) and the learning procedure projects memberships back to the consistent set (e.g., by clipping/projection $\Pi$), then the model preserves consistency throughout training and inference.

Proof. By Definition 5.2.2, the maps $\nu_V : V \to [0,1]^3$ and $\nu_E : E \to [0,1]^3$ are part of the model specification. Thus $(V, E, \nu_V, \nu_E)$ is, by definition, a single-valued neutrosophic graph.

Each layer $\Phi^{(\ell)}$ is allowed to depend on $\nu_V, \nu_E$ and therefore defines a message-passing/propagation rule that is parameterized by the neutrosophic degrees on the underlying graph. If consistency constraints are imposed, and a projection $\Pi$ onto the consistent set is applied after each update of $\nu_V, \nu_E$ (or after fuzzification), then the constrained property is preserved by construction. □

## 5.3 Plithogenic Graph Neural Network (P-GNN)

We next formalize a Plithogenic Graph Neural Network (P-GNN) as a message-passing model whose propagation is governed by (i) degrees of appurtenance of vertices/edges to attribute values and (ii) degrees of contradiction between attribute values. We also show that this framework subsumes both neutrosophic- and fuzzy-based GNNs as special cases.

A standard way to incorporate contradiction into aggregation is to blend a t-norm and t-conorm. In what follows we use the classical pair min, max, but any fixed t-norm/t-conorm pair can be used.

Definition 5.3.1 (Plithogenic binary blending). For $c \in [0,1]$ and $x, y \in [0,1]$, define

$$x \wedge_c y \;:=\; (1-c)\min\{x,y\} + c\max\{x,y\}, \qquad x \vee_c y \;:=\; (1-c)\max\{x,y\} + c\min\{x,y\}.$$

Then $\wedge_c, \vee_c : [0,1]^2 \to [0,1]$ are well-defined for every $c \in [0,1]$.

Let $X \in \mathbb{R}^{n\times d}$ be the node-feature matrix, with $x_v \in \mathbb{R}^d$ the row corresponding to $v \in V$. A P-GNN may treat $\alpha_V, \alpha_E$ as given or learned from features via fuzzification maps.

Definition 5.3.2 (Plithogenic fuzzification and defuzzification). A plithogenic fuzzification consists of maps

$$\mathcal{P}_V : \mathbb{R}^d \to [0,1]^{A_V}, \qquad \mathcal{P}_E : \mathbb{R}^{d_e} \to [0,1]^{A_E}$$

(where $[0,1]^A$ denotes the set of functions $A \to [0,1]$). Given $\mathcal{P}_V$, we define $\alpha_V$ from $X$ by $\alpha_V(v,a) := \big(\mathcal{P}_V(x_v)\big)(a)$. Similarly, if edge features exist, $\alpha_E(e,b) := \big(\mathcal{P}_E(x_e)\big)(b)$.

A plithogenic defuzzification is any map $\mathcal{D}_P$ that converts the final embeddings to task outputs, e.g. $\mathcal{D}_P : \mathbb{R}^{n\times d_L} \to \mathbb{R}^{n\times C}$ for $C$-class node classification.

We now give an explicit propagation rule (a concrete layer) whose edge weights are computed from degrees of appurtenance and contradiction.

Definition 5.3.3 (Plithogenic-weighted propagation layer). Fix a plithogenic graph

$$G_P = (V, E, \alpha_V, \alpha_E, \kappa_V, \kappa_E)$$

. For each vertex $v \in V$, choose a (measurable) selector

$$a^\star(v) \in \arg\max_{a\in A_V} \alpha_V(v,a),$$

$$s(v) := \max_{a\in A_V} \alpha_V(v,a) \in [0,1].$$

For each edge $\{u,v\} \in E$, define a contradiction coefficient

$$c_{uv} := \kappa_V\big(a^\star(u), a^\star(v)\big) \in [0,1],$$

and the induced (scalar) plithogenic edge weight

$$w_{uv} := s(u) \wedge_{c_{uv}} s(v) \in [0,1] \quad \text{(Definition 5.3.1)}.$$

Define the weighted adjacency $\widetilde{A} \in \mathbb{R}^{n\times n}_{\geq 0}$ by

$$\widetilde{A}_{uv} = \begin{cases} w_{uv}, & \{u,v\} \in E, \\ 0, & \text{otherwise,} \end{cases}$$

and the degree matrix $\widetilde{D}$ by $\widetilde{D}_{uu} := \sum_{v=1}^{n} \widetilde{A}_{uv}$. (Optionally add self-loops by replacing $\widetilde{A} \leftarrow \widetilde{A} + I_n$.)

Given embeddings $H \in \mathbb{R}^{n\times d_\ell}$ and parameters $\Theta^{(\ell)} \in \mathbb{R}^{d_\ell \times d_{\ell+1}}$, define the layer operator

$$\Phi_P^{(\ell)}(H) \;:=\; \widetilde{D}^{-1/2}\, \widetilde{A}\, \widetilde{D}^{-1/2}\, H\, \Theta^{(\ell)}.$$

Definition 5.3.4 (Plithogenic Graph Neural Network (P-GNN)). A P-GNN of depth $L$ is specified by

$$\text{P-GNN} = \Big(G,\ A_V, A_E,\ \kappa_V, \kappa_E,\ \mathcal{P}_V, \mathcal{P}_E,\ \{\Theta^{(\ell)}\}_{\ell=0}^{L-1},\ \sigma,\ \mathcal{D}_P\Big),$$

where:

- $G = (V, E)$ is a graph and $A_V, A_E$ are finite attribute-value sets;
- $\kappa_V, \kappa_E$ are contradiction functions as in Definition 2.5.12;
- $\mathcal{P}_V, \mathcal{P}_E$ are fuzzification maps (Definition 5.3.2) that produce $\alpha_V, \alpha_E$;
- $\Phi_P^{(\ell)}$ is a plithogenic propagation operator (e.g. Definition 5.3.3);
- $\sigma$ is an activation function (entrywise);
- $\mathcal{D}_P$ is a task-specific defuzzification/readout map.

The forward pass is $H^{(0)} := X$ and

$$H^{(\ell+1)} := \sigma\big(\Phi_P^{(\ell)}(H^{(\ell)})\big), \qquad \ell = 0, 1, \ldots, L-1,$$

followed by the output $Y := \mathcal{D}_P(H^{(L)})$.

We state this precisely as "containment by specialization."

Theorem 5.3.5 (P-GNN generalizes N-GNN and F-GNN). Every N-GNN and every F-GNN can be realized as a special case of a P-GNN.

Proof. We give two specializations.

(1) F-GNN as a special case. Fix $A_V = \{\star\}$ and define $\kappa_V(\star, \star) = 0$. Let $\alpha_V(v, \star) \in [0, 1]$ be any fuzzy membership/importance score for $v$. Then $a^\star(v) = \star$, $c_{uv} = 0$, and

$$w_{uv} = s(u) \wedge_0 s(v) = \min\{s(u), s(v)\}.$$

If one further chooses $s(v) \equiv 1$ (or absorbs $s(\cdot)$ into an external edge-membership $\mu_E$), then $\widetilde{A}$ reduces to the usual weighted adjacency used in fuzzy-weighted GNN layers. Hence the P-GNN layer in Definition 5.3.3 reduces to the corresponding fuzzy/weighted GNN layer.

(2) N-GNN as a special case. Let $A_V = \{T, I, F\}$ and define

$$\alpha_V(v, T) = T_V(v), \quad \alpha_V(v, I) = I_V(v),$$

$$\alpha_V(v, F) = F_V(v),$$

where $(T_V(v), I_V(v), F_V(v)) \in [0, 1]^3$ is the neutrosophic triplet assigned to $v$. Choose any fixed contradiction function $\kappa_V$ (it may be identically 0, or reflect prior knowledge). If we set $\kappa_V \equiv 0$, then $w_{uv}$ depends only on the selected appurtenance maxima and reduces to a standard neutrosophic-weighted adjacency construction in which contradiction does not intervene. Consequently, a neutrosophic-weighted propagation (and hence an N-GNN layer) is recovered as a special case of Definition 5.3.3.

Therefore, both F-GNN and N-GNN are obtained by specialization of the P-GNN data, so P-GNN generalizes both. □

The following statement is mathematically correct under the standard assumption that hesitant fuzzy memberships are finite sets.

Corollary 5.3.6 (Finite hesitant fuzzy GNNs embed into P-GNNs). Assume a hesitant fuzzy model assigns to each vertex $v \in V$ a finite set $H(v) \subset [0, 1]$ of possible membership degrees. Then there exists a P-GNN whose plithogenic representation can encode the hesitant information, and whose propagation can emulate aggregation rules based on $\max H(v)$, $\min H(v)$, or any fixed symmetric statistic of $H(v)$.

Proof. Let $A_V := \bigcup_{v \in V} H(v)$, which is finite by assumption, and define

$$\alpha_V(v, a) := \begin{cases} 1, & a \in H(v), \\ 0, & a \notin H(v). \end{cases} \qquad \kappa_V \equiv 0.$$

Then any fixed statistic $\mathrm{Stat}(H(v)) \in [0, 1]$ that depends only on the set $H(v)$ (e.g. $\max H(v)$, $\min H(v)$, or the mean of $H(v)$) can be computed from $\alpha_V(v, \cdot)$ by a deterministic map. Plugging $\mathrm{Stat}(H(\cdot))$ into the scalar score $s(v)$ in Definition 5.3.3 yields an adjacency weighting consistent with hesitant-fuzzy-style propagation rules. □

We formalize what can be guaranteed without overclaiming.

Theorem 5.3.7 (Closure of appurtenance and weights). In Definition 5.3.3, for all $u, v \in V$ we have $w_{uv} \in [0, 1]$. Consequently, the induced adjacency $\widetilde{A}$ has nonnegative entries and $\widetilde{D}$ is well-defined.

Proof. By construction $s(u), s(v) \in [0, 1]$ and $c_{uv} \in [0, 1]$. Definition 5.3.1 shows that $x \wedge_c y$ is a convex combination of $\min\{x, y\}$ and $\max\{x, y\}$, hence lies in $[0, 1]$. Therefore $w_{uv} \in [0, 1]$ and $\widetilde{A} \geq 0$. □

Theorem 5.3.8 (P-GNN is at least as expressive as N-GNN and F-GNN). The hypothesis class representable by P-GNNs contains the hypothesis classes representable by N-GNNs and by F-GNNs, under the specializations in Theorem 5.3.5.

Proof. Immediate from Theorem 5.3.5, since each N-GNN/F-GNN is realized by a P-GNN with restricted parameters. □

Since convergence is not automatic for general neural architectures, we state a standard sufficient condition.

Theorem 5.3.9 (Convergence under a contraction condition). Fix a graph $G$ and fix the plithogenic-weighted normalized propagation matrix

$$P \ := \ \widetilde{D}^{-1/2}\widetilde{A}\widetilde{D}^{-1/2}.$$

Assume $\sigma : \mathbb{R} \to \mathbb{R}$ is 1-Lipschitz (e.g. ReLU, LeakyReLU, tanh). Fix a parameter matrix $\Theta$ and consider the iteration on embeddings

$$H^{(t+1)} \ = \ \sigma\big(PH^{(t)}\Theta\big), \qquad t = 0, 1, 2, \dots$$

on the complete metric space $(\mathbb{R}^{n\times d}, \|\cdot\|_F)$. If

$$\|P\|_2 \, \|\Theta\|_2 \ < \ 1,$$

then the map $F(H) := \sigma(PH\Theta)$ is a contraction and the iterates $H^{(t)}$ converge (for any start $H^{(0)}$) to the unique fixed point $H^\star$ satisfying $H^\star = \sigma(PH^\star\Theta)$.

Proof. For any $H_1, H_2$,

$$\|F(H_1) - F(H_2)\|_F \leq \|P(H_1 - H_2)\Theta\|_F \leq \|P\|_2 \, \|\Theta\|_2 \, \|H_1 - H_2\|_F,$$

where we used that $\sigma$ is 1-Lipschitz entrywise and submultiplicativity of operator norms. If $\|P\|_2\|\Theta\|_2 < 1$, then $F$ is a contraction. Banach's fixed-point theorem yields existence and uniqueness of the fixed point and convergence of the iterates. □

We present a concrete forward-pass algorithm for a P-GNN in which plithogenic information (degrees of appurtenance and contradiction) is embedded into a nonnegative edge-weighting and then used in a normalized message-passing layer.

Remark 5.3.10 (What is (and is not) computed). Algorithm 5 is mathematically well-posed once the fuzzifiers $\mathcal{P}_V, \mathcal{P}_E$ and contradiction functions $\kappa_V, \kappa_E$ are fixed. Note that $\kappa_V$ and $\kappa_E$ are functions on attribute values (not on vertices/edges directly); the algorithm evaluates contradiction along an edge via the selected dominant values $a^\star(u), a^\star(v)$ (and analogously one may use $\kappa_E$ if edge-type contradictions are needed). Other choices are possible (e.g. attention over all values), but the above is a concrete and correct instantiation.

Theorem 5.3.11 (Algorithmic correctness). Let the inputs of Algorithm 5 be fixed. Define the plithogenic weighted adjacency $\widetilde{A}$ and degree matrix $\widetilde{D}$ as in Steps (3)–(4), and let $P = \widetilde{D}^{-1/2}\widetilde{A}\widetilde{D}^{-1/2}$. Then the algorithm outputs

$$H^{(L)} \ = \ \Big(\sigma \circ T_{L-1} \circ \cdots \circ \sigma \circ T_0\Big)(X), \qquad T_\ell(H) := P\, H\, \Theta^{(\ell)},$$

and $Y = \rho(H^{(L)})$.

Proof. Steps (1)–(3) define scalars $s_V(v) \in [0,1]$, selected labels $a^\star(v) \in A_V$, and weights $w_{uv} \in [0,1]$ for each edge $\{u,v\} \in E$, hence a symmetric nonnegative matrix $\widetilde{A}$. Step (4) forms $\widetilde{D}$ and the normalized propagation matrix $P$. Step (5) then applies, for each $\ell$, the update $H^{(\ell+1)} = \sigma(PH^{(\ell)}\Theta^{(\ell)})$ with $H^{(0)} = X$, which is exactly the stated composition. Finally Step (6) applies the readout $\rho$. □

Theorem 5.3.12 (Time complexity). Let $n = |V|$, $m = |E|$, and let $q_V := |A_V|$, $q_E := |A_E|$. Assume $\widetilde{A}$ is stored sparsely with $O(m)$ nonzeros and that multiplying a sparse $n \times n$ matrix with $O(m)$ nonzeros by an $n \times d_\ell$ dense matrix costs $O(m\, d_\ell)$. Let $\mathrm{cost}(\mathcal{P}_V)$ be the cost to evaluate $\mathcal{P}_V$ on one node feature, and similarly $\mathrm{cost}(\mathcal{P}_E)$ for one edge feature. Then the running time of Algorithm 5 is

$$O\Big(n\,\mathrm{cost}(\mathcal{P}_V) \ + \ nq_V \ + \ m\,\mathrm{cost}(\mathcal{P}_E) \ + \ mq_E \ + \ m \ + \ \sum_{\ell=0}^{L-1}\big(m\,d_\ell + n\,d_\ell d_{\ell+1}\big)\Big).$$

In the common fixed-width case $d_\ell = d$ for all $\ell$, this simplifies to

$$O\Big(n\,\mathrm{cost}(\mathcal{P}_V) + m\,\mathrm{cost}(\mathcal{P}_E) + \; nq_V + mq_E + \; L\,(md + nd^2)\Big).$$

Proof. Steps (1) and (2) evaluate the fuzzifiers at all vertices/edges and compute maxima/argmax over $A_V, A_E$, costing $O(n\,\mathrm{cost}(\mathcal{P}_V) + nq_V)$ and $O(m\,\mathrm{cost}(\mathcal{P}_E) + mq_E)$, respectively. Step (3) processes each edge once, computing $c_{uv}$ and $w_{uv}$ and inserting into $\widetilde{A}$, costing $O(m)$. Step (4) forms $\widetilde{D}$ in $O(m)$ and normalizes (diagonal operations) in $O(n)$. For each layer $\ell$, computing $PH^{(\ell)}$ costs $O(md_\ell)$, and multiplying by $\Theta^{(\ell)}$ costs $O(nd_\ell d_{\ell+1})$; applying $\sigma$ costs $O(nd_{\ell+1})$ and is dominated by the matrix products. Summing over $\ell = 0, \dots, L-1$ yields the stated bound. □

Theorem 5.3.13 (Space complexity). Assume sparse storage of $\widetilde{A}$ with $O(m)$ nonzeros and in-place layer updates (i.e., we do not store all $\{H^{(\ell)}\}_{\ell=0}^{L}$ simultaneously). Then the working memory of Algorithm 5 is

$$O\Big(m \;+\; nd_{\max} \;+\; nq_V \;+\; mq_E \;+\; \sum_{\ell=0}^{L-1} d_\ell d_{\ell+1}\Big),$$

where $d_{\max} := \max_\ell d_\ell$. If one streams appurtenance values (keeping only $s_V(\cdot), a^\star(\cdot)$ and $s_E(\cdot)$), then the $nq_V + mq_E$ term can be reduced to $O(n + m)$.

Proof. Sparse $\widetilde{A}$ and (optionally) $P$ require $O(m)$ storage, and the degree vector/diagonal needs $O(n)$. The current embedding $H^{(\ell)}$ requires $O(nd_\ell)$; with in-place updates the dominant term is $O(nd_{\max})$. Storing all appurtenance vectors $\alpha_V(v,\cdot)$ and $\alpha_E(e,\cdot)$ costs $O(nq_V + mq_E)$, while storing only maxima/argmax and edge scores costs $O(n+m)$. Model parameters $\Theta^{(\ell)}$ contribute $\sum_\ell O(d_\ell d_{\ell+1})$. Adding these terms yields the claim. □

## 5.4 Uncertain Graph Neural Network

An Uncertain Graph Neural Network models noisy, incomplete, or probabilistic graph data by learning robust node and edge representations, propagating uncertainty-aware messages, and improving prediction reliability under ambiguity in practice. Throughout, fix an uncertain model $M$ with degree-domain $\mathrm{Dom}(M) \subseteq [0,1]^k$ ($k \geq 1$), and let $\mathcal{G}_M = (V, E, \mu_M)$ be an uncertain graph of type $M$ in the sense of Definition 2.5.16, i.e., $\mu_M : V \cup E \to \mathrm{Dom}(M)$. We write $N(v) := \{u \in V : \{u,v\} \in E\}$ for the (open) neighborhood of $v$.

Definition 5.4.1 (Uncertain graph neural network (U-GNN) of type $M$). Let $\mathcal{G}_M = (V, E, \mu_M)$ be a finite uncertain graph of type $M$. A U-GNN of type $M$ on $\mathcal{G}_M$ is a tuple

$$\text{U-GNN}_M = \big(\mathcal{G}_M,\ d_0, \dots, d_L,\ h^{(0)},\ (\mathrm{Msg}^{(l)}, \mathrm{Agg}^{(l)}, \mathrm{Upd}^{(l)})_{l=0}^{L-1},\ \mathrm{Readout}\big),$$

where:

- $L \in \mathbb{N}$ is the number of layers and $d_l \in \mathbb{N}$ are feature dimensions.
- $h^{(0)} : V \to \mathbb{R}^{d_0}$ is the initial vertex-feature map.
- For each layer $l = 0, \dots, L-1$:
  - $\mathrm{Msg}^{(l)} : \mathbb{R}^{d_l} \times \mathbb{R}^{d_l} \times \mathrm{Dom}(M)^3 \to \mathbb{R}^{d_l}$ is a message function;
  - $\mathrm{Agg}^{(l)}$ is a permutation-invariant aggregation operator mapping each finite multiset of $\mathbb{R}^{d_l}$ to a vector in $\mathbb{R}^{d_l}$ (e.g., sum, mean, max);
  - $\mathrm{Upd}^{(l)} : \mathbb{R}^{d_l} \times \mathbb{R}^{d_l} \to \mathbb{R}^{d_{l+1}}$ is an update function.
- $\mathrm{Readout} : \mathbb{R}^{d_L} \to \mathcal{Y}$ is a task-dependent output map (e.g., $\mathcal{Y} = \mathbb{R}^C$ for $C$-class logits).

Given these components, the forward propagation is defined recursively by

$$m_v^{(l)} := \mathrm{Agg}^{(l)}\Big(\{\mathrm{Msg}^{(l)}(h_v^{(l)}, h_u^{(l)}, \mu_M(v), \mu_M(u), \mu_M(\{u,v\})) : u \in N(v)\}\Big),$$

$$h_v^{(l+1)} := \mathrm{Upd}^{(l)}\big(h_v^{(l)}, m_v^{(l)}\big), \qquad (v \in V,\ l = 0, \dots, L-1),$$

and the node-level output is $y_v := \mathrm{Readout}(h_v^{(L)})$ for each $v \in V$.

Remark 5.4.2 (Optional degree-propagation layer). Definition 5.4.1 keeps the uncertainty-degree assignment $\mu_M$ fixed (as graph structure). If one wishes to update degrees across layers, it suffices to additionally specify maps

$$\mathrm{Deg}_V^{(l)} : \mathrm{Dom}(M) \times \mathrm{Dom}(M)^{t_l} \to \mathrm{Dom}(M), \qquad \mathrm{Deg}_E^{(l)} : \mathrm{Dom}(M) \times \mathrm{Dom}(M)^{s_l} \to \mathrm{Dom}(M),$$

and then define layerwise degree functions $\mu_M^{(l)} : V \cup E \to \mathrm{Dom}(M)$ by

$$\mu_M^{(0)} := \mu_M, \qquad \mu_M^{(l+1)}(v) := \mathrm{Deg}_V^{(l)}\big(\mu_M^{(l)}(v), (\mu_M^{(l)}(u))_{u \in N(v)}\big),$$

$$\mu_M^{(l+1)}(e) := \mathrm{Deg}_E^{(l)}\big(\mu_M^{(l)}(e), (\mu_M^{(l)}(v))_{v \in e}\big),$$

provided the right-hand sides are interpreted with a fixed ordering (or via symmetric maps) so that the definitions are unambiguous. All results below remain valid with $\mu_M$ replaced by $\mu_M^{(l)}$.

Theorem 5.4.3 (U-GNN carries an uncertain-graph structure at every layer). Let U-GNN$_M$ be as in Definition 5.4.1. Then:

1. The underlying triple $(V, E, \mu_M)$ is an uncertain graph of type $M$.
2. Moreover, if degrees are updated as in Remark 5.4.2, then for every $l = 0, \dots, L$ the triple $\mathcal{G}_M^{(l)} := (V, E, \mu_M^{(l)})$ is an uncertain graph of type $M$.

Proof. (1) This is immediate from the data of Definition 5.4.1: $(V, E)$ is a finite simple graph and $\mu_M : V \cup E \to \mathrm{Dom}(M)$ assigns to each vertex and edge a degree in the fixed domain $\mathrm{Dom}(M)$. Hence $(V, E, \mu_M)$ is an uncertain graph of type $M$.

(2) Assume degree propagation is included. By construction $\mu_M^{(0)} = \mu_M$ maps $V \cup E$ into $\mathrm{Dom}(M)$. Suppose inductively that $\mu_M^{(l)} : V \cup E \to \mathrm{Dom}(M)$. For a vertex $v \in V$, the update rule defines $\mu_M^{(l+1)}(v)$ as the value of $\mathrm{Deg}_V^{(l)}$ applied to elements of $\mathrm{Dom}(M)$; since $\mathrm{Deg}_V^{(l)}$ has codomain $\mathrm{Dom}(M)$, we obtain $\mu_M^{(l+1)}(v) \in \mathrm{Dom}(M)$. Similarly, for $e \in E$, $\mu_M^{(l+1)}(e) \in \mathrm{Dom}(M)$ because $\mathrm{Deg}_E^{(l)}$ has codomain $\mathrm{Dom}(M)$. Thus $\mu_M^{(l+1)} : V \cup E \to \mathrm{Dom}(M)$, and hence $\mathcal{G}_M^{(l+1)}$ is an uncertain graph of type $M$. By induction, the claim holds for all $l$. □

Theorem 5.4.4 (Well-definedness of the U-GNN forward pass). Let U-GNN$_M$ be as in Definition 5.4.1. Assume that for each $l = 0, \dots, L-1$:

- $\mathrm{Msg}^{(l)}$ is a function $\mathbb{R}^{d_l} \times \mathbb{R}^{d_l} \times \mathrm{Dom}(M)^3 \to \mathbb{R}^{d_l}$;
- $\mathrm{Agg}^{(l)}$ maps every finite multiset of $\mathbb{R}^{d_l}$ to an element of $\mathbb{R}^{d_l}$;
- $\mathrm{Upd}^{(l)}$ is a function $\mathbb{R}^{d_l} \times \mathbb{R}^{d_l} \to \mathbb{R}^{d_{l+1}}$;
- $\mathrm{Readout} : \mathbb{R}^{d_L} \to \mathcal{Y}$ is a function.

Then the quantities $m_v^{(l)}$, $h_v^{(l)}$, and $y_v$ defined by the recursion in Definition 5.4.1 exist and are uniquely determined for all $v \in V$ and $l = 0, \dots, L$. In particular, the U-GNN forward propagation is well-defined.

If, additionally, degree-propagation maps $\mathrm{Deg}_V^{(l)}$, $\mathrm{Deg}_E^{(l)}$ as in Remark 5.4.2 are included, then the degree updates $\mu_M^{(l)}$ are also well-defined and satisfy $\mu_M^{(l)}(x) \in \mathrm{Dom}(M)$ for all $x \in V \cup E$ and all $l$.

Proof. Fix a layer $l \in \{0, \dots, L-1\}$ and a vertex $v \in V$. For each neighbor $u \in N(v)$, the tuple $(h_v^{(l)}, h_u^{(l)}, \mu_M(v), \mu_M(u), \mu_M(\{u, v\}))$ lies in the domain of $\mathrm{Msg}^{(l)}$ because $h_v^{(l)}, h_u^{(l)} \in \mathbb{R}^{d_l}$ and $\mu_M(\cdot) \in \mathrm{Dom}(M)$ by definition of an uncertain graph. Hence each message $\mathrm{Msg}^{(l)}(\cdots) \in \mathbb{R}^{d_l}$. Since $N(v)$ is finite, the collection of messages is a finite multiset in $\mathbb{R}^{d_l}$, so $\mathrm{Agg}^{(l)}$ produces a unique aggregate vector $m_v^{(l)} \in \mathbb{R}^{d_l}$. Applying $\mathrm{Upd}^{(l)}$ yields $h_v^{(l+1)} \in \mathbb{R}^{d_{l+1}}$. Thus, given $h^{(l)}$, the next-layer features $h^{(l+1)}$ are uniquely determined.

Starting from the given initialization $h^{(0)}$, iterating the above argument for $l = 0, 1, \dots, L-1$ shows by induction that all $h_v^{(l)}$ and $m_v^{(l)}$ exist and are uniquely determined. Finally, since Readout is a function on $\mathbb{R}^{d_L}$, the outputs $y_v = \mathrm{Readout}(h_v^{(L)})$ are well-defined.

If degree propagation is included, then $\mu_M^{(0)} = \mu_M$ is given. Assuming $\mu_M^{(l)}$ is defined and valued in $\mathrm{Dom}(M)$, the update rules apply $\mathrm{Deg}_V^{(l)}$, $\mathrm{Deg}_E^{(l)}$ to arguments in $\mathrm{Dom}(M)$ and return values in $\mathrm{Dom}(M)$ by codomain specification. Hence $\mu_M^{(l+1)}$ is well-defined and $\mathrm{Dom}(M)$-valued. An induction on $l$ completes the proof. □

## 5.5 Soft Graph Neural Network

A soft graph represents a parameterized graph structure, where each parameter defines a subgraph over a fixed vertex and edge set [251, 252]. Related concepts, such as HyperSoft Graphs, are also known [253, 254]. A Soft Graph Neural Network integrates soft set theory with graph learning, representing parameterized uncertainty for flexible, context-sensitive reasoning tasks. Soft set theory has been applied in a wide range of areas, including machine learning and artificial intelligence (AI) (e.g.[255, 256, 257]).

Definition 5.5.1 (Soft Graph). [252] Let $G = (V, E)$ be a (simple undirected) graph and $C$ a nonempty set of parameters. A soft graph over $G$ with parameters $C$ is a quadruple

$$\big(G,\, C,\, A,\, B\big),$$

where

$$A : C \longrightarrow \mathcal{P}(V), \qquad B : C \longrightarrow \mathcal{P}(E),$$

such that for every $c \in C$,

$$B(c) \;\subseteq\; \big\{\{u, v\} \in E : u \in A(c),\, v \in A(c)\big\}.$$

For each $c \in C$, the pair $(A(c),\, B(c))$ is called the soft subgraph of $G$ at parameter $c$.

Definition 5.5.2 (Soft Graph Neural Network (SoftGNN)). Let $G = (V, E)$ be a finite (crisp) graph and let $A$ be a nonempty set of parameters. A soft graph over $G$ is a soft-set style family

$$\mathcal{G}_{\mathrm{soft}} = (G, F, A), \qquad F : A \to \{\text{subgraphs of } G\},$$

where for each $a \in A$ we write

$$F(a) = G_a = (V_a, E_a), \qquad V_a \subseteq V, \quad E_a \subseteq E \cap \binom{V_a}{2}.$$

(Thus each parameter $a$ selects a subgraph $G_a$ of the base graph $G$.)

Fix input node features $X \in \mathbb{R}^{|V| \times d_0}$ (row $i$ corresponds to $v_i \in V$), and set $h_i^{(0)} := X_{i\cdot} \in \mathbb{R}^{d_0}$. A Soft Graph Neural Network of depth $T$ on $\mathcal{G}_{\mathrm{soft}}$ is a message-passing architecture that, for each layer $t = 0, 1, \dots, T-1$, updates node representations $h_i^{(t)} \in \mathbb{R}^{d_t}$ by:

(1) Parameter-wise messages. For each parameter $a \in A$ and each node $v_i \in V_a$, define the neighborhood

$$\mathcal{N}_a(i) := \{\, j \in V_a \mid \{v_i, v_j\} \in E_a \,\},$$

and compute a parameter-specific aggregated message

$$m_{i,a}^{(t+1)} := \mathrm{Agg}_{j \in \mathcal{N}_a(i)}\, \phi_m^{(t)}\big(h_i^{(t)}, h_j^{(t)}, a\big) \;\in\; \mathbb{R}^{r_t},$$

where $\phi_m^{(t)}$ is a learnable message map and Agg is a permutation-invariant aggregator (e.g., sum/mean/max). If $v_i \notin V_a$, set $m_{i,a}^{(t+1)} := \mathbf{0}$.

(2) Soft (across-parameter) aggregation. Let $A_i := \{a \in A \mid v_i \in V_a\}$ be the set of parameters in which node $v_i$ participates. Choose nonnegative parameter weights $\omega_i : A_i \to [0, \infty)$ (fixed or learnable) and define

$$\overline{m}_i^{(t+1)} := \begin{cases} \dfrac{\sum_{a \in A_i} \omega_i(a)\, m_{i,a}^{(t+1)}}{\sum_{a \in A_i} \omega_i(a)}, & A_i \neq \emptyset, \\ \mathbf{0}, & A_i = \emptyset, \end{cases} \qquad \overline{m}_i^{(t+1)} \in \mathbb{R}^{r_t}.$$

(3) Node update. Update each node embedding by a learnable map $\phi_u^{(t)}$:

$$h_i^{(t+1)} := \phi_u^{(t)}\big(h_i^{(t)}, \overline{m}_i^{(t+1)}\big) \in \mathbb{R}^{d_{t+1}}.$$

Optionally, for graph-level tasks, a permutation-invariant readout $\phi_r$ produces

$$z_{\mathcal{G}_{\text{soft}}} := \phi_r\Big(\{h_i^{(T)} \mid v_i \in V\}\Big).$$

Remark. If $|A| = 1$ and $F(a) = G$, then the construction reduces to an ordinary message-passing GNN on $G$.

Definition 5.5.3 (Soft Graph Neural Network layer (soft-set/parameterized sense)). Let $V$ be a finite node set, let $A$ be a finite parameter set, and let

$$\mathcal{E} : A \to \mathcal{P}(V \times V)$$

assign to each $a \in A$ a (directed or undirected) edge set $\mathcal{E}(a)$. The triple $\mathcal{S} = (V, A, \mathcal{E})$ is called a soft graph (a parameterized family of graphs). Let $X \in \mathbb{R}^{|V| \times d_0}$ be node features and set $h_v^{(0)} := X_v \in \mathbb{R}^{d_0}$.

An $\ell$-th soft graph neural network (SGNN) layer maps $\{h_v^{(\ell)}\}_{v \in V}$ to $\{h_v^{(\ell+1)}\}_{v \in V}$ by parameter-wise message passing followed by parameter aggregation: for each $a \in A$ and $v \in V$,

$$m_{v,a}^{(\ell)} := \sum_{u \in N_a(v)} \phi^{(\ell)}\big(h_u^{(\ell)}, h_v^{(\ell)}, a\big), \qquad \tilde{h}_{v,a}^{(\ell+1)} := \sigma\Big(W_0^{(\ell)} h_v^{(\ell)} + m_{v,a}^{(\ell)}\Big),$$

where $N_a(v) := \{u \in V : (u, v) \in \mathcal{E}(a)\}$ (or $\{u : \{u, v\} \in \mathcal{E}(a)\}$ in the undirected case), $\phi^{(\ell)}$ is a learnable message function, $W_0^{(\ell)}$ is a learnable matrix, and $\sigma$ is a nonlinearity. Finally,

$$h_v^{(\ell+1)} := \operatorname{Agg}\Big(\{\tilde{h}_{v,a}^{(\ell+1)} : a \in A\}\Big),$$

where Agg is a permutation-invariant aggregation over parameters (e.g. a weighted sum with learnable weights $\pi_a \geq 0$, $\sum_{a \in A} \pi_a = 1$). Stacking layers yields an SGNN.

Example 5.5.4 (Concrete real-world example: multi-modal public transport SGNN). Consider a city that wants station embeddings for congestion prediction. Let

$$V = \{s_1, s_2, s_3, s_4\}$$

be four transit stations (e.g. Shinjuku, Shibuya, Shinagawa, Ueno), and let the parameter set represent transport modes

$$A = \{\text{rail}, \text{bus}\}.$$

Define mode-specific edge sets (a simplified toy instance):

$$\mathcal{E}(\text{rail}) = \big\{\{s_1, s_2\}, \{s_2, s_3\}\big\}, \qquad \mathcal{E}(\text{bus}) = \big\{\{s_1, s_3\}, \{s_3, s_4\}, \{s_2, s_4\}\big\}.$$

Thus, $\mathcal{S} = (V, A, \mathcal{E})$ is a soft graph encoding that rail and bus connectivity differ.

Assign each station a feature vector

$$x_v = (\text{average daily passengers at } v,\ \text{average delay rate at } v) \in \mathbb{R}^2, \qquad h_v^{(0)} := x_v.$$

Apply one SGNN layer from Definition 5.5.3 with (for illustration) a linear message $\phi^{(0)}(h_u, h_v, a) = W_a^{(0)} h_u$ depending on the mode $a$:

$$m_{v,a}^{(0)} = \sum_{u \in N_a(v)} W_a^{(0)} h_u^{(0)}, \qquad \tilde{h}_{v,a}^{(1)} = \sigma\Big(W_0^{(0)} h_v^{(0)} + m_{v,a}^{(0)}\Big).$$

For example, station $s_3$ receives rail-messages only from $s_2$ but bus-messages from $s_1$ and $s_4$:

$$m_{s_3,\text{rail}}^{(0)} = W_{\text{rail}}^{(0)} h_{s_2}^{(0)}, \qquad m_{s_3,\text{bus}}^{(0)} = W_{\text{bus}}^{(0)} h_{s_1}^{(0)} + W_{\text{bus}}^{(0)} h_{s_4}^{(0)}.$$

Finally aggregate the mode-specific embeddings, e.g. by a convex combination with learnable mode-importance weights $\pi_{\text{rail}}, \pi_{\text{bus}} \geq 0$ and $\pi_{\text{rail}} + \pi_{\text{bus}} = 1$:

$$h_{s_3}^{(1)} = \pi_{\text{rail}} \tilde{h}_{s_3,\text{rail}}^{(1)} + \pi_{\text{bus}} \tilde{h}_{s_3,\text{bus}}^{(1)}.$$

Intuitively, the network learns how strongly rail vs. bus connectivity should influence each station's representation for downstream tasks such as short-term congestion or delay propagation prediction.

Theorem 5.5.5 (Well-definedness of Soft Graph Neural Networks). Consider a soft graph $\mathcal{G}_{\text{soft}} = (G, F, A)$ as in the SoftGNN definition above, where $G = (V, E)$ is finite and $A \neq \emptyset$ is finite. Assume the following for each layer $t = 0, 1, \ldots, T-1$:

(i) (Message map) There exists a function

$$\phi_m^{(t)} : \mathbb{R}^{d_t} \times \mathbb{R}^{d_t} \times A \to \mathbb{R}^{r_t}.$$

(ii) (Neighbor aggregator) Agg is a permutation-invariant map on finite multisets in $\mathbb{R}^{r_t}$, with values in $\mathbb{R}^{r_t}$ (e.g., sum, mean, or coordinatewise max).

(iii) (Parameter weights) For each node $v_i \in V$, the weight map

$$\omega_i : A_i \to [0, \infty) \qquad \text{with} \qquad A_i := \{a \in A : \; v_i \in V_a\}$$

satisfies

$$\sum_{a \in A_i} \omega_i(a) > 0 \quad \text{whenever } A_i \neq \emptyset.$$

(iv) (Update map) There exists a function

$$\phi_u^{(t)} : \mathbb{R}^{d_t} \times \mathbb{R}^{r_t} \to \mathbb{R}^{d_{t+1}}.$$

Then the Soft Graph Neural Network recursion is well-defined: for every layer $t = 0, 1, \ldots, T-1$ and every node $v_i \in V$, the quantities

$$m_{i,a}^{(t+1)} \in \mathbb{R}^{r_t}, \qquad \overline{m}_i^{(t+1)} \in \mathbb{R}^{r_t}, \qquad h_i^{(t+1)} \in \mathbb{R}^{d_{t+1}}$$

exist uniquely and are independent of the ordering of neighbors in $\mathcal{N}_a(i)$ and of the ordering of parameters in $A_i$. Consequently, after $T$ layers, $\{h_i^{(T)}\}_{v_i \in V}$ is uniquely determined by $\mathcal{G}_{\text{soft}}$, $X$, and the learnable maps/weights. If a permutation-invariant readout $\phi_r$ is used, then $z_{\mathcal{G}_{\text{soft}}}$ is also well-defined.

Proof. We proceed by induction on the layer index $t$.

Base step ($t = 0$). By assumption, the input feature matrix $X \in \mathbb{R}^{|V| \times d_0}$ is given, so each

$$h_i^{(0)} = X_{i\cdot} \in \mathbb{R}^{d_0}$$

is well-defined.

Induction step. Assume $h_i^{(t)} \in \mathbb{R}^{d_t}$ is already well-defined for every $v_i \in V$. We show that layer $t+1$ is well-defined.

1) Parameter-wise messages $m_{i,a}^{(t+1)}$. Fix $v_i \in V$ and $a \in A$. Since $G_a = (V_a, E_a)$ is a subgraph of the finite graph $G$, the neighborhood

$$\mathcal{N}_a(i) = \{j \in V_a : \{v_i, v_j\} \in E_a\}$$

is a finite set (or empty). For each $j \in \mathcal{N}_a(i)$, the induction hypothesis gives $h_i^{(t)}, h_j^{(t)} \in \mathbb{R}^{d_t}$, and by (i),

$$\phi_m^{(t)}(h_i^{(t)}, h_j^{(t)}, a) \in \mathbb{R}^{r_t}.$$

Therefore, $\mathrm{Agg}_{j\in\mathcal{N}_a(i)}\, \phi_m^{(t)}(\cdots)$ is defined and lies in $\mathbb{R}^{r_t}$ by (ii). If $v_i \notin V_a$, the definition sets $m_{i,a}^{(t+1)} = \mathbf{0} \in \mathbb{R}^{r_t}$. Hence $m_{i,a}^{(t+1)} \in \mathbb{R}^{r_t}$ is well-defined in all cases.

Moreover, because Agg is permutation-invariant, $m_{i,a}^{(t+1)}$ does not depend on the ordering of neighbors in $\mathcal{N}_a(i)$.

2) Soft aggregation $\overline{m}_i^{(t+1)}$. Fix $v_i \in V$. The set $A_i = \{a \in A : v_i \in V_a\}$ is finite because $A$ is finite.

- If $A_i = \emptyset$, then $\overline{m}_i^{(t+1)} := \mathbf{0} \in \mathbb{R}^{r_t}$, so it is well-defined.
- If $A_i \neq \emptyset$, then by (iii) the denominator

  $$\sum_{a\in A_i} \omega_i(a)$$

  is strictly positive, so division is valid. Since each $m_{i,a}^{(t+1)} \in \mathbb{R}^{r_t}$, the weighted average

  $$\overline{m}_i^{(t+1)} = \frac{\sum_{a\in A_i} \omega_i(a)\, m_{i,a}^{(t+1)}}{\sum_{a\in A_i} \omega_i(a)}$$

  is a well-defined vector in $\mathbb{R}^{r_t}$.

This expression is independent of the ordering of parameters in $A_i$, because finite sums of vectors/scalars are order-independent.

3) Node update $h_i^{(t+1)}$. By the induction hypothesis, $h_i^{(t)} \in \mathbb{R}^{d_t}$, and by Step 2, $\overline{m}_i^{(t+1)} \in \mathbb{R}^{r_t}$. Hence, by (iv),

$$h_i^{(t+1)} = \phi_u^{(t)}(h_i^{(t)}, \overline{m}_i^{(t+1)}) \in \mathbb{R}^{d_{t+1}}$$

is well-defined and unique.

Thus all node embeddings at layer $t+1$ are well-defined. This completes the induction.

Therefore, the entire depth-$T$ SoftGNN is well-defined. If a graph-level readout

$$z_{\mathcal{G}_{\text{soft}}} = \phi_r(\{h_i^{(T)} : v_i \in V\})$$

is applied and $\phi_r$ is permutation-invariant on finite multisets, then $z_{\mathcal{G}_{\text{soft}}}$ is also well-defined and independent of the ordering of the nodes. □

Proposition 5.5.6 (Well-definedness of the layer formulation in Definition 5.5.3). Assume $V$ and $A$ are finite, and for each layer $\ell$:

- $\phi^{(\ell)}$ maps to $\mathbb{R}^{d_{\ell+1}}$ (or a compatible message space),
- $W_0^{(\ell)}$ has compatible dimensions,
- $\sigma$ is a well-defined nonlinearity on the corresponding Euclidean space, and
- Agg is permutation-invariant on finite multisets of vectors in that space.

Then the SGNN layer in Definition 5.5.3 is well-defined, and stacking such layers yields a well-defined SGNN.

Proof. For each $a \in A$ and $v \in V$, the neighbor set $N_a(v)$ is finite because $V$ is finite. Hence the sum

$$m_{v,a}^{(\ell)} = \sum_{u \in N_a(v)} \phi^{(\ell)}(h_u^{(\ell)}, h_v^{(\ell)}, a)$$

is a finite sum of vectors and is therefore well-defined. The affine map $W_0^{(\ell)} h_v^{(\ell)} + m_{v,a}^{(\ell)}$ and the nonlinearity $\sigma$ are well-defined by assumption, so $\tilde{h}_{v,a}^{(\ell+1)}$ exists. Finally, since $A$ is finite and Agg is permutation-invariant on finite multisets, the parameter aggregation

$$h_v^{(\ell+1)} = \mathrm{Agg}\big(\{\tilde{h}_{v,a}^{(\ell+1)} : a \in A\}\big)$$

is well-defined and independent of the ordering of $A$. Iterating over layers proves the claim. □

## 5.6 Rough Graph Neural Networks

Rough graphs model lower/upper approximation-based uncertainty on vertices and edges [258, 259]; Rough Graph Neural Networks learn embeddings through boundary-aware message passing for uncertain graph reasoning tasks (cf.[260]). Rough set theory has been applied in a wide range of areas, including machine learning and artificial intelligence (AI) (e.g.[261, 262, 263]).

Definition 5.6.1 (Rough Set). [264, 265] Let $U$ be a universe of discourse and $R$ a relation on $U$ that induces a partition of $U$ into equivalence classes. For a subset $X \subseteq U$, the lower approximation of $X$ with respect to $R$ (denoted $R(X)$) is the set of elements that are certainly in $X$ given the information provided by $R$:

$$R(X) = \{x \in U \mid [x]_R \subseteq X\}.$$

The upper approximation of $X$ (denoted $R(X)$) is the set of elements that possibly belong to $X$:

$$R(X) = \{x \in U \mid [x]_R \cap X \neq \emptyset\}.$$

The pair $(R(X), R(X))$ is called the Rough Set approximation of $X$ with respect to $R$.

Definition 5.6.2 (Rough Graph). [266, 267, 268] Let $G = (V, E)$ be a graph where $V$ is the set of vertices and $E$ is the set of edges. Let $R$ be an attribute set on $E$, inducing an equivalence relation on the edges. For any edge set $X \subseteq E$, the lower approximation of $X$ with respect to $R$ (denoted $R(X)$) is defined as:

$$R(X) = \{e \in E \mid [e]_R \subseteq X\},$$

where $[e]_R$ denotes the equivalence class of $e$ under $R$. The upper approximation of $X$ (denoted $R(X)$) is defined as:

$$R(X) = \{e \in E \mid [e]_R \cap X \neq \emptyset\}.$$

A graph $G = (V, E)$ is called an R-rough graph if $X$ is not exactly definable under $R$, and it is characterized by the pair $(R(X), R(X))$, where $R(X)$ is the lower approximation graph and $R(X)$ is the upper approximation graph.

Definition 5.6.3 (Rough Graph Neural Network (RGNN)). Let $G = (V, E)$ be a finite graph, let $\sim_R$ be an equivalence relation on $E$, and let $X \subseteq E$ be a designated edge subset. Let

$$E^- = \underline{R}(X), \qquad E^+ = \overline{R}(X), \qquad E^\partial = E^+ \setminus E^-$$

be the lower, upper, and boundary edge sets, respectively.

Fix input node features

$$H^{(0)} = X^{(0)} \in \mathbb{R}^{|V| \times d_0},$$

where the row $h_v^{(0)} \in \mathbb{R}^{d_0}$ corresponds to node $v \in V$. A Rough Graph Neural Network of depth $T$ is a message-passing architecture that, for each layer $\ell = 0, 1, \ldots, T-1$, updates node embeddings

$$h_v^{(\ell)} \in \mathbb{R}^{d_\ell} \qquad (v \in V)$$

using the three rough channels (certain, boundary, possible) as follows.

(1) Channel-wise neighborhoods. For each $v \in V$, define

$$\mathcal{N}^-(v) := \{u \in V : \{u,v\} \in E^-\},$$
$$\mathcal{N}^\partial(v) := \{u \in V : \{u,v\} \in E^\partial\},$$
$$\mathcal{N}^+(v) := \{u \in V : \{u,v\} \in E^+\}.$$

(2) Channel-wise messages (boundary-aware message passing). Let

$$\phi_c^{(\ell)} : \mathbb{R}^{d_\ell} \times \mathbb{R}^{d_\ell} \to \mathbb{R}^{r_\ell} \qquad (c \in \{-, \partial, +\})$$

be learnable message maps, and let $\mathrm{Agg}_c$ be permutation-invariant aggregators (e.g., sum/mean/max) on finite multisets in $\mathbb{R}^{r_\ell}$. Define

$$m_{v,c}^{(\ell+1)} := \mathrm{Agg}_c\Big(\big\{\phi_c^{(\ell)}(h_u^{(\ell)}, h_v^{(\ell)}) : u \in \mathcal{N}^c(v)\big\}\Big) \in \mathbb{R}^{r_\ell},$$

for each $c \in \{-, \partial, +\}$, with the convention that if $\mathcal{N}^c(v) = \emptyset$, then $m_{v,c}^{(\ell+1)} := \mathbf{0} \in \mathbb{R}^{r_\ell}$.

(3) Rough (cross-channel) aggregation. Let $\lambda_v^{(\ell)}(c) \geq 0$ be channel weights ($c \in \{-, \partial, +\}$), fixed or learnable, satisfying

$$\sum_{c \in \{-,\partial,+\}} \lambda_v^{(\ell)}(c) > 0 \qquad \text{for all } v \in V.$$

Define the rough aggregated message

$$\overline{m}_v^{(\ell+1)} := \frac{\lambda_v^{(\ell)}(-)\, m_{v,-}^{(\ell+1)} + \lambda_v^{(\ell)}(\partial)\, m_{v,\partial}^{(\ell+1)} + \lambda_v^{(\ell)}(+)\, m_{v,+}^{(\ell+1)}}{\lambda_v^{(\ell)}(-) + \lambda_v^{(\ell)}(\partial) + \lambda_v^{(\ell)}(+)} \in \mathbb{R}^{r_\ell}.$$

(4) Node update. Let

$$\phi_u^{(\ell)} : \mathbb{R}^{d_\ell} \times \mathbb{R}^{r_\ell} \to \mathbb{R}^{d_{\ell+1}}$$

be a learnable update map. Define

$$h_v^{(\ell+1)} := \phi_u^{(\ell)}\big(h_v^{(\ell)}, \overline{m}_v^{(\ell+1)}\big) \in \mathbb{R}^{d_{\ell+1}}.$$

(5) Optional readout. For graph-level tasks, a permutation-invariant readout $\phi_r$ yields

$$z_{G_R(X)} := \phi_r\big(\{h_v^{(T)} : v \in V\}\big).$$

The resulting architecture is called a Rough Graph Neural Network (RGNN).

Remark 5.6.4 (Special case: ordinary GNN). If the rough approximation is exact, i.e.,

$$E^- = E^+ \quad \text{and hence} \quad E^\partial = \emptyset,$$

then the boundary channel vanishes. Choosing $\lambda_v^{(\ell)}(\partial) = 0$ and $\lambda_v^{(\ell)}(-) = \lambda_v^{(\ell)}(+)$, the RGNN reduces to an ordinary message-passing GNN on the graph $(V, E^-) = (V, E^+)$.

Example 5.6.5 (RGNN on an uncertain contact network with rough edge evidence). We illustrate Definition 5.6.3 on a small contact network in which edge evidence is uncertain and grouped into equivalence classes (e.g., the same sensor modality or the same confidence category).

Step 1: Base graph and an edge equivalence relation. Let $G = (V, E)$ be an undirected graph with

$$V = \{v_1, v_2, v_3, v_4\}, \qquad E = \big\{\{v_1, v_2\}, \{v_2, v_3\}, \{v_3, v_4\}, \{v_1, v_4\}\big\}.$$

Assume an edge attribute $R$ induces an equivalence relation $\sim_R$ on $E$ with two classes:

$$C_1 = \big\{\{v_1, v_2\}, \{v_2, v_3\}\big\}, \qquad C_2 = \big\{\{v_3, v_4\}, \{v_1, v_4\}\big\}.$$

(For example, $C_1$ are contacts detected by a high-confidence sensor and $C_2$ by a low-confidence sensor.)

Step 2: Designated edge set and rough approximations. Let the designated edge subset be

$$X = \big\{\{v_1, v_2\}, \{v_3, v_4\}\big\} \subseteq E.$$

Then the lower approximation consists of equivalence classes contained in $X$, and the upper approximation consists of classes intersecting $X$:

$$E^- = \underline{R}(X) = \emptyset \quad (\text{since neither } C_1 \subseteq X \text{ nor } C_2 \subseteq X),$$

$$E^+ = \overline{R}(X) = C_1 \cup C_2 = E \quad (\text{since } C_1 \cap X \neq \emptyset,\ C_2 \cap X \neq \emptyset).$$

Hence the boundary edge set is

$$E^\partial = E^+ \setminus E^- = E.$$

Intuitively, all edges are possible under the available coarse evidence, while none is certain.

Step 3: Input node features. Take one-dimensional initial features ($d_0 = 1$):

$$h_{v_1}^{(0)} = 1, \qquad h_{v_2}^{(0)} = 0, \qquad h_{v_3}^{(0)} = 2, \qquad h_{v_4}^{(0)} = 1.$$

Step 4: One RGNN layer (simplified choices). Consider a single layer $\ell = 0$ with

$$\phi_-^{(0)}(h_u, h_v) = h_u, \qquad \phi_\partial^{(0)}(h_u, h_v) = h_u, \qquad \phi_+^{(0)}(h_u, h_v) = h_u,$$

and choose $\mathrm{Agg}_c = \sum$ (sum aggregation) for all channels. Since $E^- = \emptyset$, we have $\mathcal{N}^-(v) = \emptyset$ for all $v$, hence

$$m_{v,-}^{(1)} = 0.$$

Because $E^\partial = E^+ = E$, the boundary and possible neighborhoods coincide with the ordinary graph neighborhood:

$$\mathcal{N}^\partial(v) = \mathcal{N}^+(v) = \{u \in V : \{u, v\} \in E\}.$$

Therefore,

$$m_{v,\partial}^{(1)} = \sum_{u \in \mathcal{N}(v)} h_u^{(0)}, \qquad m_{v,+}^{(1)} = \sum_{u \in \mathcal{N}(v)} h_u^{(0)}.$$

Compute these sums:

$$\begin{aligned}
\mathcal{N}(v_1) &= \{v_2, v_4\} &\Rightarrow\quad & m_{v_1,\partial}^{(1)} = 0 + 1 = 1,\\
\mathcal{N}(v_2) &= \{v_1, v_3\} &\Rightarrow\quad & m_{v_2,\partial}^{(1)} = 1 + 2 = 3,\\
\mathcal{N}(v_3) &= \{v_2, v_4\} &\Rightarrow\quad & m_{v_3,\partial}^{(1)} = 0 + 1 = 1,\\
\mathcal{N}(v_4) &= \{v_3, v_1\} &\Rightarrow\quad & m_{v_4,\partial}^{(1)} = 2 + 1 = 3.
\end{aligned}$$

(And $m_{v,+}^{(1)} = m_{v,\partial}^{(1)}$ for each $v$.)

Choose channel weights

$$\lambda_v^{(0)}(-) = 0, \qquad \lambda_v^{(0)}(\partial) = 1, \qquad \lambda_v^{(0)}(+) = 1 \qquad (\forall v \in V),$$

so the cross-channel aggregated message becomes

$$\overline{m}_v^{(1)} = \frac{1 \cdot m_{v,\partial}^{(1)} + 1 \cdot m_{v,+}^{(1)}}{1 + 1} = m_{v,\partial}^{(1)}.$$

Finally, let the update be a simple affine map (here $d_1 = 1$):

$$\phi_u^{(0)}(h_v^{(0)}, \overline{m}_v^{(1)}) = h_v^{(0)} + \overline{m}_v^{(1)}.$$

Hence

$$h_v^{(1)} = h_v^{(0)} + \overline{m}_v^{(1)}.$$

Thus the updated embeddings are

$$h_{v_1}^{(1)} = 1 + 1 = 2, \qquad h_{v_2}^{(1)} = 0 + 3 = 3, \qquad h_{v_3}^{(1)} = 2 + 1 = 3, \qquad h_{v_4}^{(1)} = 1 + 3 = 4.$$

Interpretation. Because the evidence is entirely boundary/possible in this example ($E^- = \emptyset$), the RGNN primarily propagates information through uncertain edges, and the boundary-aware aggregation explicitly separates this uncertainty channel from the "certain" channel. This illustrates how RGNN can model coarse, class-based evidence while still performing message passing on node features.

Theorem 5.6.6 (Well-definedness of Rough Graph Neural Networks). Let $G = (V, E)$ be a finite graph, let $\sim_R$ be an equivalence relation on $E$, and let $X \subseteq E$. Assume the data and maps in Definition 5.6.3 satisfy:

(i) For each layer $\ell$, each channel message map

$$\phi_c^{(\ell)} : \mathbb{R}^{d_\ell} \times \mathbb{R}^{d_\ell} \to \mathbb{R}^{r_\ell} \quad (c \in \{-, \partial, +\})$$

is well-defined.

(ii) For each layer $\ell$ and each channel $c \in \{-, \partial, +\}$, $\mathrm{Agg}_c$ is a permutation-invariant aggregator on finite multisets in $\mathbb{R}^{r_\ell}$, with output in $\mathbb{R}^{r_\ell}$.

(iii) For each layer $\ell$ and node $v \in V$, the channel weights satisfy

$$\lambda_v^{(\ell)}(c) \geq 0 \quad (c \in \{-, \partial, +\}), \qquad \sum_{c \in \{-,\partial,+\}} \lambda_v^{(\ell)}(c) > 0.$$

(iv) For each layer $\ell$, the update map

$$\phi_u^{(\ell)} : \mathbb{R}^{d_\ell} \times \mathbb{R}^{r_\ell} \to \mathbb{R}^{d_{\ell+1}}$$

is well-defined.

Then the RGNN recursion in Definition 5.6.3 is well-defined: for every layer $\ell = 0, 1, \ldots, T-1$ and every node $v \in V$, the vectors

$$m_{v,c}^{(\ell+1)} \in \mathbb{R}^{r_\ell} \quad (c \in \{-, \partial, +\}), \qquad \overline{m}_v^{(\ell+1)} \in \mathbb{R}^{r_\ell}, \qquad h_v^{(\ell+1)} \in \mathbb{R}^{d_{\ell+1}}$$

exist uniquely and are independent of the ordering of neighbors in each $\mathcal{N}^c(v)$. If $\phi_r$ is permutation-invariant, then $z_{G_R(X)}$ is also well-defined.

Proof. We use induction on the layer index $\ell$.

Base layer ($\ell = 0$). The initial feature matrix $H^{(0)} \in \mathbb{R}^{|V| \times d_0}$ is given, so each $h_v^{(0)} \in \mathbb{R}^{d_0}$ is well-defined.

Induction step. Assume $h_v^{(\ell)} \in \mathbb{R}^{d_\ell}$ is well-defined for every $v \in V$. We show that all quantities at layer $\ell + 1$ are well-defined.

1) Well-definedness of neighborhoods. Since $V$ is finite and $E^-, E^\partial, E^+ \subseteq E$, each neighborhood $\mathcal{N}^c(v) \subseteq V$ $(c \in \{-, \partial, +\})$ is finite.

2) Well-definedness of channel-wise messages. Fix $v \in V$ and $c \in \{-, \partial, +\}$. For every $u \in \mathcal{N}^c(v)$, the induction hypothesis gives $h_u^{(\ell)}, h_v^{(\ell)} \in \mathbb{R}^{d_\ell}$. By (i),

$$\phi_c^{(\ell)}(h_u^{(\ell)}, h_v^{(\ell)}) \in \mathbb{R}^{r_\ell}.$$

Hence $\mathrm{Agg}_c$ is applied to a finite multiset of vectors in $\mathbb{R}^{r_\ell}$, so by (ii),

$$m_{v,c}^{(\ell+1)} \in \mathbb{R}^{r_\ell}$$

is well-defined. If $\mathcal{N}^c(v) = \emptyset$, the definition sets $m_{v,c}^{(\ell+1)} = \mathbf{0} \in \mathbb{R}^{r_\ell}$, which is also well-defined. Permutation invariance of $\mathrm{Agg}_c$ implies that $m_{v,c}^{(\ell+1)}$ does not depend on the ordering of $\mathcal{N}^c(v)$.

3) Well-definedness of rough aggregation. By Step 2, each $m_{v,-}^{(\ell+1)}, m_{v,\partial}^{(\ell+1)}, m_{v,+}^{(\ell+1)} \in \mathbb{R}^{r_\ell}$. By (iii), all weights are nonnegative and the denominator

$$\lambda_v^{(\ell)}(-) + \lambda_v^{(\ell)}(\partial) + \lambda_v^{(\ell)}(+)$$

is strictly positive. Therefore the weighted average

$$\overline{m}_v^{(\ell+1)} \in \mathbb{R}^{r_\ell}$$

is well-defined and unique.

4) Well-definedness of node update. Since $h_v^{(\ell)} \in \mathbb{R}^{d_\ell}$ (induction hypothesis) and $\overline{m}_v^{(\ell+1)} \in \mathbb{R}^{r_\ell}$ (Step 3), assumption (iv) implies

$$h_v^{(\ell+1)} = \phi_u^{(\ell)}(h_v^{(\ell)}, \overline{m}_v^{(\ell+1)}) \in \mathbb{R}^{d_{\ell+1}}$$

is well-defined.

Thus all node embeddings at layer $\ell + 1$ are well-defined. The induction is complete.

Finally, if a graph-level readout

$$z_{G_R(X)} = \phi_r\big(\{h_v^{(T)} : v \in V\}\big)$$

is used and $\phi_r$ is permutation-invariant on finite multisets, then $z_{G_R(X)}$ is well-defined and independent of the ordering of the nodes. □

## 5.7 Fuzzy Hypergraph Neural Network

A Fuzzy Hypergraph Neural Network integrates hypergraph neural message passing with graded (fuzzy) vertex–hyperedge incidence [196]. Equivalently, it can be viewed as a neural architecture whose underlying combinatorial object is a fuzzy hypergraph. Fuzzy hypergraphs have been studied extensively (see, e.g., [269, 270, 271]).

Definition 5.7.1 (Fuzzy hypergraph). [271] Let $V$ be a finite nonempty set of vertices. A fuzzy hyperedge on $V$ is a fuzzy subset $e$ of $V$, i.e., a map $\mu_e : V \to [0,1]$ with $\mathrm{supp}(e) := \{v \in V \mid \mu_e(v) > 0\} \neq \emptyset$. Let $\mathcal{E}$ be a finite family of fuzzy hyperedges on $V$. The pair $\mathcal{H} = (V, \mathcal{E})$ is called a fuzzy hypergraph if

$$V = \bigcup_{e \in \mathcal{E}} \mathrm{supp}(e).$$

The height of $\mathcal{H}$ is

$$h(\mathcal{H}) := \max_{e \in \mathcal{E}} \max_{v \in V} \mu_e(v).$$

For $e, f \in \mathcal{E}$, write $e \le f$ if $\mu_e(v) \le \mu_f(v)$ for all $v \in V$. The fuzzy hypergraph $\mathcal{H}$ is called

- simple if there are no repeated hyperedges in $\mathcal{E}$ and $e \le f \Rightarrow e = f$ for all $e, f \in \mathcal{E}$;
- support-simple if $e \le f$ and $\mathrm{supp}(e) = \mathrm{supp}(f)$ imply $e = f$.

Example 5.7.2 (Fuzzy hypergraph: graded team participation in projects). Let

$$V = \{\mathrm{A}, \mathrm{B}, \mathrm{C}, \mathrm{D}\}$$

be a set of four employees. We model participation in projects by fuzzy hyperedges, where $\mu_e(v) \in [0,1]$ represents the intensity of involvement of employee $v$ in project $e$.

Define three fuzzy hyperedges $e_1, e_2, e_3$ on $V$ by the membership functions:

| | A | B | C | D |
|---|---|---|---|---|
| $\mu_{e_1}(\cdot)$ | 0.9 | 0.6 | 0.2 | 0 |
| $\mu_{e_2}(\cdot)$ | 0 | 0.7 | 0.8 | 0.3 |
| $\mu_{e_3}(\cdot)$ | 0.4 | 0 | 0.5 | 0.6 |

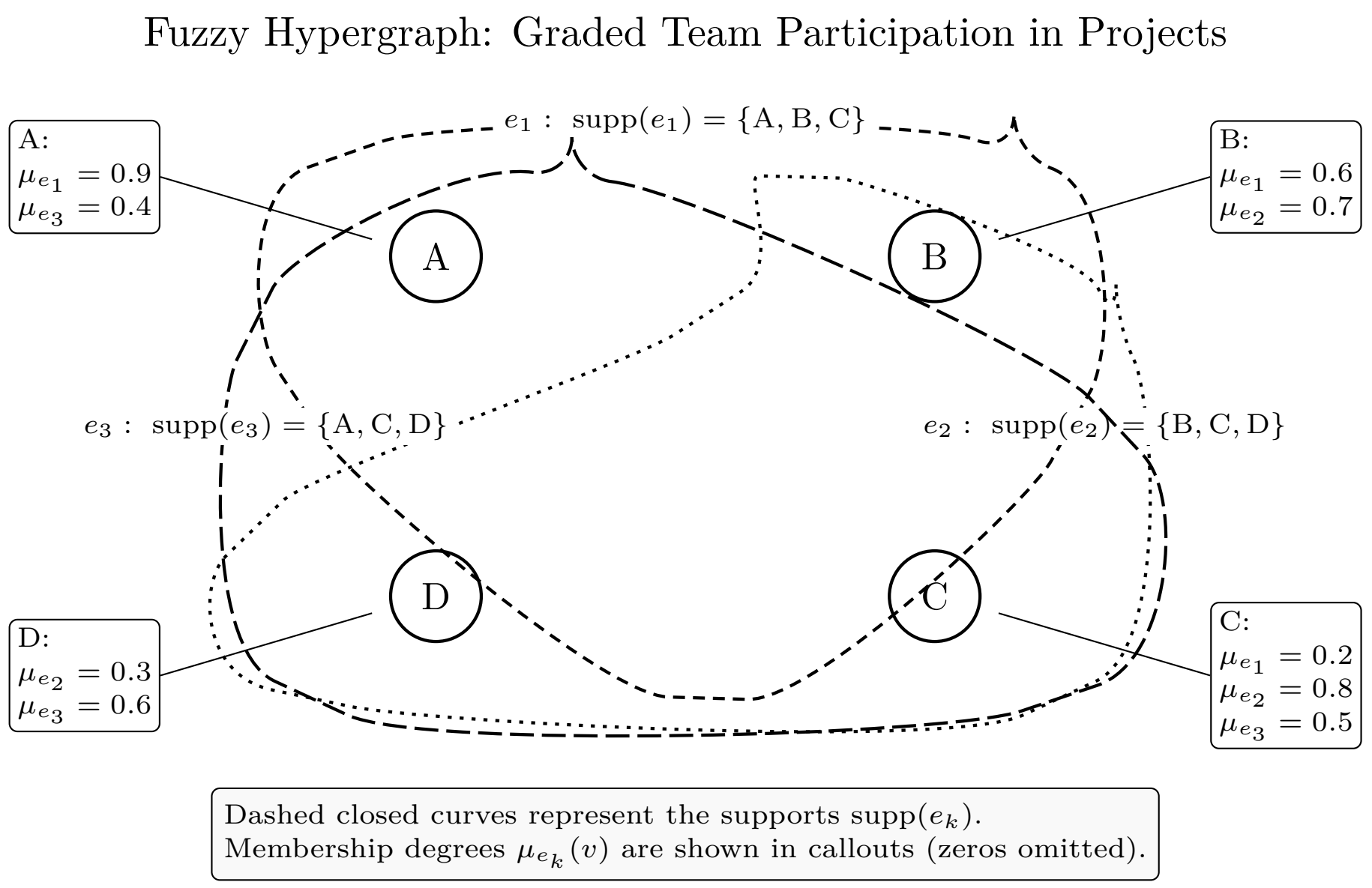

Figure 5.1: A fuzzy hypergraph with three fuzzy hyperedges $e_1, e_2, e_3$ on $V = \{\mathrm{A}, \mathrm{B}, \mathrm{C}, \mathrm{D}\}$.

Then each $e_k$ is a fuzzy hyperedge because its support is nonempty:

$$\mathrm{supp}(e_1) = \{\mathrm{A}, \mathrm{B}, \mathrm{C}\}, \quad \mathrm{supp}(e_2) = \{\mathrm{B}, \mathrm{C}, \mathrm{D}\}, \quad \mathrm{supp}(e_3) = \{\mathrm{A}, \mathrm{C}, \mathrm{D}\}.$$

Let

$$\mathcal{E} = \{e_1, e_2, e_3\}.$$

Since

$$\bigcup_{e \in \mathcal{E}} \mathrm{supp}(e) = \mathrm{supp}(e_1) \cup \mathrm{supp}(e_2) \cup \mathrm{supp}(e_3) = \{\mathrm{A}, \mathrm{B}, \mathrm{C}, \mathrm{D}\} = V,$$

the pair $\mathcal{H} = (V, \mathcal{E})$ is a fuzzy hypergraph in the sense of Definition 5.7.1.

Height. The height is the maximum membership value across all fuzzy hyperedges and vertices:

$$h(\mathcal{H}) = \max\{0.9, 0.8, 0.6, 0.7, 0.5, 0.6, 0.3, 0.4, 0.2\} = 0.9.$$

Simplicity check. There are no repeated hyperedges. Moreover, none of the three hyperedges is pointwise dominated by another: for example, $e_1 \not\leq e_2$ because $\mu_{e_1}(\mathrm{A}) = 0.9 > \mu_{e_2}(\mathrm{A}) = 0$, and $e_2 \not\leq e_1$ because $\mu_{e_2}(\mathrm{C}) = 0.8 > \mu_{e_1}(\mathrm{C}) = 0.2$. Similarly, $e_3$ is incomparable with both $e_1$ and $e_2$. Hence $\mathcal{H}$ is (in particular) simple and support-simple.

This fuzzy hypergraph captures graded, overlapping group participation: a person can belong to multiple project-hyperedges with different strengths, reflecting partial commitment. For reference, an overview schematic is provided in Fig. 5.1.

Definition 5.7.3 ($c$-level (crisp) hypergraph). Let $\mathcal{H} = (V, \mathcal{E})$ be a fuzzy hypergraph and fix $c \in (0, 1]$. For each $e \in \mathcal{E}$, the $c$-cut of $e$ is the crisp set

$$e_c := \{ v \in V \mid \mu_e(v) \geq c \}.$$

Define the $c$-level hypergraph $\mathcal{H}_c = (V_c, \mathcal{E}_c)$ by

$$V_c := \bigcup_{e \in \mathcal{E}} e_c, \qquad \mathcal{E}_c := \{ e_c \mid e \in \mathcal{E}, \ e_c \neq \emptyset \},$$

where duplicates in $\mathcal{E}_c$ may be removed (i.e., $\mathcal{E}_c$ can be viewed as a set).

Theorem 5.7.4 (Specializations). (cf.[272, 273]) A fuzzy hypergraph generalizes (i) crisp hypergraphs and (ii) fuzzy graphs (with crisp vertex set).

Proof. (i) Let $\mathcal{H}^* = (V, \mathcal{E}^*)$ be a (crisp) hypergraph. For each $e \in \mathcal{E}^*$, define $\mu_e = \mathbf{1}_e$, the indicator of $e$. Then $(V, \mathcal{E}^*)$ becomes a fuzzy hypergraph in the sense of Definition 5.7.1, and all membership values are in $\{0, 1\}$.

(ii) Let $G = (V, E)$ be a (simple) graph together with a fuzzy edge membership function $\mu : E \to (0, 1]$ (equivalently, a weighted graph with weights in $(0, 1]$). For each edge $\{u, v\} \in E$, define a fuzzy hyperedge $e_{uv}$ on $V$ by

$$\mu_{e_{uv}}(u) = \mu_{e_{uv}}(v) = \mu(\{u, v\}), \qquad \mu_{e_{uv}}(w) = 0 \ (w \in V \setminus \{u, v\}).$$

Let $\mathcal{E} := \{e_{uv} \mid \{u, v\} \in E\}$ and set $\mathcal{H} = (V, \mathcal{E})$. Then every fuzzy hyperedge has support of size 2, and $\mathcal{H}$ encodes exactly the fuzzy graph data. If the covering condition $V = \bigcup_{e \in \mathcal{E}} \operatorname{supp}(e)$ fails due to isolated vertices, one may either restrict to the non-isolated vertex set or adjoin singleton fuzzy hyperedges to cover them. □

Definition 5.7.5 (Fuzzy incidence matrix and degrees). Let $\mathcal{H} = (V, \mathcal{E})$ be a fuzzy hypergraph with $V = \{v_1, \ldots, v_n\}$ and $\mathcal{E} = \{e_1, \ldots, e_m\}$. The fuzzy incidence matrix is $H_f \in \mathbb{R}^{n \times m}$ given by

$$(H_f)_{ij} \ := \ \mu_{e_j}(v_i).$$

Let $w : \mathcal{E} \to \mathbb{R}_{\geq 0}$ be a (nonnegative) hyperedge-weight function and set $W := \operatorname{diag}(w(e_1), \ldots, w(e_m))$. Define the (weighted) fuzzy vertex degree and fuzzy hyperedge degree by

$$d_V(v_i) \ := \ \sum_{j=1}^{m} (H_f)_{ij} \, w(e_j), \qquad d_E(e_j) \ := \ \sum_{i=1}^{n} (H_f)_{ij}.$$

Let $D_V := \operatorname{diag}(d_V(v_1), \ldots, d_V(v_n))$ and $D_E := \operatorname{diag}(d_E(e_1), \ldots, d_E(e_m))$. Whenever inverses such as $D_V^{-1/2}$ or $D_E^{-1}$ are used, we assume the corresponding diagonal entries are positive (otherwise one may use a pseudo-inverse or a standard $\varepsilon$-regularization).

Theorem 5.7.6 (Crisp case). If $\mu_{e_j}(v_i) \in \{0, 1\}$ for all $i, j$, then $H_f$ is the usual (binary) incidence matrix of a crisp hypergraph.

Proof. If $\mu_{e_j}(v_i) \in \{0, 1\}$, then $(H_f)_{ij} = 1$ iff $v_i \in e_j$ and $(H_f)_{ij} = 0$ otherwise. This is exactly the classical incidence representation. □

Definition 5.7.7 (Fuzzy hypergraph Laplacian). With the notation of Definition 5.7.5, the fuzzy hypergraph Laplacian is

$$\Delta_f \ := \ I_n \ - \ D_V^{-1/2} \, H_f \, W \, D_E^{-1} \, H_f^{\top} \, D_V^{-1/2} \ \in \ \mathbb{R}^{n \times n}.$$

Theorem 5.7.8 (Laplacian generalization). The Laplacian $\Delta_f$ generalizes the (crisp) hypergraph Laplacian. In particular, if all memberships are binary, then $\Delta_f$ reduces to the standard hypergraph Laplacian (cf.[198, 197]).

Proof. Under binary memberships, $H_f$ becomes the classical incidence matrix $H$ (Theorem 5.7.6), and the degree matrices $D_V, D_E$ coincide with the usual weighted vertex-degree and hyperedge-degree matrices. Substituting $H_f = H$ into Definition 5.7.7 yields the standard formula. □

Definition 5.7.9 (Fuzzy hypergraph neural network (F-HGNN)). Let $\mathcal{H} = (V, \mathcal{E})$ be a fuzzy hypergraph with fuzzy incidence matrix $H_f$ and weights $W$. Given node features $X \in \mathbb{R}^{n \times d}$ and learnable parameters $\Theta \in \mathbb{R}^{d \times c}$, one F-HGNN convolution layer is

$$Y \ = \ \sigma\Big( D_V^{-1/2} \, H_f \, W \, D_E^{-1} \, H_f^{\top} \, D_V^{-1/2} \, X \, \Theta \Big),$$

where $\sigma$ is an activation function (e.g., ReLU) applied entrywise.

Definition 5.7.10 (Multi-layer F-HGNN). A depth-$L$ F-HGNN is obtained by stacking layers:

$$X^{(0)} := X, \qquad X^{(\ell+1)} = \sigma\Big(D_V^{-1/2}\, H_f\, W\, D_E^{-1}\, H_f^{\top}\, D_V^{-1/2}\, X^{(\ell)}\, \Theta^{(\ell)}\Big), \quad \ell = 0, \ldots, L-1.$$

Theorem 5.7.11 (Generalization of HGNN and weighted GNN). The F-HGNN framework contains the standard HGNN as a special case. Moreover, if every fuzzy hyperedge has support size at most 2, then the propagation operator $S := D_V^{-1/2} H_f W D_E^{-1} H_f^{\top} D_V^{-1/2}$ coincides with a normalized propagation matrix of a weighted (fuzzy) graph, hence the F-HGNN reduces to a (fuzzy/weighted) graph convolutional layer.

Proof. If all memberships are binary, then $H_f$ is the classical incidence matrix and Definition 5.7.9 becomes the usual HGNN convolution.

Assume now $|\operatorname{supp}(e_j)| \le 2$ for all $j$. Define a symmetric weighted adjacency $A_f \in \mathbb{R}^{n\times n}_{\ge 0}$ by

$$(A_f)_{ik} \;:=\; \sum_{j=1}^{m} \frac{w(e_j)}{d_E(e_j)}\, \mu_{e_j}(v_i)\, \mu_{e_j}(v_k).$$

Then, by direct expansion of the matrix product, $H_f W D_E^{-1} H_f^{\top} = A_f$, and therefore $S = D_V^{-1/2} A_f D_V^{-1/2}$, which is the standard normalized propagation form used for weighted graph convolutions. □

Theorem 5.7.12 (Structure preservation). An F-HGNN, as defined above, operates on a fixed fuzzy hypergraph $\mathcal{H} = (V, \mathcal{E})$: the fuzzy incidence $H_f$ (hence the membership functions $\{\mu_e\}_{e\in\mathcal{E}}$) is an input to the layer and is not altered by the network computations.

Proof. In Definitions 5.7.9–5.7.10, the only trainable objects are the parameter matrices $\{\Theta^{(\ell)}\}$, while $H_f, W, D_V, D_E$ are determined by $\mathcal{H}$ and the chosen weights. Thus the layer maps feature matrices $X^{(\ell)}$ to new feature matrices $X^{(\ell+1)}$ while keeping the underlying fuzzy hypergraph structure fixed. □

## 5.8 Uncertain Hypergraph Neural Network

Since $\operatorname{Dom}(M) \subseteq [0,1]^k$ may be vector-valued, we explicitly choose evaluation maps that convert model degrees into nonnegative scalars used by hypergraph convolution.

Definition 5.8.1 (Degree evaluation maps). Let $M$ be an uncertain model with degree-domain $\operatorname{Dom}(M)$. A pair $(\phi_V, \phi_E)$ of maps is called admissible if

$$\phi_V : \operatorname{Dom}(M) \to [0,\infty), \qquad \phi_E : \operatorname{Dom}(M) \to (0,\infty),$$

where $\phi_E$ is strictly positive. (For example, in a neutrosophic model one may take $\phi(T, I, F) = T$, or another application-dependent score.)

Definition 5.8.2 (Uncertainty-aware incidence and weights). Let $\mathcal{H}_M = (V, E, \mu_M)$ be an uncertain hypergraph of type $M$, and fix admissible evaluation maps $(\phi_V, \phi_E)$ as in Definition 5.8.1. Define:

- the vertex-weight diagonal $S_V \in \mathbb{R}^{n\times n}$ by

$$S_V \;:=\; \operatorname{diag}\big(\phi_V(\mu_M(v_1)), \ldots, \phi_V(\mu_M(v_n))\big);$$

- the uncertainty-weighted incidence matrix $\widetilde{H} \in \mathbb{R}^{n\times m}$ by

$$\widetilde{H} \;:=\; S_V\, H;$$

- the uncertainty-weight diagonal $W_M \in \mathbb{R}^{m\times m}$ by

$$W_M \;:=\; \operatorname{diag}\big(\phi_E(\mu_M(e_1)), \ldots, \phi_E(\mu_M(e_m))\big).$$

Define the associated (possibly weighted) degree diagonals

$$D_V \;:=\; \mathrm{diag}\big(d_V(v_1),\ldots,d_V(v_n)\big), \qquad D_E \;:=\; \mathrm{diag}\big(d_E(e_1),\ldots,d_E(e_m)\big),$$

where

$$d_V(v_i) \;:=\; \sum_{j=1}^{m} \widetilde{H}_{ij}\,(W_M)_{jj}, \qquad d_E(e_j) \;:=\; \sum_{i=1}^{n} \widetilde{H}_{ij}.$$

Finally, define diagonal generalized inverses by the standard convention

$$(D_V^{-1/2})_{ii} \;:=\; \begin{cases} d_V(v_i)^{-1/2}, & d_V(v_i)>0,\\ 0, & d_V(v_i)=0, \end{cases} \qquad (D_E^{-1})_{jj} \;:=\; \begin{cases} d_E(e_j)^{-1}, & d_E(e_j)>0,\\ 0, & d_E(e_j)=0. \end{cases}$$

Definition 5.8.3 (Uncertain hypergraph propagation operator). With the notation of Definition 5.8.2, define

$$\mathcal{S}_M \;:=\; D_V^{-1/2}\,\widetilde{H}\,W_M\,D_E^{-1}\,\widetilde{H}^{\top}\,D_V^{-1/2} \;\in\; \mathbb{R}^{n\times n}.$$

Definition 5.8.4 (Uncertain Hypergraph Neural Network (U-HGNN)). Let $\mathcal{H}_M = (V, E, \mu_M)$ be an uncertain hypergraph of type $M$. Fix admissible evaluation maps $(\phi_V, \phi_E)$ and let $\mathcal{S}_M$ be the propagation operator from Definition 5.8.3.

An Uncertain Hypergraph Neural Network of type $M$ (briefly, U-HGNN) with $L$ layers is specified by trainable matrices

$$\Theta^{(l)} \in \mathbb{R}^{d_l\times d_{l+1}} \qquad (l=0,\ldots,L-1),$$

an activation function $\sigma:\mathbb{R}\to\mathbb{R}$ applied entrywise, and an initial feature matrix $X^{(0)}\in\mathbb{R}^{n\times d_0}$. The forward propagation is defined by

$$X^{(l+1)} \;:=\; \sigma\Big(\mathcal{S}_M\,X^{(l)}\,\Theta^{(l)}\Big) \qquad (l=0,\ldots,L-1).$$

A task-dependent readout map $\mathrm{Readout}:\mathbb{R}^{n\times d_L}\to\mathcal{Y}$ (e.g., rowwise Softmax for node classification) produces the final output $Y := \mathrm{Readout}(X^{(L)})$.

Theorem 5.8.5 (U-HGNN carries the structure of an uncertain hypergraph). Every U-HGNN as in Definition 5.8.4 is defined on an uncertain hypergraph $\mathcal{H}_M = (V, E, \mu_M)$ of type $M$, and all uncertainty-aware operators $\widetilde{H}, W_M, D_V, D_E, \mathcal{S}_M$ are constructed functorially from $\mathcal{H}_M$ and the fixed evaluation maps $(\phi_V, \phi_E)$. In particular, the underlying uncertain-hypergraph data $(V, E, \mu_M)$ are preserved as part of the model specification.

Proof. By Definition 5.8.4, a U-HGNN takes as input (and is parametrized by) an uncertain hypergraph $\mathcal{H}_M = (V, E, \mu_M)$ of type $M$. Definition 5.8.2 constructs $\widetilde{H}$ and $W_M$ directly from $\mu_M$ via $(\phi_V, \phi_E)$, and then defines $D_V, D_E$ and their diagonal generalized inverses. Definition 5.8.3 constructs $\mathcal{S}_M$ from these matrices. No step modifies $(V, E, \mu_M)$; hence the uncertain-hypergraph structure is retained in the U-HGNN. □

Theorem 5.8.6 (U-HGNN has the structure of a hypergraph neural network). Let $\mathcal{H}_M = (V, E, \mu_M)$ be an uncertain hypergraph of type $M$. If $\phi_V \equiv 1$ on $\mathrm{Dom}(M)$ and $\phi_E$ is constant on $\mathrm{Dom}(M)$, then the U-HGNN update

$$X^{(l+1)} = \sigma\Big(\mathcal{S}_M X^{(l)}\Theta^{(l)}\Big)$$

reduces to the standard HGNN-style hypergraph convolution on the underlying crisp hypergraph $(V, E)$ (up to a constant rescaling of hyperedge weights). Consequently, U-HGNN generalizes (matrix-based) Hypergraph Neural Networks.

Proof. Assume $\phi_V \equiv 1$. Then $S_V = I_n$ and hence $\widetilde{H} = H$ (the usual incidence matrix). If $\phi_E$ is constant, then $W_M = c\,I_m$ for some constant $c > 0$, i.e., hyperedge weights are uniform up to scaling. Therefore,

$$\mathcal{S}_M = D_V^{-1/2}\,H\,(cI_m)\,D_E^{-1}\,H^{\top}\,D_V^{-1/2} = c\cdot\big(D_V^{-1/2}HD_E^{-1}H^{\top}D_V^{-1/2}\big),$$

which is exactly the standard HGNN propagation operator (with uniform hyperedge weights, up to the scalar $c$). Substituting into the layer rule yields an HGNN-style convolution. Thus U-HGNN contains HGNN as a special case and therefore has the HGNN structure. □

Theorem 5.8.7 (Well-definedness of U-HGNN forward propagation). Let $\mathcal{H}_M = (V, E, \mu_M)$ be a finite uncertain hypergraph of type $M$. Fix admissible evaluation maps $(\phi_V, \phi_E)$ and define $\mathcal{S}_M$ as in Definitions 5.8.2–5.8.3. Then:

1. All matrices $\widetilde{H}, W_M, D_V, D_E, D_V^{-1/2}, D_E^{-1}, \mathcal{S}_M$ are well-defined with compatible dimensions.

2. For any layer widths $d_0, \ldots, d_L$, any parameters $\Theta^{(l)} \in \mathbb{R}^{d_l \times d_{l+1}}$, and any input $X^{(0)} \in \mathbb{R}^{n \times d_0}$, the recursion $X^{(l+1)} = \sigma(\mathcal{S}_M X^{(l)} \Theta^{(l)})$ produces uniquely determined matrices $X^{(l)} \in \mathbb{R}^{n \times d_l}$ for all $l = 0, \ldots, L$.

Hence the U-HGNN forward pass is well-defined.

Proof. (1) Since $V$ and $E$ are finite, $H \in \{0,1\}^{n \times m}$ is well-defined. By admissibility, $\phi_V(\mu_M(v_i)) \in [0, \infty)$ and $\phi_E(\mu_M(e_j)) \in (0, \infty)$, so $S_V \in \mathbb{R}^{n \times n}$ and $W_M \in \mathbb{R}^{m \times m}$ are well-defined diagonal matrices, hence $\widetilde{H} = S_V H \in \mathbb{R}^{n \times m}$ is well-defined. The formulas for $d_V(v_i)$ and $d_E(e_j)$ are finite sums of real numbers, so $D_V, D_E$ exist as diagonal matrices. The generalized inverses $D_V^{-1/2}$ and $D_E^{-1}$ are well-defined entrywise by the stated convention, even when some degrees are zero. Therefore the product $\mathcal{S}_M = D_V^{-1/2} \widetilde{H} W_M D_E^{-1} \widetilde{H}^\top D_V^{-1/2}$ is a product of conformable matrices and is well-defined in $\mathbb{R}^{n \times n}$.

(2) Fix $l \in \{0, \ldots, L-1\}$. If $X^{(l)} \in \mathbb{R}^{n \times d_l}$, then $X^{(l)} \Theta^{(l)} \in \mathbb{R}^{n \times d_{l+1}}$ and $\mathcal{S}_M (X^{(l)} \Theta^{(l)}) \in \mathbb{R}^{n \times d_{l+1}}$. Applying $\sigma$ entrywise yields a unique matrix $X^{(l+1)} \in \mathbb{R}^{n \times d_{l+1}}$. Starting from the given $X^{(0)}$, induction on $l$ gives existence and uniqueness of all $X^{(l)}$. □

## 5.9 Uncertain n-SuperHyperGraph Neural Network

Throughout, let $\mathcal{S}_M^{(n)} = (V_n, E, \mu_M)$ be an uncertain $n$-SuperHyperGraph of type $M$ as in Definition 2.5.18. Write $N := |V_n|$ and $m := |E|$, and fix orderings

$$V_n = \{v_1, \ldots, v_N\}, \qquad E = \{e_1, \ldots, e_m\}.$$

Since each $e_j \in E$ is a nonempty subset of $V_n$, the pair $(V_n, E)$ is a (crisp) hypergraph on the vertex set $V_n$; the adjective "$n$-super" refers to the fact that each $v_i \in V_n$ is itself an element of $\mathcal{P}^n(V_0)$.

Definition 5.9.1 (Admissible degree evaluation maps). Let $M$ be an uncertain model with degree-domain $\mathrm{Dom}(M) \subseteq [0,1]^k$. A pair $(\phi_V, \phi_E)$ is called admissible if

$$\phi_V : \mathrm{Dom}(M) \to [0, \infty), \qquad \phi_E : \mathrm{Dom}(M) \to (0, \infty).$$

(Thus $\phi_E$ is strictly positive.)

Definition 5.9.2 (Incidence of an $n$-SuperHyperGraph). The (crisp) incidence matrix of $(V_n, E)$ is the matrix $H^{(n)} \in \{0,1\}^{N \times m}$ defined by

$$H_{ij}^{(n)} := \begin{cases} 1, & v_i \in e_j, \\ 0, & \text{otherwise.} \end{cases}$$

Definition 5.9.3 (Uncertainty-aware matrices). Let $\mathcal{S}_M^{(n)} = (V_n, E, \mu_M)$ be an uncertain $n$-SuperHyperGraph of type $M$ and fix admissible $(\phi_V, \phi_E)$ as in Definition 5.9.1. Define:

- the diagonal supervertex weight matrix

$$S_V^{(n)} := \mathrm{diag}\big(\phi_V(\mu_M(v_1)), \ldots, \phi_V(\mu_M(v_N))\big) \in \mathbb{R}^{N \times N};$$

- the uncertainty-weighted incidence matrix

$$\widetilde{H}^{(n)} := S_V^{(n)} H^{(n)} \in \mathbb{R}^{N \times m};$$

- the diagonal superedge weight matrix

$$W_M^{(n)} := \operatorname{diag}\big(\phi_E(\mu_M(e_1)), \ldots, \phi_E(\mu_M(e_m))\big) \in \mathbb{R}^{m\times m}.$$

Define the associated (weighted) degree diagonals

$$D_V^{(n)} := \operatorname{diag}\big(d_V^{(n)}(v_1), \ldots, d_V^{(n)}(v_N)\big), \qquad D_E^{(n)} := \operatorname{diag}\big(d_E^{(n)}(e_1), \ldots, d_E^{(n)}(e_m)\big),$$

where

$$d_V^{(n)}(v_i) := \sum_{j=1}^{m} \widetilde{H}_{ij}^{(n)} (W_M^{(n)})_{jj}, \qquad d_E^{(n)}(e_j) := \sum_{i=1}^{N} \widetilde{H}_{ij}^{(n)}.$$

To avoid division by zero, define diagonal generalized inverses by

$$\big((D_V^{(n)})^{-1/2}\big)_{ii} := \begin{cases} \big(d_V^{(n)}(v_i)\big)^{-1/2}, & d_V^{(n)}(v_i) > 0, \\ 0, & d_V^{(n)}(v_i) = 0, \end{cases} \qquad \big((D_E^{(n)})^{-1}\big)_{jj} := \begin{cases} \big(d_E^{(n)}(e_j)\big)^{-1}, & d_E^{(n)}(e_j) > 0, \\ 0, & d_E^{(n)}(e_j) = 0. \end{cases}$$

Definition 5.9.4 (Uncertain $n$-SuperHyperGraph propagation operator). With the notation of Definition 5.9.3, define

$$\mathbf{P}_M^{(n)} := (D_V^{(n)})^{-1/2}\, \widetilde{H}^{(n)}\, W_M^{(n)}\, (D_E^{(n)})^{-1}\, (\widetilde{H}^{(n)})^{\top}\, (D_V^{(n)})^{-1/2} \in \mathbb{R}^{N\times N}.$$

Definition 5.9.5 (Uncertain $n$-SuperHyperGraph Neural Network (U-$n$SHGNN)). Let $\mathcal{S}_M^{(n)} = (V_n, E, \mu_M)$ be an uncertain $n$-SuperHyperGraph of type $M$. Fix admissible evaluation maps $(\phi_V, \phi_E)$ and let $\mathbf{P}_M^{(n)}$ be as in Definition 5.9.4.

An Uncertain $n$-SuperHyperGraph Neural Network of type $M$ (briefly, U-$n$SHGNN) with $L$ layers consists of:

- layer widths $d_0, \ldots, d_L \in \mathbb{N}$;
- trainable matrices $\Theta^{(\ell)} \in \mathbb{R}^{d_\ell \times d_{\ell+1}}$ for $\ell = 0, \ldots, L-1$;
- an activation function $\sigma : \mathbb{R} \to \mathbb{R}$ applied entrywise;
- an input feature matrix $X^{(0)} \in \mathbb{R}^{N\times d_0}$ indexed by the supervertices $V_n$;
- a task-dependent readout map $\operatorname{Readout} : \mathbb{R}^{N\times d_L} \to \mathcal{Y}$.

The forward propagation is defined recursively by

$$X^{(\ell+1)} := \sigma\Big(\mathbf{P}_M^{(n)}\, X^{(\ell)}\, \Theta^{(\ell)}\Big) \qquad (\ell = 0, \ldots, L-1),$$

and the output is $Y := \operatorname{Readout}\big(X^{(L)}\big)$.

Theorem 5.9.6 (U-$n$SHGNN is built on an uncertain hypergraph structure). Let $\mathcal{S}_M^{(n)} = (V_n, E, \mu_M)$ be an uncertain $n$-SuperHyperGraph of type $M$. Then $\mathcal{S}_M^{(n)}$ canonically determines an uncertain hypergraph of type $M$ in the sense of Definition 2.5.17, namely the triple

$$\mathcal{H}_M^{(n)} := (V_n, E, \mu_M),$$

where $(V_n, E)$ is regarded as a hypergraph with vertex set $V_n$. Moreover, every U-$n$SHGNN (Definition 5.9.5) is defined functorially from this uncertain hypergraph data together with the fixed evaluation maps $(\phi_V, \phi_E)$.

Proof. By Definition 2.5.18, $E \subseteq \mathcal{P}(V_n) \setminus \{\emptyset\}$, hence $(V_n, E)$ is (by definition) a hypergraph whose vertices are the elements of $V_n$ and whose hyperedges are the elements of $E$. The same map $\mu_M : V_n \cup E \to \operatorname{Dom}(M)$ therefore makes $(V_n, E, \mu_M)$ an uncertain hypergraph of type $M$ as in Definition 2.5.17. Definitions 5.9.2–5.9.4 and Definition 5.9.5 construct all operators used by the network from $(V_n, E, \mu_M)$ and $(\phi_V, \phi_E)$ without altering this underlying data. □

Theorem 5.9.7 (U-$n$SHGNN contains HGNN as a special case). Let $\mathcal{S}_M^{(n)} = (V_n, E, \mu_M)$ be an uncertain $n$-SuperHyperGraph of type $M$. If $\phi_V \equiv 1$ on $\mathrm{Dom}(M)$ and $\phi_E$ is constant on $\mathrm{Dom}(M)$, then the propagation operator $\mathbf{P}_M^{(n)}$ reduces (up to a global scalar) to the standard HGNN-style hypergraph propagation operator on the underlying hypergraph $(V_n, E)$. Consequently, the U-$n$SHGNN layer update $X^{(\ell+1)} = \sigma(\mathbf{P}_M^{(n)} X^{(\ell)} \Theta^{(\ell)})$ reduces to a standard HGNN update on $(V_n, E)$.

Proof. If $\phi_V \equiv 1$, then $S_V^{(n)} = I_N$, hence $\widetilde{H}^{(n)} = H^{(n)}$. If $\phi_E$ is constant, then $W_M^{(n)} = c\, I_m$ for some $c > 0$. Substituting into Definition 5.9.4 yields

$$\mathbf{P}_M^{(n)} = (D_V^{(n)})^{-1/2}\, H^{(n)}\, (cI_m)\, (D_E^{(n)})^{-1}\, (H^{(n)})^\top\, (D_V^{(n)})^{-1/2} = c\cdot (D_V^{(n)})^{-1/2} H^{(n)} (D_E^{(n)})^{-1} (H^{(n)})^\top (D_V^{(n)})^{-1/2},$$

which is exactly the standard normalized hypergraph propagation operator (with uniform hyperedge weights), up to the scalar factor $c$. Inserting this into the layer rule gives the usual HGNN-style convolution on $(V_n, E)$. □

Theorem 5.9.8 (Well-definedness of U-$n$SHGNN). Let $\mathcal{S}_M^{(n)} = (V_n, E, \mu_M)$ be a finite uncertain $n$-SuperHyperGraph of type $M$. Fix admissible evaluation maps $(\phi_V, \phi_E)$ and construct $\mathbf{P}_M^{(n)}$ as in Definitions 5.9.3–5.9.4. Then:

1. all matrices $H^{(n)}, \widetilde{H}^{(n)}, S_V^{(n)}, W_M^{(n)}, D_V^{(n)}, D_E^{(n)}, \mathbf{P}_M^{(n)}$ are well-defined with compatible dimensions;
2. for any widths $d_0, \ldots, d_L$, any parameters $\Theta^{(\ell)} \in \mathbb{R}^{d_\ell \times d_{\ell+1}}$, any input $X^{(0)} \in \mathbb{R}^{N \times d_0}$, and any entrywise activation $\sigma$, the recursion

$$X^{(\ell+1)} = \sigma\Big(\mathbf{P}_M^{(n)} X^{(\ell)} \Theta^{(\ell)}\Big) \qquad (\ell = 0, \ldots, L-1)$$

   produces uniquely determined matrices $X^{(\ell)} \in \mathbb{R}^{N \times d_\ell}$ for all $\ell = 0, \ldots, L$.

Hence the U-$n$SHGNN forward pass is well-defined.

Proof. (1) Since $V_n$ and $E$ are finite, the incidence matrix $H^{(n)} \in \{0,1\}^{N \times m}$ is well-defined. Admissibility gives $\phi_V(\mu_M(v_i)) \in [0, \infty)$ and $\phi_E(\mu_M(e_j)) \in (0, \infty)$, so $S_V^{(n)} \in \mathbb{R}^{N \times N}$ and $W_M^{(n)} \in \mathbb{R}^{m \times m}$ exist as diagonal matrices and $\widetilde{H}^{(n)} = S_V^{(n)} H^{(n)} \in \mathbb{R}^{N \times m}$ is well-defined. The degree values $d_V^{(n)}(v_i)$ and $d_E^{(n)}(e_j)$ are finite sums of real numbers, hence $D_V^{(n)}$ and $D_E^{(n)}$ exist as diagonal matrices. The generalized inverses $(D_V^{(n)})^{-1/2}$ and $(D_E^{(n)})^{-1}$ are well-defined entrywise by the stated convention, even if some degrees vanish. Therefore the matrix product defining $\mathbf{P}_M^{(n)}$ is conformable and yields an element of $\mathbb{R}^{N \times N}$.

(2) Fix $\ell \in \{0, \ldots, L-1\}$ and assume $X^{(\ell)} \in \mathbb{R}^{N \times d_\ell}$. Then $X^{(\ell)}\Theta^{(\ell)} \in \mathbb{R}^{N \times d_{\ell+1}}$ and $\mathbf{P}_M^{(n)}(X^{(\ell)}\Theta^{(\ell)}) \in \mathbb{R}^{N \times d_{\ell+1}}$. Applying $\sigma$ entrywise produces a unique matrix $X^{(\ell+1)} \in \mathbb{R}^{N \times d_{\ell+1}}$. Induction on $\ell$ yields existence and uniqueness of $X^{(\ell)}$ for all layers. □

Corollary 5.9.9 (Reduction to uncertain hypergraph neural networks). If $n = 0$ and $V_n = V_0$, then $\mathcal{S}_M^{(0)} = (V_0, E, \mu_M)$ is an uncertain hypergraph of type $M$, and the U-0SHGNN construction coincides with the usual uncertainty-aware hypergraph neural network defined on $(V_0, E, \mu_M)$ (i.e., an uncertain hypergraph neural network).

Proof. For $n = 0$, $\mathcal{P}^0(V_0) = V_0$, hence $V_n = V_0$ and $(V_n, E) = (V_0, E)$ is an ordinary hypergraph. All matrices and updates in Definitions 5.9.2–5.9.5 then coincide with the corresponding uncertain hypergraph constructions on $(V_0, E, \mu_M)$. □

**Algorithm 5:** Plithogenic Graph Neural Network (P-GNN): a plithogenic-weighted normalized message passing

**Input:** A (simple) graph $G = (V, E)$ with $n := |V|$;
node-feature matrix $X \in \mathbb{R}^{n \times d}$ (row $x_v$ for $v \in V$);
(optional) edge-feature map $z : E \to \mathbb{R}^{d_e}$;
finite attribute-value sets $A_V, A_E$;
plithogenic fuzzifiers $\mathcal{P}_V : \mathbb{R}^d \to [0,1]^{A_V}$, $\mathcal{P}_E : \mathbb{R}^{d_e} \to [0,1]^{A_E}$;
contradiction functions $\kappa_V : A_V \times A_V \to [0,1]$, $\kappa_E : A_E \times A_E \to [0,1]$;
number of layers $L$;
learnable matrices $\{\Theta^{(\ell)}\}_{\ell=0}^{L-1}$ with $\Theta^{(\ell)} \in \mathbb{R}^{d_\ell \times d_{\ell+1}}$ and $d_0 = d$;
activation $\sigma$ (entrywise);
readout $\rho$ (e.g. row-wise Softmax).
**Output:** Predictions $Y$ (e.g. $Y \in \mathbb{R}^{n \times C}$).

(1) Plithogenic appurtenance on vertices
foreach $v \in V$ do
    $\alpha_V(v, \cdot) \leftarrow \mathcal{P}_V(x_v) \in [0,1]^{A_V}$
    $s_V(v) \leftarrow \max_{a \in A_V} \alpha_V(v, a) \in [0,1]$
    choose $a^\star(v) \in \arg\max_{a \in A_V} \alpha_V(v, a)$
end
(2) Optional appurtenance on edges
if edge features are available then
    foreach edge $e \in E$ do
        $\alpha_E(e, \cdot) \leftarrow \mathcal{P}_E(z_e) \in [0,1]^{A_E}$
        $s_E(e) \leftarrow \max_{b \in A_E} \alpha_E(e, b) \in [0,1]$
    end
end
else
    set $s_E(e) \leftarrow 1$ for all $e \in E$
end
(3) Build a plithogenic weighted adjacency
Initialize a sparse matrix $\widetilde{A} \in \mathbb{R}_{\geq 0}^{n \times n}$ as all zeros
foreach edge $e = \{u, v\} \in E$ do
    $c_{uv} \leftarrow \kappa_V\big(a^\star(u), a^\star(v)\big) \in [0,1]$
    // Plithogenic blending: convex combination of t-norm and t-conorm (here min, max)
    $w_{uv} \leftarrow s_E(e)\Big((1 - c_{uv}) \min\{s_V(u), s_V(v)\} + c_{uv} \max\{s_V(u), s_V(v)\}\Big) \in [0,1]$
    $\widetilde{A}_{uv} \leftarrow w_{uv}$, $\widetilde{A}_{vu} \leftarrow w_{uv}$
end
(4) Normalization
Compute $\widetilde{D} \in \mathbb{R}^{n \times n}$ with $\widetilde{D}_{uu} = \sum_{v=1}^{n} \widetilde{A}_{uv}$
(Optionally add self-loops: $\widetilde{A} \leftarrow \widetilde{A} + I_n$ and recompute $\widetilde{D}$)
Set $P \leftarrow \widetilde{D}^{-1/2} \widetilde{A} \widetilde{D}^{-1/2}$
(5) Layer-wise propagation
Set $H^{(0)} \leftarrow X$
for $\ell \leftarrow 0$ to $L - 1$ do
    $H^{(\ell+1)} \leftarrow \sigma\big(P\, H^{(\ell)}\, \Theta^{(\ell)}\big)$
end
(6) Readout
$Y \leftarrow \rho\big(H^{(L)}\big)$
return $Y$

# 6 Other SuperHyperGraph Concepts

In this section, we explore concepts related to SuperHyperGraphs that are not directly connected to the topics discussed above.

## 6.1 Multilevel $k$-way Hypergraph Partitioning

Multilevel graph partitioning is a standard strategy to divide a graph into $k$ parts by coarsening, initial partitioning, and uncoarsening/refinement [274, 275]. In hypergraph theory, multilevel hypergraph partitioning and multilevel $k$-way hypergraph partitioning are widely studied, especially due to applications in VLSI design [276, 277, 278, 279, 280]. In this section we formalize multilevel $k$-way partitioning for hypergraphs and then extend it, in a mathematically consistent way, to $n$-SuperHyperGraphs via the expanded-hypergraph construction.

Definition 6.1.1 ($k$-way vertex partition and balance constraint). Let $\mathcal{H} = (\mathcal{V}, \mathcal{E})$ be a finite hypergraph. A $k$-way partition of $\mathcal{V}$ is a family $\Pi = \{\mathcal{V}_1, \dots, \mathcal{V}_k\}$ such that $\bigcup_{i=1}^{k} \mathcal{V}_i = \mathcal{V}$ and $\mathcal{V}_i \cap \mathcal{V}_j = \emptyset$ for $i \neq j$. If vertex weights $\omega : \mathcal{V} \to \mathbb{R}_{>0}$ are given, write $\omega(\mathcal{S}) = \sum_{v \in \mathcal{S}} \omega(v)$ and $\Omega = \omega(\mathcal{V})$. For an imbalance factor $c \geq 1$, the partition $\Pi$ is $c$-balanced if

$$\frac{\Omega}{kc} \;\leq\; \omega(\mathcal{V}_i) \;\leq\; c\,\frac{\Omega}{k} \qquad (i = 1, \dots, k).$$

(For unweighted vertices, take $\omega \equiv 1$, so $\omega(\mathcal{V}_i) = |\mathcal{V}_i|$.)

Definition 6.1.2 (Spanning number of a hyperedge). Let $\Pi = \{\mathcal{V}_1, \dots, \mathcal{V}_k\}$ be a $k$-way partition of $\mathcal{V}$. For a hyperedge $e \in \mathcal{E}$ define the spanning number (or connectivity)

$$\lambda_e(\Pi) \;:=\; \big|\{\, i \in \{1, \dots, k\} \mid e \cap \mathcal{V}_i \neq \emptyset \,\}\big|.$$

A hyperedge is cut if $\lambda_e(\Pi) > 1$.

Definition 6.1.3 (Common $k$-way cut objectives). Let $\mathcal{H} = (\mathcal{V}, \mathcal{E})$ be a hypergraph with hyperedge weights $w : \mathcal{E} \to \mathbb{R}_{>0}$, and let $\Pi$ be a $k$-way partition of $\mathcal{V}$. Common objective functions include:

$$\begin{aligned}
f_{\mathrm{cut}}(\Pi) &:= \sum_{e \in \mathcal{E}} w(e)\, \mathbf{1}[\lambda_e(\Pi) > 1] && \text{(cut-net / min-cut)},\\
f_{k-1}(\Pi) &:= \sum_{e \in \mathcal{E}} (\lambda_e(\Pi) - 1)\, w(e)\, \mathbf{1}[\lambda_e(\Pi) > 1] && \text{($k-1$ metric / connectivity)},\\
f_{\mathrm{SOED}}(\Pi) &:= \sum_{e \in \mathcal{E}} \lambda_e(\Pi)\, w(e)\, \mathbf{1}[\lambda_e(\Pi) > 1] && \text{(SOED: sum of external degrees)}.
\end{aligned}$$

Definition 6.1.4 (Multilevel $k$-way hypergraph partitioning). (cf. [278]) Given a hypergraph $\mathcal{H} = (\mathcal{V}, \mathcal{E})$, an integer $k \geq 2$, an imbalance factor $c \geq 1$, and a chosen objective $f$ (e.g., Definition 6.1.3), multilevel $k$-way hypergraph partitioning seeks a $c$-balanced $k$-way partition $\Pi = \{\mathcal{V}_1, \dots, \mathcal{V}_k\}$ minimizing $f(\Pi)$.

A multilevel method typically proceeds in three phases:

- Coarsening: build a hierarchy $\mathcal{H}_0, \mathcal{H}_1, \dots, \mathcal{H}_\ell$ with $\mathcal{H}_0 = \mathcal{H}$, where $\mathcal{H}_{i+1}$ is obtained from $\mathcal{H}_i$ by merging (clustering) vertices so that $|\mathcal{V}_{i+1}| \ll |\mathcal{V}_i|$.
- Initial partitioning: compute a $k$-way partition of the smallest hypergraph $\mathcal{H}_\ell$.
- Uncoarsening/refinement: project the partition back through the hierarchy and refine it at each level to reduce $f$ while maintaining the balance constraint.

Definition 6.1.5 (Expansion map for $n$-SuperHyperGraph objects). Let $V_0$ be a finite base vertex set and fix $n \in \mathbb{N}$. Define $\mathcal{P}^0(V_0) = V_0$ and $\mathcal{P}^{k+1}(V_0) = \mathcal{P}(\mathcal{P}^k(V_0))$. Define the expansion map $\mathrm{Expand}_n : \mathcal{P}^n(V_0) \to \mathcal{P}(V_0)$ recursively by

$$\mathrm{Expand}_n(x) = \begin{cases} \{x\}, & x \in V_0, \\ \bigcup_{u \in x} \mathrm{Expand}_n(u), & x \in \mathcal{P}^n(V_0) \setminus V_0. \end{cases}$$

If $e$ is a finite set of such objects (e.g., a superedge), define

$$\mathrm{Expand}_n(e) \;:=\; \bigcup_{v \in e} \mathrm{Expand}_n(v) \;\subseteq\; V_0.$$

Definition 6.1.6 (Expanded hypergraph of an $n$-SuperHyperGraph). Let $H = (V, E)$ be an $n$-SuperHyperGraph with base set $V_0$ and assume $E$ is a finite family of nonempty superedges, each superedge $e \in E$ being a finite nonempty set of elements of $\mathcal{P}^n(V_0)$. Given a weight function $w : E \to \mathbb{R}_{>0}$, define the expanded hypergraph

$$\mathcal{H}^{\exp}(H) \;:=\; (V_0, E^{\exp}, w^{\exp})$$

by

$$E^{\exp} \;:=\; \{\mathrm{Expand}_n(e) \mid e \in E\} \setminus \{\emptyset\}, \qquad w^{\exp}(e^{\exp}) \;:=\; \sum_{\substack{e \in E \\ \mathrm{Expand}_n(e) = e^{\exp}}} w(e).$$

(Thus duplicate expanded hyperedges are merged by summing weights.)

Definition 6.1.7 (Multilevel $k$-way $n$-SuperHyperGraph partitioning). Let $H = (V, E)$ be an $n$-SuperHyperGraph with base set $V_0$, let $k \geq 2$ and $c \geq 1$. Choose a hypergraph objective $f$ (Definition 6.1.3) and apply it to the expanded hypergraph $\mathcal{H}^{\exp}(H) = (V_0, E^{\exp}, w^{\exp})$.

A multilevel $k$-way $n$-SuperHyperGraph partitioning of $H$ is a multilevel $k$-way hypergraph partitioning (Definition 6.1.4) of the expanded hypergraph $\mathcal{H}^{\exp}(H)$; i.e., it produces a $c$-balanced partition $\Pi = \{V_{0,1}, \ldots, V_{0,k}\}$ of the base vertex set $V_0$ minimizing $f(\Pi)$ computed with respect to $(V_0, E^{\exp}, w^{\exp})$.

Theorem 6.1.8 (Reduction to standard multilevel $k$-way hypergraph partitioning). Multilevel $k$-way $n$-SuperHyperGraph partitioning (Definition 6.1.7) generalizes multilevel $k$-way hypergraph partitioning (Definition 6.1.4). More precisely, if $\mathcal{H} = (V_0, \mathcal{E}, w)$ is a hypergraph and we view it as an $n$-SuperHyperGraph $H = (V, E)$ by taking $V = V_0$ (embedded as singletons in $\mathcal{P}^n(V_0)$) and $E = \mathcal{E}$, then $\mathcal{H}^{\exp}(H) = \mathcal{H}$ and both partitioning problems coincide.

Proof. Embed each base vertex $v \in V_0$ as itself (equivalently, as the singleton object at level 0), and let each hyperedge $e \in \mathcal{E}$ be treated as a superedge containing only base vertices. Then for every $v \in V_0$ we have $\mathrm{Expand}_n(v) = \{v\}$, and hence for every $e \in \mathcal{E}$,

$$\mathrm{Expand}_n(e) = \bigcup_{v \in e} \mathrm{Expand}_n(v) = \bigcup_{v \in e} \{v\} = e.$$

Therefore $E^{\exp} = \mathcal{E}$ and $w^{\exp} = w$ (up to the harmless convention of merging duplicates), so $\mathcal{H}^{\exp}(H) = (V_0, \mathcal{E}, w) = \mathcal{H}$.

Consequently, the feasible set (all $c$-balanced $k$-way partitions of $V_0$) and the objective values computed via $\lambda_e(\Pi)$ agree in both formalisms. Hence the multilevel $k$-way $n$-SuperHyperGraph partitioning problem reduces exactly to the standard multilevel $k$-way hypergraph partitioning problem. □

## 6.2 SuperHyperGraph Random Walk

A (graph) random walk is a discrete-time Markov chain whose transition probabilities are determined by the edge structure of the graph, and it is a standard tool for modeling stochastic processes on networks [281, 282]. This idea has been extended to hypergraphs, yielding hypergraph random walks [283, 284]. In this subsection we formulate a mathematically consistent random-walk model for $n$-SuperHyperGraphs.

Definition 6.2.1 (Discrete-time Markov chain). (cf. [285, 286, 287]) Let $S$ be a finite or countable set. A stochastic process $\{X_t\}_{t\geq 0}$ with $X_t \in S$ is a discrete-time Markov chain if there exists a transition matrix $P = (p_{xy})_{x,y\in S}$ such that for all $t \geq 0$ and all $x, y \in S$,

$$\Pr(X_{t+1} = y \mid X_t = x, X_{t-1}, \ldots, X_0) = \Pr(X_{t+1} = y \mid X_t = x) = p_{xy},$$

and each row of $P$ is stochastic:

$$p_{xy} \geq 0, \qquad \sum_{y\in S} p_{xy} = 1 \qquad (\forall x \in S).$$

An initial distribution $\pi_0(x) = \Pr(X_0 = x)$ completes the specification.

Definition 6.2.2 (Weighted hypergraph random walk). [283, 288] Let $\mathcal{H} = (V, E)$ be a finite hypergraph with positive hyperedge weights $\omega : E \to \mathbb{R}_{>0}$. Assume that every vertex has at least one incident hyperedge:

$$E(v) := \{e \in E \mid v \in e\} \neq \emptyset \qquad (\forall v \in V).$$

Fix also, for each $e \in E$, a strictly positive within-edge weight function $\gamma_e : e \to \mathbb{R}_{>0}$ (uniform choice corresponds to $\gamma_e \equiv 1$).

A hypergraph random walk on $\mathcal{H}$ is the Markov chain on state space $V$ defined by: from the current vertex $v \in V$,

1. select an incident hyperedge $e \in E(v)$ with probability

$$\Pr(e \mid v) = \frac{\omega(e)}{\sum_{e'\in E(v)} \omega(e')},$$

2. then select the next vertex $u \in e$ with probability

$$\Pr(u \mid e) = \frac{\gamma_e(u)}{\sum_{x\in e} \gamma_e(x)}.$$

Hence the one-step transition probabilities are

$$p_{vu} := \Pr(X_{t+1} = u \mid X_t = v) = \sum_{\substack{e\in E \\ v\in e,\ u\in e}} \frac{\omega(e)}{\sum_{e'\in E(v)} \omega(e')} \cdot \frac{\gamma_e(u)}{\sum_{x\in e} \gamma_e(x)}.$$

Definition 6.2.3 ($n$-SuperHyperGraph). Let $V_0$ be a finite base set and fix $n \in \mathbb{N}$. Write $\mathcal{P}^0(V_0) = V_0$ and $\mathcal{P}^{k+1}(V_0) = \mathcal{P}(\mathcal{P}^k(V_0))$. An $n$-SuperHyperGraph is a pair $H = (V, E)$ such that

$$V \subseteq \mathcal{P}^n(V_0) \setminus \{\emptyset\}, \qquad E \subseteq \mathcal{P}(V) \setminus \{\emptyset\}.$$

Elements of $V$ are called supervertices, and elements of $E$ are superedges.

Definition 6.2.4 ($n$-SuperHyperGraph random walk). Let $H = (V, E)$ be a finite $n$-SuperHyperGraph with positive superedge weights $\omega : E \to \mathbb{R}_{>0}$. Assume every supervertex has at least one incident superedge:

$$E(v) := \{e \in E \mid v \in e\} \neq \emptyset \qquad (\forall v \in V).$$

Fix also strictly positive within-superedge weights $\gamma_e : e \to \mathbb{R}_{>0}$ for each $e \in E$.

An $n$-SuperHyperGraph random walk on $H$ is the discrete-time process $\{X_t\}_{t\geq 0}$ on state space $V$ defined by the two-stage sampling rule: from the current supervertex $v \in V$,

1. select an incident superedge $e \in E(v)$ with probability

$$\Pr(e \mid v) = \frac{\omega(e)}{\sum_{e' \in E(v)} \omega(e')},$$

2. then select the next supervertex $u \in e$ with probability

$$\Pr(u \mid e) = \frac{\gamma_e(u)}{\sum_{x \in e} \gamma_e(x)}.$$

Equivalently, the one-step transition probabilities are

$$p_{vu} := \Pr(X_{t+1} = u \mid X_t = v) = \sum_{\substack{e \in E \\ v \in e,\ u \in e}} \frac{\omega(e)}{\sum_{e' \in E(v)} \omega(e')} \cdot \frac{\gamma_e(u)}{\sum_{x \in e} \gamma_e(x)}.$$

Theorem 6.2.5 (Markov property). The process in Definition 6.2.4 is a discrete-time Markov chain on the state space $V$ with transition matrix $P = (p_{vu})_{v,u \in V}$.

Proof. By construction, the conditional distribution of $X_{t+1}$ given $X_t = v$ is determined solely by: (i) choosing $e \in E(v)$ according to $\Pr(e \mid v)$ and (ii) choosing $u \in e$ according to $\Pr(u \mid e)$. Hence $\Pr(X_{t+1} = u \mid X_t = v, X_{t-1}, \dots, X_0) = \Pr(X_{t+1} = u \mid X_t = v) = p_{vu}$.

Moreover, $p_{vu} \geq 0$ is immediate. For fixed $v \in V$, summing over $u \in V$ yields

$$\sum_{u \in V} p_{vu} = \sum_{e \in E(v)} \Pr(e \mid v) \sum_{u \in e} \Pr(u \mid e) = \sum_{e \in E(v)} \Pr(e \mid v) \cdot 1 = 1,$$

so each row of $P$ sums to 1. Therefore $\{X_t\}$ is a discrete-time Markov chain. □

Theorem 6.2.6 (Support of transitions respects superedges). If $p_{vu} > 0$, then there exists a superedge $e \in E$ such that $v \in e$ and $u \in e$. Equivalently, the walk moves only between supervertices that co-occur in at least one superedge.

Proof. If $p_{vu} > 0$, then at least one summand in Definition 6.2.4 is strictly positive. This requires the existence of a superedge $e \in E$ with $v \in e$ and $u \in e$. □

Theorem 6.2.7 (Generalization of hypergraph random walks). The $n$-SuperHyperGraph random walk generalizes the hypergraph random walk. More precisely, if $H = (V, E)$ is an $n$-SuperHyperGraph such that $V = V_0$ (embedded as base-level vertices) and $E \subseteq \mathcal{P}(V_0) \setminus \{\emptyset\}$, then Definition 6.2.4 reduces to the hypergraph random walk in Definition 6.2.2.

Proof. Under the stated assumption, the state space $V$ is exactly the vertex set $V_0$ and each superedge is an ordinary hyperedge (a subset of $V_0$). The incident-edge sets $E(v)$ and the two-stage sampling rule coincide with Definition 6.2.2, hence the transition probabilities agree. □

Corollary 6.2.8 (Reduction to graph random walks). If, in addition to the assumptions of Theorem 6.2.7, every hyperedge has size 2, then the walk reduces to a (possibly weighted) random walk on an ordinary graph.

Remark 6.2.9 (Base-vertex random walk via expansion). If one prefers a random walk on the base vertices $V_0$ even when supervertices are nested, one can first form an expanded hypergraph (by recursively expanding supervertices into elements of $V_0$) and then apply Definition 6.2.2 on that expanded hypergraph. This yields a base-level stochastic process induced by the superhypergraph structure.

## 6.3 Superhypergraph Turán Problem

The classical hypergraph Turán problem asks for the maximum number of edges in an $r$-uniform hypergraph on $n$ vertices that avoids a fixed forbidden $r$-uniform hypergraph $F$ as a (labeled) subhypergraph (cf. [289, 290, 291]). In this subsection we give a mathematically consistent formulation of an analogous problem for $n$-superhypergraphs (supervertices drawn from an iterated power set). Throughout, $r \geq 2$ is an integer.

Definition 6.3.1 (Forbidden subgraph/subhypergraph). Let $F = (V(F), E(F))$ and $G = (V(G), E(G))$ be (hyper)graphs of the same uniformity. We say that $G$ contains $F$ as a subhypergraph if there exists an injective map $\phi : V(F) \to V(G)$ such that for every edge $e \in E(F)$ we have

$$\phi(e) \ := \ \{\phi(v) \mid v \in e\} \in E(G).$$

If no such $\phi$ exists, then $G$ is called $F$-free.

Definition 6.3.2 ($r$-uniform hypergraph Turán number and density). Let $F$ be an $r$-uniform hypergraph. The Turán number $\mathrm{ex}_r(n, F)$ is

$$\mathrm{ex}_r(n, F) \ := \ \max\bigl\{|E(G)| : \ G \text{ is an } F\text{-free } r\text{-uniform hypergraph with } |V(G)| = n\bigr\}.$$

The Turán density is defined by

$$\pi(F) \ := \ \lim_{n\to\infty} \frac{\mathrm{ex}_r(n, F)}{\binom{n}{r}},$$

whenever the limit is taken as the standard Turán density (for $r$-uniform hypergraphs this limit is known to exist).

Definition 6.3.3 ($n$-SuperHypergraph). Fix a finite base set $V_0$ and $n \in \mathbb{N}_0$. Write $\mathcal{P}^0(V_0) = V_0$ and $\mathcal{P}^{k+1}(V_0) = \mathcal{P}(\mathcal{P}^k(V_0))$. An $n$-superhypergraph is a pair $H = (V, E)$ where

$$V \subseteq \mathcal{P}^n(V_0) \setminus \{\emptyset\} \quad \text{and} \quad E \subseteq \mathcal{P}(V) \setminus \{\emptyset\}.$$

Elements of $V$ are called supervertices and elements of $E$ are called superedges.

Definition 6.3.4 ($r$-uniform $n$-SuperHypergraph). An $n$-superhypergraph $H = (V, E)$ is $r$-uniform if every superedge $e \in E$ has cardinality $r$, i.e., $|e| = r$ for all $e \in E$.

Definition 6.3.5 ($n$-SuperHypergraph Turán number and density). Let $F$ be an $r$-uniform $n$-superhypergraph. For $N \in \mathbb{N}$, define

$$\mathrm{ex}_r^{(n)}(N, F) \ := \ \max\bigl\{|E(G)| : \ G \text{ is an } F\text{-free } r\text{-uniform } n\text{-superhypergraph with } |V(G)| = N\bigr\}.$$

A natural density normalization (by the number of all $r$-subsets of an $N$-element vertex set) is

$$\pi^{(n)}(F) \ := \ \limsup_{N\to\infty} \frac{\mathrm{ex}_r^{(n)}(N, F)}{\binom{N}{r}}.$$

Theorem 6.3.6 (Hypergraphs as the case $n = 0$). An $r$-uniform hypergraph is precisely an $r$-uniform 0-superhypergraph.

Proof. If $n = 0$, then $\mathcal{P}^0(V_0) = V_0$. Hence an 0-superhypergraph $H = (V, E)$ satisfies $V \subseteq V_0$ and $E \subseteq \mathcal{P}(V) \setminus \{\emptyset\}$, which is exactly the usual definition of a (simple) hypergraph. The condition $|e| = r$ for all $e \in E$ is the standard $r$-uniformity condition. □

Definition 6.3.7 (Iterated singleton embedding). Fix $n \in \mathbb{N}_0$. Define $s_n : V_0 \to \mathcal{P}^n(V_0)$ recursively by

$$s_0(v) = v, \qquad s_{k+1}(v) = \{s_k(v)\} \ \ (k \geq 0).$$

Then $s_n$ is injective.

Theorem 6.3.8 (Embedding hypergraphs into $n$-superhypergraphs). Let $H = (V_0, E)$ be an $r$-uniform hypergraph and let $n \geq 0$. Define an $n$-superhypergraph $H^{(n)} = (V^{(n)}, E^{(n)})$ by

$$V^{(n)} := s_n(V_0), \qquad E^{(n)} := \big\{\{s_n(v) \mid v \in e\} \;\big|\; e \in E\big\}.$$

Then $H^{(n)}$ is $r$-uniform and $H$ is isomorphic (as an $r$-uniform hypergraph) to $H^{(n)}$.

Proof. For each $e \in E$ we have $|e| = r$, and since $s_n$ is injective,

$$\big|\{s_n(v) \mid v \in e\}\big| = |e| = r,$$

so $H^{(n)}$ is $r$-uniform. The map $s_n : V_0 \to V^{(n)}$ is a bijection and it sends edges of $H$ to edges of $H^{(n)}$ by construction, hence it is an isomorphism. □

Theorem 6.3.9 (Formal generalization of the Turán problem). Let $F$ be an $r$-uniform hypergraph and let $F^{(n)}$ be its embedded $r$-uniform $n$-superhypergraph as in Theorem 6.3.8. Then for every $N \in \mathbb{N}$,

$$\mathrm{ex}_r(N, F) \;=\; \mathrm{ex}_r^{(n)}(N, F^{(n)}).$$

Consequently, $\pi(F) = \pi^{(n)}(F^{(n)})$ (with the density normalization of Definition 6.3.5).

Proof. An $r$-uniform $n$-superhypergraph is, combinatorially, just an $r$-uniform hypergraph on the vertex set $V$ (the internal nature of supervertices does not affect incidence). Therefore forbidding $F^{(n)}$ as a subhypergraph is the same as forbidding the corresponding $r$-uniform incidence pattern of $F$. Hence the extremal edge count among $F$-free $r$-uniform hypergraphs on $N$ vertices equals the extremal edge count among $F^{(n)}$-free $r$-uniform $n$-superhypergraphs on $N$ vertices, i.e. $\mathrm{ex}_r(N, F) = \mathrm{ex}_r^{(n)}(N, F^{(n)})$. Dividing by $\binom{N}{r}$ and passing to $N \to \infty$ yields the density identity. □

Remark 6.3.10. If one introduces an additional constraint tying the supervertices to a fixed base size $|V_0|$, or if one forbids configurations in the expanded base-level hypergraph obtained from superedges, then genuinely new extremal behavior can occur. The formulation above is the cleanest Turán-type extension that is purely incidence-based.

## 6.4 Binary decision $n$-superhypertree

Binary decision hypertrees are rooted binary decision trees used to evaluate Boolean functions (cf. [292, 293]). We give a consistent extension in which nodes may be labeled by $n$-level supervertices.

Definition 6.4.1 (Hyperdiagram). Let $G = \{x_1, \ldots, x_n\}$ be a finite set. A hyperdiagram on $G$ is a pair $H = (G, \{E_k\}_{k=1}^m)$ such that $E_k \subseteq G$ and $|E_k| \geq 1$ for each $k \in \{1, \ldots, m\}$.

Definition 6.4.2 ($n$-Superhyperdiagram). Let $V_0$ be a finite base set and $n \in \mathbb{N}_0$. An $n$-superhyperdiagram is a pair $H = (V, \{E_k\}_{k=1}^m)$ such that

$$V \subseteq \mathcal{P}^n(V_0) \setminus \{\emptyset\}, \qquad E_k \subseteq V, \;\; |E_k| \geq 1 \;\; (1 \leq k \leq m).$$

The elements of $V$ are called supervertices and each $E_k$ is a superedge.

Theorem 6.4.3. An $n$-superhyperdiagram generalizes a hyperdiagram: when $n = 0$ and $V = V_0$, Definition 6.4.2 reduces to Definition 6.4.1.

Proof. If $n = 0$, then $\mathcal{P}^0(V_0) = V_0$, so $V \subseteq V_0$ and each $E_k \subseteq V$ with $|E_k| \geq 1$. Taking $V = V_0$ gives exactly a hyperdiagram on $V_0$. □

Definition 6.4.4 (Binary decision hypertree). Let $V_0 = \{x_1, \ldots, x_n\}$ be a set of Boolean variables and let $f : \{0,1\}^{V_0} \to \{0,1\}$ be a Boolean function. A binary decision hypertree is a rooted binary tree $T$ such that:

- each internal node is labeled by a variable $x_i \in V_0$;

- each internal node has exactly two outgoing edges labeled 0 and 1 (corresponding to the assignment of that variable);
- each leaf is labeled by an output value in $\{0, 1\}$;
- along any root-to-leaf path, the edge labels define a partial assignment $\alpha$, and the leaf label agrees with $f$ restricted to $\alpha$ (in particular, if the path assigns all variables then the leaf label is $f(\alpha)$).

Definition 6.4.5 (Binary decision $n$-superhypertree). Let $V_0$ and $f$ be as in Definition 6.4.4, and fix $n \in \mathbb{N}_0$. A binary decision $n$-superhypertree is a rooted binary tree $T$ such that:

- each node $p$ carries a state label $\lambda(p) \in \mathcal{P}^n(V_0)$ (a supervertex label);
- each internal node is additionally labeled by a queried variable $x_i \in V_0$;
- each internal node has exactly two outgoing edges labeled 0 and 1 (assigning the queried variable);
- each leaf is labeled by an output value in $\{0, 1\}$, consistent with $f$ under the assignment encoded by the path from the root.

No restriction is imposed on how $\lambda(p)$ is updated along the tree; different choices encode different superhyper-level state representations (e.g., sets of remaining variables, sets of sets of candidate variables, etc.).

Theorem 6.4.6. Binary decision $n$-superhypertrees generalize binary decision hypertrees: when $n = 0$ and one takes the state labels $\lambda(p) \in V_0$ to coincide with the queried variable at $p$, Definition 6.4.5 reduces to Definition 6.4.4.

Proof. If $n = 0$, then $\mathcal{P}^0(V_0) = V_0$. Choosing $\lambda(p)$ to equal the variable queried at $p$ yields exactly a rooted binary decision tree that branches on variables and labels leaves by outputs of $f$, which is a binary decision hypertree. □

# 7 Conclusion of this book

## 7.1 Conclusion

This book has established a rigorous mathematical foundation for developing SuperHyperGraph Neural Networks (SHGNNs) and Plithogenic Graph Neural Networks, thereby extending neural-network methodology beyond classical graphs and hypergraphs to nested, set-valued, and attribute-contradiction–aware relational structures.

Concretely, we formalized how hierarchical supervertices (arising from iterated powerset constructions) and higher-order superhyperedges can be converted into learning operators through well-defined lifting, propagation, and projection mechanisms, including normalized incidence-based convolutions and Laplacian-type operators on expanded representations. In parallel, we showed how plithogenic modeling can be incorporated into graph learning by encoding multi-valued attributes together with contradiction degrees, enabling message passing and aggregation that respect both graded appurtenance and incompatibility among attribute values.

As a result, the proposed frameworks provide principled tools for representing and learning from complex systems in which interactions are genuinely higher-order, hierarchical, and uncertain—for example, multi-level microservice call flows (pods $\to$ services $\to$ applications), overlapping group interactions with nested organization, and decision-making networks in which edge semantics depend on context-dependent, partially contradictory attributes.

## 7.2 Future Directions of this Research

This section highlights potential future directions for this research. A key objective is the practical implementation and experimental validation of the SuperHyperGraph Neural Network (SHGNN). Through computational experiments, we hope to discover related concepts that make the SHGNN more suitable for practical applications.

Another promising avenue is the exploration of extensions to SuperHyperGraph Neural Networks incorporating Fuzzy sets [47] and Neutrosophic sets [51]. This includes developing and validating frameworks such as Fuzzy SuperHyperGraph Neural Networks and Neutrosophic SuperHyperGraph Neural Networks. These frameworks aim to generalize Fuzzy Neural Networks [294] and Neutrosophic Neural Networks [295] by integrating the structural advantages of hypergraphs, laying the groundwork for advanced representations and computations. Additionally, future research could explore considerations involving Directed SuperHyperGraphs and their applications [201].

In addition to the concepts mentioned above, numerous frameworks for handling uncertainty, such as Soft Set (Soft Graph) [296, 297, 298], hypersoft set[63, 299, 300], Rough Set (Rough Graph) [301, 302], Hyperfuzzy set[303, 304], and Plithogenic Set (Plithogenic Graph) [97, 98], are well-known in the literature. Future research could explore how these concepts behave when applied to Graph Neural Networks, Hypergraph Neural Networks, and SuperHyperGraph Neural Networks. Such investigations could also shed light on whether these extensions result in more efficient and effective networks. This area holds significant potential for advancing understanding and innovation.

# Disclaimer

## Funding

This study did not receive any financial or external support from organizations or individuals.

## Acknowledgments

We extend our sincere gratitude to everyone who provided insights, inspiration, and assistance throughout this research. We particularly thank our readers for their interest and acknowledge the authors of the cited works for laying the foundation that made our study possible. We also appreciate the support from individuals and institutions that provided the resources and infrastructure needed to produce and share this book. Finally, we are grateful to all those who supported us in various ways during this project.

## Data Availability

This research is purely theoretical, involving no data collection or analysis. We encourage future researchers to pursue empirical investigations to further develop and validate the concepts introduced here.

## Ethical Approval

As this research is entirely theoretical in nature and does not involve human participants or animal subjects, no ethical approval is required.

## Use of Generative AI and AI-Assisted Tools

We use generative AI and AI-assisted tools for tasks such as English grammar checking, and We do not employ them in any way that violates ethical standards.

## Conflicts of Interest

The authors confirm that there are no conflicts of interest related to the research or its publication.

## Disclaimer

This work presents theoretical concepts that have not yet undergone practical testing or validation. Future researchers are encouraged to apply and assess these ideas in empirical contexts. While every effort has been made to ensure accuracy and appropriate referencing, unintentional errors or omissions may still exist. Readers are advised to verify referenced materials on their own. The views and conclusions expressed here are the authors' own and do not necessarily reflect those of their affiliated organizations.

# Appendix (List of Tables)

*

# Appendix (List of Figures)

*

Hypergraphs generalize classical graphs by allowing a single edge to connect multiple vertices, thereby providing a natural framework for modeling higher-order interactions. Superhypergraphs further extend this paradigm by accommodating nested, set-valued entities and relations, enabling the representation of hierarchical and multi-level structures that cannot be adequately captured by ordinary graphs or hypergraphs.

In parallel, neural networks—particularly Graph Neural Networks (GNNs)—have become a standard methodology for learning from relational data. Recent years have witnessed significant advances in Hypergraph Neural Networks (HGNNs), along with growing interest in their theoretical foundations.

To address uncertainty and multi-aspect attributes in complex networks, several graded and multi-valued graph frameworks have been developed, including fuzzy graphs and neutrosophic graphs. The plithogenic graph framework unifies and refines these approaches by incorporating multi-valued attributes together with explicit membership and contradiction mechanisms, offering a flexible and expressive representation for heterogeneous and partially inconsistent information.

This book develops the theoretical foundations of SuperHyperGraph Neural Networks (SHGNNs) and Plithogenic Graph Neural Networks, extending neural message-passing principles to these advanced higher-order structures. We introduce rigorous definitions, establish core structural properties, and prove well-definedness results for fundamental constructions. Particular emphasis is placed on strengthened formulations of Soft Graph Neural Networks and Rough Graph Neural Networks within these generalized settings. Although the present volume focuses primarily on mathematical development rather than empirical validation, it outlines directions for future computational and experimental research.

This book constitutes an expanded and substantially improved version of an earlier paper (Takaaki Fujita and Florentin Smarandache. *Superhypergraph neural networks and plithogenic graph neural networks: Theoretical foundations. Advancing Uncertain Combinatorics through Graphization, Hyperization, and Uncertainization: Fuzzy, Neutrosophic, Soft, Rough, and Beyond*, 2025). The current revision refines definitions, theorems, and proofs; streamlines previously elementary passages; improves the scope and accuracy of the related literature; and enhances overall coherence, structure, and readability.